%% file: main.tex
\documentclass{article}

\usepackage{PRIMEarxiv}

\usepackage[utf8]{inputenc}
\usepackage[T1]{fontenc}

\usepackage{url}

\usepackage{booktabs}
\usepackage{multirow}

\usepackage{graphicx}
\usepackage{float}
\graphicspath{{figures/}}

\usepackage{amsfonts}
\usepackage{amsmath}
\usepackage{amssymb}
\usepackage{nicefrac}

\usepackage{microtype}

\usepackage{xcolor}
\usepackage{colortbl}

\usepackage{fancyhdr}

\usepackage{hyperref}
\hypersetup{
  colorlinks=true,
  linkcolor=blue,
  citecolor=blue,
  urlcolor=blue
}

\title{Is Abliteration Getting Harder? Not Uniformly:\\ Refusal Removal by Direction-Based Ablation and ARA in Heretic on the Qwen3.8, Gemma~4, Olmo~3.1 and Mistral Small~3.2 LLMs Versus Their Predecessors}

\author{
  Richard J. Young \\
  University of Nevada Las Vegas \\
  Department of Information Systems \\
  Las Vegas, NV, USA \\
  \texttt{ryoung@unlv.edu} \\
}

\begin{document}
\maketitle

\begin{abstract}
\input{sections/01_abstract}
\end{abstract}


\input{sections/02_introduction}
\input{sections/03_related_work}
\input{sections/04_methods}
\input{sections/05_results}

\input{sections/06_discussion}   %
\input{sections/07_conclusion}

\section*{Ethics Statement}
The study involved no human participants and no personal data and was not
human-subjects research, so no IRB review was required. It creates and
releases language models whose refusal behaviour has been removed, which is
dual-use. The models are released openly with two exceptions. The five
Gemma-2-27B variants are withheld (their repositories are private), because the
Gemma Prohibited Use Policy may forbid redistributing them (see below). The
five OLMo-2-32B variants (DiM, MPOA, SOM, ARA and $\gamma$-MPOA) are gated on
Hugging Face with automatic approval, because OLMo-2-32B is the only base model
studied for which no third-party public abliteration was found. For the other nine bases
(the Gemma-2-27B variants among them are withheld regardless),
refusal-removed derivatives were already public: for five of them, public
releases were evaluated here (\S\ref{sec:res-secondary}), and for the other
four (Gemma-2-27B, Mistral-Small-2409, Qwen2.5-7B and Qwen3.5-4B),
third-party abliterated or uncensored derivatives were found on Hugging
Face\footnote{For example
\href{https://huggingface.co/byroneverson/gemma-2-27b-it-abliterated}{\nolinkurl{byroneverson/gemma-2-27b-it-abliterated}},
\href{https://huggingface.co/byroneverson/Mistral-Small-Instruct-2409-abliterated}{\nolinkurl{byroneverson/Mistral-Small-Instruct-2409-abliterated}}
and
\href{https://huggingface.co/huihui-ai/Huihui-Qwen3.5-4B-abliterated}{\nolinkurl{huihui-ai/Huihui-Qwen3.5-4B-abliterated}};
accessed 8 November 2026.}
whose residual refusal was not evaluated. The argument is therefore that
comparable models are already available, not that they are equally effective.
The release serves two purposes that the configurations and journals alone do
not: it lets others re-run the audit on the evaluated weights, which matters
because re-exports from stored configurations are not exactly reproducible
(the ARA refits drifted when rebuilt on other hardware;
Appendix~\ref{app:dev-rescore}), and it provides fixed edited models against
which defences and detectors can be evaluated. Responses to harmful prompts
are stored privately for scoring and are not released. Released models (all
except the five withheld Gemma-2-27B variants) keep the licence of their base model: Apache-2.0 (the Qwen, OLMo and
Mistral-Small-3.2 bases and Gemma-4-31B), to which the use restrictions of
the OpenRAIL++ licence are added; or, for Mistral-Small-2409, the Mistral Research
Licence, which permits non-commercial research use. Apache-2.0 and the
Mistral Research Licence permit redistributing such derivatives, the latter
for research purposes only. The Gemma Prohibited Use Policy, which the Gemma
Terms incorporate, forbids ``attempts to override or circumvent safety
filters'', so this study cannot state that redistributing the refusal-removed
Gemma-2-27B derivatives is permitted, and they are not released
(Appendix~\ref{sec:setup-export}). The
automatic-approval gate on the OLMo-2-32B variants requires a logged-in
Hugging Face account and shares the requester's contact details with the
author, so access is recorded; it does not limit who can obtain the weights. A licence states the permitted use; it is not a
technical control.

\section*{Data and Code Availability}
The 42 primary models and the four $\gamma$-MPOA control models are on
Hugging Face under \url{https://huggingface.co/richardyoung}
(\texttt{<base>-<METHOD>-heretic}); the five Gemma-2-27B variants are
withheld (private repositories), the five OLMo-2-32B variants are gated with
automatic approval, and the others are public. The release of code and data consists of
the code (the Heretic fork branches \texttt{paper33-plugins} and
\texttt{paper33-gamma}, the evaluation harness and the analysis scripts), the
configuration of every run, the Optuna study journals and run logs, and the
judged labels of every held-out response (judge, keyword and degeneracy labels
per prompt and model, without the response text). The held-out prompts come
from the public sets of Table~\ref{tab:refusal-sets}. The analysis code,
harness, configurations, journals, logs and labels are at
\url{https://github.com/ricyoung/abliteration-studies}, and the fork branches
at \url{https://github.com/ricyoung/heretic}. The Heretic fork is released
under AGPL-3.0, the licence of upstream Heretic; the other code and the data
are released under OpenRAIL++.

\bibliographystyle{unsrt}
\bibliography{references}

\newpage
\appendix
\input{sections/08_appendix}

\end{document}

%% file: sections/01_abstract.tex
Abliteration, a fine-tuning-free weight edit, removes refusal from open-weight
large language models (LLMs). The developer of Arbitrary-Rank Ablation (ARA), an
optimisation-based variant, reports that direction-based abliteration, which
ablates a refusal direction, increasingly fails on newer models. On five
release pairs, refusal removal did not become uniformly harder. Qwen2.5-32B and
Qwen3.8-27B, Gemma-2-27B and Gemma-4-31B, OLMo-2-32B and Olmo-3.1-32B,
Mistral-Small-2409 and Mistral-Small-3.2, and Qwen2.5-7B and Qwen3.5-4B were
edited in Heretic~2.0 by three or four direction-based methods
(difference-in-means, MPOA, SOM and, on the small pair, J-space) and ARA under
the same prompts, nominal search budget, first-token KL divergence cap and
selection rule; removal counted as harder when the automatic WildGuard judge
found more of 972 held-out harmful prompts (JailbreakBench, XSTest,
StrongREJECT, HarmBench and an author-compiled malicious-code bank) still
refused. After direction-based removal, newer releases retained $-$1 to +37
percentage points more refusal, and after ARA $-$26 to +34. With run-to-run
variance included, between-pair heterogeneity of the direction-based effect
ranged from high to low with the scale and variance model ($I^2$ 20--80\%).
ARA lost less effectiveness than direction-based methods in Gemma and OLMo and
more in Qwen; each verdict rests on one ARA run, and some changed with the
divergence budget or variance model. In Gemma the direction-based increase came
from the malicious-code bank and reversed on public sets. Contrary to
prediction, keyword counts at the tool's window did not undercount judged
refusal on average, but differed by a mean 7 points per model. Audits of a
single pair, method or prompt source can reach opposite conclusions. Whether a
newer model resisted removal depended on the release pair, method and budget,
not on generation alone.

%% file: sections/02_introduction.tex
\section{Introduction}
\label{sec:intro}

Open-weight language models ship with safety training that teaches them to
refuse harmful requests, but their public weights allow this behaviour to be
edited after release. Since 2024, fine-tuning-free techniques called
abliteration, which need no end-to-end gradient training, have made such
editing cheap: with a few hundred contrasting prompts and a
single GPU, they modify a few weight matrices so that the model stops
refusing \cite{arditi2024refusal,labonne2024abliteration}, without the
gradient-based fine-tuning on harmful data used by earlier attacks
\cite{qi2024finetuning,lermen2023lora}. Many abliterated
models are published on Hugging Face, and fully automated tools are publicly
available \cite{heretic2025}. Whether safety training withstands such editing
therefore matters to attackers, to defenders and to anyone who audits whether
newer releases are more robust than older ones.

The original technique rests on the finding that, in many chat models,
refusal is mediated by a single residual-stream direction whose removal from
the weights suppresses refusal while leaving most other behaviour intact
\cite{arditi2024refusal}. Some successors change only how this direction is
removed, to limit collateral damage \cite{lai2025mpoa}, whereas others remove
several directions or a refusal ``cone''
\cite{piras2025som,wollschlager2025geometry}. Arbitrary-Rank Ablation (ARA),
now the default of the open-source Heretic~2.0 tool,
abandons directions and optimises each targeted weight matrix to move harmful
outputs toward harmless ones while keeping harmless outputs fixed
\cite{weidmann2026ara}. The methods are motivated by different views of where
refusal lives, from one direction to a low-dimensional subspace or cone, a
non-linear region (so far suggested only by visualisation
\cite{hildebrandt2025refusal}) or a distribution across modules. They also
differ in edit rank (one to six for the direction-based methods used here, 50
for ARA), objective and search space, so comparing them compares families of
edits, not geometric hypotheses about refusal.
If refusal is not one-dimensional, single-direction
defences (\S\ref{sec:related}) may give a false sense of security, yet recent
work offers mixed evidence on the single-direction picture
\cite{joad2026there,wang2026on,pan2025hidden}.

Whether newer model generations resist refusal removal remains unclear. The
author of ARA reports that older models ``can usually be decensored quite
easily with standard abliteration'', whereas ARA ``shows its strengths
precisely with the latest generation of models, where traditional directional
ablation tends to be woefully inadequate'' \cite{weidmann2026ara}. One explanation
would be that newer safety training spreads refusal over more dimensions or
modules, consistent with calls for alignment ``more than just a few tokens
deep'' \cite{qi2025safety}, with evidence that harmfulness and refusal are
encoded separately \cite{zhao2025harmfulness}, and with training methods that
make refusal survive ablation of its direction
\cite{yu2025refusal,xie2025beyond}. Adversarial safety fine-tuning of one 7B
model, however, moved a still low-dimensional refusal representation to an
earlier layer rather than expanding it \cite{lan2026dynamic}. Other
explanations would not involve the representation of refusal at all:
architectural changes can weaken a given edit (in post-normalised blocks part
of an ablated direction reaches the residual stream; Appendix~\ref{app:setup-models}),
later layers can partly compensate for an edit \cite{mcgrath2023hydra}, and the estimate of the
direction can be contaminated by its prompt sets.
To the best of
current knowledge, the ARA author's claim has not been tested independently: the nearest
evidence concerns safety-pretraining checkpoints of one 1.7B model
\cite{agnihotri2025granular}, ARA has no published evaluation, and existing
abliteration comparisons \cite{young2025comparative} predate both ARA and
current open models. A second
gap is methodological: tools are usually evaluated with the cheap signals they
optimise (keyword-matched refusals over short generations and first-token KL
divergence), which can disagree sharply with stronger judges: keyword matching
agrees poorly with human labels \cite{xie2025sorrybench}, and conclusions can
depend on the classifier used \cite{young2026measuring}. In one public
comparison, keyword matching reported 3--5 refusals per 100 prompts where an
LLM judge found more than 33 (Heretic PR~\#196, an unreviewed practitioner
report\footnote{Heretic pull requests (PR~\#) and issues (\#) at
\url{https://github.com/p-e-w/heretic} are cited in this paper as unreviewed
practitioner reports, not as peer-reviewed evidence. Each is at
\url{https://github.com/p-e-w/heretic/pull/<N>} or \url{.../issues/<N>}; they
were retrieved on 2--5 November 2026 and may have changed since.}).
Aggregate refusal rates can also hide failures in domains such as
cybersecurity \cite{hadetskyi2026not} and malicious code generation, for which
validated prompt banks and systematic corpus reviews now exist
\cite{young2026refusal,young2026a,young2026code}. A method can therefore
appear successful by avoiding refusal phrases without complying, or by
complying with degraded, useless output \cite{nikolic2025jailbreak}.

The present study addresses both gaps through a controlled comparison of
refusal-removal methods on an earlier and a later release of four open-weight
model families and one smaller pair, five release pairs in all. To current
knowledge it is the first independent comparison of ARA with direction-based
methods across model releases; it rests on one ARA run per model. The primary purpose
was to determine whether refusal removal becomes harder in newer model
generations. Removal is called harder for a model when more refusal remains
after the same search budget and selection rule, that is, when its residual
refusal (the share of held-out harmful prompts whose response an automatic
judge, WildGuard, labels a refusal) is higher. The judge is independent of the
optimisation objective but has not been validated against human labels on
edited models. This is a budget-bound measure: residual refusal of the trial
selected after 100 search trials at first-token KL divergence $\le 0.1$, not
the difficulty of removal for an attacker with more budget.
It was hypothesised (H1) that,
with every model selected under the same cap on first-token KL divergence
($\le 0.1$), residual refusal after ablation would be higher for the newer
model of each pair than for its predecessor, for every method. Because a
newer original may already refuse more before any edit, the generation effect
is also reported relative to the originals' own difference (baseline-adjusted),
which separates higher starting refusal from resistance to removal; this
contrast was added after H1 was stated and is exploratory.
The secondary purpose was to determine whether this difficulty depends
on the class of edit. It was hypothesised (H2) that the generation effect would
interact with method class, such that optimisation-based ablation (ARA) would
lose less effectiveness on newer models than direction-based methods
(diff-in-means, projected/magnitude-preserving abliteration (MPOA),
self-organising-map (SOM) ablation and, on a small pair, Jacobian-space
(J-space) ablation). Because the two classes differ in rank, objective and
search space as well as in their motivating geometry, H2 compares edit
families and cannot by itself separate these factors. A third, exploratory aim was to test whether the cheap
optimisation signals agree with independent measures; it was hypothesised
(H3) that keyword-based refusal counts would underestimate residual refusal
relative to an automatic judge, and that the size of this gap would differ
between method classes. All methods run inside Heretic~2.0 with the same
prompts and nominal search budget on four main-study pairs (Qwen, Gemma, OLMo,
Mistral; 22--32B parameters)
and a small Qwen pair (7B, 4B).
Models selected by a fixed rule are released (except the five Gemma-2-27B
variants, which are withheld; Appendix~\ref{sec:setup-export}), run logs, configurations and
study journals are released with the code (Data and Code Availability), and
every model is scored by one harness,
independent of every optimisation objective, for judged refusal, capability
(knowledge, reasoning, instruction following \cite{young2025when} and code
generation) and full-sequence distribution shift.

The study makes three contributions:
\begin{itemize}
\item a held-out test of whether refusal removal is harder in newer releases,
  with the release pair as the unit: the generation effect differed in sign
  and size between pairs, the only raw per-pair effect that exceeded
  run-to-run variation under both variance models, at the point estimate of the
  run-to-run standard deviation, was the OLMo direction-based increase, and
  some per-pair verdicts depended on the source of the test prompts
  (\S\ref{sec:res-h1});
\item a preliminary comparison of ARA with direction-based methods under one
  nominal search budget, KL cap and selection rule: ARA lost less
  effectiveness on the newer model in two pairs, more in one in its point
  estimate (indeterminate with run-to-run variance) and about as much in two,
  but each contrast rests on one ARA run whose export was
  refitted after the search, and the verdicts changed with the KL budget and
  the variance model (\S\ref{sec:res-h2});
\item evidence on the signals that abliteration tools optimise: keyword
  counts at the tool's 100-token window matched judged refusal on average but
  disagreed with it by several points per model, and first-token KL
  divergence did not bound capability loss (\S\ref{sec:res-h3},
  \S\ref{sec:res-secondary}).
\end{itemize}

\begin{figure*}[!t]
\centering
\includegraphics[width=\textwidth]{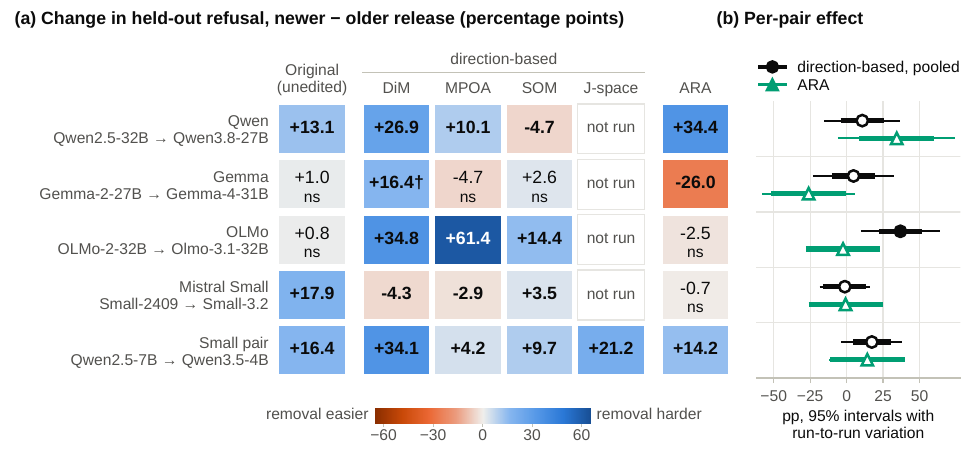}
\caption{\textbf{Refusal removal did not become uniformly harder.} (a) Change in held-out refusal from the older to
the newer release of each pair (newer minus older, percentage points; WildGuard judge, 972 harmful prompts not used
for search or selection), for the unedited originals and for the model each removal method selected (fewest dev
refusals at first-token KL $\le$ 0.1, seed 1). Blue: the newer release kept more refusal (removal harder); orange:
less (removal easier). Bold: significant (methods: exact McNemar, Holm over the 23 H1 tests; originals: exact
McNemar, unadjusted); ns: not significant; $\dagger$ failed edit (Gemma-4-31B DiM left more refusal than its own
original). (b) Per-pair effect of the direction-based methods (pooled over DiM, MPOA and SOM; J-space added on the
small pair) and of ARA, with 95\% intervals that add run-to-run variation at its point estimate to prompt sampling:
thick bars assume constant run-to-run variance, thin bars variance that scales with the refusal level. Filled
marker: both intervals exclude zero, which holds only for the OLMo direction-based increase. Refusal levels and
Wilson intervals of every cell are in Figure~\ref{fig:heldout-generation}.}
\label{fig:at-a-glance}
\end{figure*}

Figure~\ref{fig:at-a-glance} summarises the held-out answer to H1 and the per-pair effects.

%% file: sections/03_related_work.tex
\section{Related Work}
\label{sec:related}

Refusal-removal methods are reviewed below by what they assume about the
geometry of refusal, followed by work on how post-training shapes a behaviour,
defences and evaluation practice (definitions
in Appendix~\ref{app:method-defs}).

\subsection{Single-Direction Ablation}

Arditi et al.~\cite{arditi2024refusal} showed causally, across 13 chat models,
that adding the difference between mean harmful and harmless activations
induces refusal and ablating it suppresses refusal. Contemporaneous work read
and steered such directions from the top down: representation engineering
extracted directions for concepts including harmlessness
\cite{zou2023representation}, and contrastive activation addition steered
refusal, among other behaviours, in Llama~2 chat models with mean-difference
vectors added to the residual stream \cite{panickssery2023steering}.
Orthogonalising
the weights that write to the residual stream against this direction yields a
fine-tuning-free edit that needs only a few hundred prompts
\cite{labonne2024abliteration,failspy_abliterator}, consistent with the linear
representation hypothesis \cite{park2024linear}. The edit is fast and
interpretable, but the direction also carries part of the harmless mean, an
offset that an affine model of refusal handles explicitly
\cite{marshall2024refusal}; changed row norms shift activation scales, and
results depend on a hand-chosen layer and strength, with large capability
differences between tools \cite{young2025comparative}
(Appendix~\ref{sec:m-dim}).

Most successors address this collateral damage rather than the
one-dimensional assumption. Projected abliteration removes the component along
the harmless mean \cite{lai2024projected}, magnitude-preserving abliteration
(MPOA) restores row norms \cite{lai2025mpoa}, and Heretic automates the choice
of layer, strength and direction with TPE \cite{heretic2025,akiba2019optuna}. A
null-space-constrained variant \cite{jwest33abliterator} adopts the AlphaEdit
constraint from knowledge editing \cite{fang2025alphaedit,meng2022locating},
which may also block the refusal edit when harmful and harmless activations
share directions. Surgical refusal ablation orthogonalises the direction
against capability and style directions \cite{cristofano2026surgical}, whereas
Petrov~\cite{petrov2026on} found that a topic-matched harmless baseline yielded
no functional refusal direction. The evidence remains thin: the projection and
norm-preservation variants are documented mainly in blog posts, and both
preprints evaluate models of at most 14B parameters.

\subsection{Multi-Direction and Subspace Methods}

A second group holds that one mean difference captures only the dominant
component of refusal. Gabliteration projects out the top singular vectors of
per-prompt activation differences \cite{gabliteration2024}, OBLITERATUS adds
whitened, iterative and spectral variants \cite{obliteratus2026}, fast subspace
estimation finds multi-dimensional refusal subspaces \cite{winninger2026fast},
and weighted refusal direction optimisation (WRDO) learns weights over
per-prompt directions, with Heretic as a baseline \cite{evseev2026wrdo}.
Wollschl\"ager et al.~\cite{wollschlager2025geometry} learned refusal
directions by gradient descent, found multi-dimensional concept cones, and
showed that orthogonality does not imply independence under intervention.
Piras et al.~\cite{piras2025som} proved that self-organising maps (SOMs)
generalise difference-in-means and reported that ablating several SOM
directions outperforms the single-direction baseline. Each added direction,
however, removes capacity. The premise is also
contested: Joad et al.~\cite{joad2026there} found distinct refusal directions
that produced nearly identical refusal--over-refusal trade-offs when steered,
Wang et al.~\cite{wang2026on} found safety-triggered refusal to be
one-dimensionally steerable, and in Llama-3-8B one dominant direction governed
refusal while smaller orthogonal directions, such as role-play, modulated it
\cite{pan2025hidden}. A multi-dimensional representation therefore does not
establish that multi-directional removal is necessary.

\subsection{Optimisation-Based Methods}

A third group replaces the estimated direction with an explicit objective.
Arbitrary-Rank Ablation (ARA) fits, for each targeted module, a low-rank
update that keeps harmless outputs fixed and moves harmful outputs toward them
\cite{weidmann2026ara}, so any refusal mechanism carried by module outputs is
in scope up to the adapter's rank. Its author reports that directional ablation
increasingly fails on recent models and that ARA outperforms it on gpt-oss-20b
(Appendix~\ref{sec:m-ara}), but this evidence comes from a pull request, and
the nearest-neighbour objective may overfit its prompt sets. A low-rank edit may suffice because safety-critical weights occupy about 3\% of parameters and 2.5\% of rank,
separately from utility \cite{wei2024assessing}, although newer models may
spread safety more widely. Optimal-transport ablation also
matches distributions, with a closed-form Gaussian map \cite{nanfack2026ot},
but no public code was found (Appendix~\ref{sec:m-not}).

\subsection{Representation-Space Methods}

A fourth group changes the space in which refusal is sought. Jacobian-space
(J-space) ablation computes the direction in a workspace defined by a Jacobian
lens and pulls it back to the residual stream
\cite{sekinal2026jspace,gurnee2026verbalizable,neuronpedia2026jacobian}, to
isolate the part of the refusal signal the model uses downstream. By its
authors' own measurements, however, it removes little refusal, and the
reference code is a single-model forward hook (Appendix~\ref{sec:m-jspace}). Related findings suggest that the
residual-stream mean difference mixes signals: harmfulness is encoded
separately from refusal \cite{zhao2025harmfulness,yeo2025understanding}, and at
fixed intervention strength a direction estimated at a layer's output, before
the residual stream, controlled refusal whereas one estimated in the summed
stream did not \cite{bosco2026locating}. Where a direction is estimated may
matter as much as how many are removed.

\subsection{Post-Training Mechanisms and Self-Repair}

Two further literatures bear on why a newer release could resist directional
removal without a higher-dimensional refusal representation. Mechanistic
studies of alignment algorithms found that preference optimisation (DPO) left
a model's toxic capability in place and learned to bypass it
\cite{lee2024mechanistic}, and that safety fine-tuning applied a minimal
transformation to MLP weights that routes unsafe inputs towards the null
space of those weights \cite{jain2024makes}. How a release's post-training
recipe implements refusal may therefore matter as much as how many directions
refusal occupies. Ablation studies have also found self-repair, in which later
layers compensate for an ablated component \cite{mcgrath2023hydra}; a model
that compensates more strongly would retain more refusal after the same edit.
Neither property was measured in the models studied here.

\subsection{Defences and Evaluation Problems}

Defences largely target the single-direction estimator. Adversarial training
against ablation of the refusal feature cut its attack success from 53\% to
10\% on Llama-3-8B \cite{yu2025refusal}; extended-refusal fine-tuning spreads
the refusal signal across tokens, limiting refusal drops under abliteration to
at most 10\% versus 70--80\% in baseline models \cite{abushairah2025defense};
and fine-tuning under random ablation of the direction teaches models to
rebuild refusal, cutting attack success by about 95\% \cite{xie2025beyond}.
Refusal aliases obscure the direction \cite{truong2026abliteration}; decoy
directions corrupt the contrastive estimator, cutting a Heretic weight-level
attack on Llama-3-8B-Instruct from 88.7\% to 18\% success
\cite{muhamed2026decoy}; and decoy hardening concedes refusal removal but turns
most answers to hazardous requests into falsified decoys
\cite{russinovich2026fools}. Circuit breakers instead act on harmful-output
representations \cite{zou2024circuit}, and tamper-resistant training targets
fine-tuning attacks \cite{tamirisa2025tamper,rosati2024representation}, yet
such fine-tuning defences still allowed 79--98\% success to a best-of-96
single-layer abliteration on Llama-3.1-8B, against 99\% without defence
\cite{kuo2026openweight}. Training-time defences could make later releases
resist directional removal, whereas defences built around direction
estimation may lose value if removal no longer relies on it.

Evaluation practice is a second weakness. Tools are optimised and reported
with keyword-matched refusal counts and first-token KL divergence, yet
conclusions can depend on the classifier \cite{young2026measuring}, including
in abliteration studies \cite{agnihotri2025granular}, and in one unreviewed
practitioner report keyword counts fell far below an LLM judge's (Heretic
PR~\#196).
Against human labels, keyword matching reached
Cohen's $\kappa=0.37$, WildGuard 0.61 and fine-tuned LLM judges about 0.81
\cite{xie2025sorrybench}, although on unmodified models keyword matching
tracked a GPT-4 judge closely (gaps of 0.2--5.5 points across six models) \cite{cui2025orbench}, so large gaps may arise only after
modification. Compliance is also not competence:
single-vector projection raised a security-task score only from 0.46 to 0.50
while raising unsafe compliance from 0.10 to 0.47 \cite{david2026ablating};
abliterated code models complied with requests to inject vulnerabilities but succeeded at
rates bounded by model size \cite{carleo2026willing}; and jailbreaks of models
aligned to refuse benign topics cut accuracy by up to 92\%
\cite{nikolic2025jailbreak}. Off-target effects include more optimistic
decisions on a task that elicited no refusals \cite{fafua2026abliteration} and
script-dependent bias that abliteration did not remove \cite{ko2026uncensored}.
Guardrails have proved brittle
under adversarial pressure \cite{young2025evaluating}, and durability
claims are easily overstated because small changes to the attack
configuration can reverse them \cite{qi2025durability}.

\subsection{Positioning of the Present Study}

The present study differs from this literature in four respects. First,
whether removal becomes harder in newer model generations has been tested only
indirectly: with one directional pipeline, susceptibility across 24 open
models was model-specific rather than size-dependent, with safety-training
methodology and architecture the strongest predictors \cite{hadetskyi2026not};
in one 1.7B model the full safety-pretraining recipe was the most robust
to abliteration, and Qwen3 resisted it in one setup \cite{agnihotri2025granular}; and the earlier cross-tool comparison
covered 7--14B models that predate ARA and the current generation
\cite{young2025comparative}. A companion benchmark \cite{young2026paper30}
uses the same harness to compare Heretic 1.x, Heretic 2.0 and community
releases within the Qwen2.5 family. The two studies share the evaluation
harness, the held-out prompt set and 19 of the models evaluated here: the
Qwen2.5-7B, Qwen2.5-32B and Qwen3.5-4B originals, their seed-1 DiM, MPOA, SOM
and ARA models (and J-space on the two small models), and the two community
releases of Qwen2.5-32B. For these 19 models the companion study uses the same
stored responses (identical checksums) and therefore the same judge and
keyword labels. The Qwen3.8-27B, Gemma, OLMo and Mistral Small models, the
seed-2 replicates, the $\gamma$-MPOA controls and the other six community
releases are used only here, and the comparison between releases (H1, H2) is
reported only here.
Here, earlier- and later-generation models are
paired within families. Second, all four method classes (representation-space
methods on the small pair only) share one framework, prompt sets, budget and
trial-selection rule under one cap on first-token KL divergence. Third, a harness independent of all optimisation
objectives scores judged refusal on held-out prompts, capability and
full-sequence distribution shift. Fourth, the study provides, to current
knowledge, the first independent comparison of ARA with direction-based
methods across model releases, with one ARA run per model.
Table~\ref{tab:methods} summarises the methods considered. Presets of the
external OBLITERATUS toolkit are evaluated in the companion paper
\cite{young2026paper30}; null-space-constrained
abliteration, refusal cones, WRDO and optimal-transport ablation are not
evaluated (Appendix~\ref{sec:m-not}); and fine-tuning attacks, which remove
safety training cheaply but require gradient updates
\cite{qi2024finetuning,lermen2023lora}, are out of scope.

\begin{table*}[t]
\centering\small
\caption{Refusal-removal methods considered in this study, as instantiated
(\S\ref{sec:setup-methods}). ``Search'' = automated hyperparameter
optimisation; ``Export'' = produces a standalone model. SVD multi-direction
(OBLITERATUS) is evaluated in the companion paper \cite{young2026paper30}, not
here; null-space-constrained
abliteration and refusal cones are not evaluated (Appendix~\ref{sec:m-not}).
Formal definitions are given in Appendix~\ref{app:method-defs}.}
\label{tab:methods}
\setlength{\tabcolsep}{4pt}%
\begin{tabular}{@{}llllll@{}}
\toprule
Method & Assumption about refusal & Edit & Search & Export & Implementation \\
\midrule
Diff-in-means \cite{arditi2024refusal} & 1 direction & rank-1 projection & TPE & yes & Heretic 2.0 (no proj./norm) \\
Projected / MPOA \cite{lai2024projected,lai2025mpoa} & 1 direction, $\perp$ harmless & rank-1 + norm$^{a}$ & TPE & yes & Heretic 2.0 \\
SVD multi-dir.\ \cite{gabliteration2024} & $k$-dim subspace & rank-$k$ projection & -- & yes & OBLITERATUS$^{b}$ \\
SOM \cite{piras2025som} & $k$ cluster directions & rank-$k$ + norm$^{a}$ & TPE & yes & Heretic 2.0 fork$^{c}$ \\
ARA \cite{weidmann2026ara} & none (per-module map) & rank-50 LoRA & TPE + L-BFGS & yes & Heretic 2.0 \\
J-space & 1 direction in J-space & rank-1 + norm$^{a}$ & TPE & yes & Heretic 2.0 fork$^{d}$ \\
\bottomrule
\end{tabular}

\smallskip
\footnotesize $^{a}$Row-norm restoration approximated by a LoRA adapter of rank 3
(MPOA, J-space) or 6 (SOM).
$^{b}$Presets \texttt{advanced} and \texttt{spectral\_cascade} of OBLITERATUS
v0.1.4 \cite{obliteratus2026}; not evaluated in this study. These presets and
\texttt{basic} were run on Qwen2.5-7B-Instruct and scored with this study's
harness in the companion paper \cite{young2026paper30}; v0.1.4 does not
support Qwen3.5-4B, so it offers no older--newer contrast.
$^{c}$Port of Heretic PR~\#196 with the deviations listed in
Appendix~\ref{sec:setup-deviations}.
$^{d}$Small pair only; directions from \texttt{jspace-refusal} baked into the
weights without harmless-mean projection; the reference forward hook is planned.
\end{table*}

%% file: sections/04_methods.tex
\section{Methods}
\label{sec:setup}

The study crosses model generation with refusal-removal method under one
selection rule, one evaluation harness independent of the optimisation
objective and the same nominal budget of 100 trials (exceptions in
Appendix~\ref{sec:setup-deviations}), since method comparisons depend on the
search budget \cite{dodge2019show}; further detail is in
Appendix~\ref{app:setup-details}. Figure~\ref{fig:workflow} summarises the
workflow.

\begin{figure*}[tp]
\centering
\includegraphics[width=\textwidth]{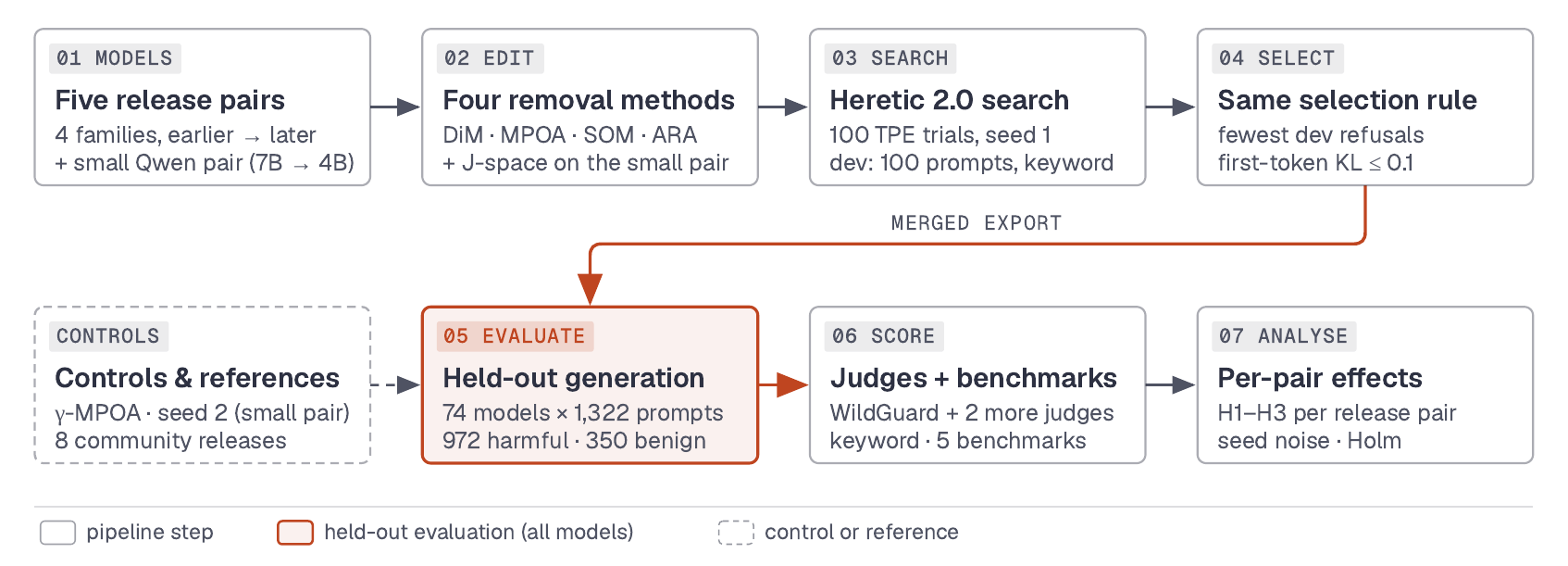}
\caption{\textbf{Study workflow.} Five earlier-to-later pairs of instruction-tuned models are edited with four refusal-removal methods in Heretic 2.0 (five on the small pair, which adds J-space), each in a 100-trial Optuna TPE study that minimises dev keyword refusals and first-token KL divergence; the trial with the fewest dev refusals at KL $\le$ 0.1 is exported as standalone weights. The originals, the selected models and eight public community abliterations are evaluated on held-out data (judged refusal on 1,322 prompts, of which the 44 that exactly (36) or approximately (8) match a prompt of the search dataset are flagged; teacher-forced divergence on the 350 benign prompts; five capability benchmarks), and the statistics test H1--H3 and compare dev-set with held-out refusal. The highlighted step is the held-out evaluation, applied to all 74 models; the dashed box marks the $\gamma$-MPOA control, the seed-2 replicate of the small pair and the community releases, which join the same evaluation. Keyword refusal is scored on the first 100, 256 and 512 generated tokens.}
\label{fig:workflow}
\end{figure*}

\subsection{Models and Data}
\label{sec:setup-models}

The main study uses one dense instruction-tuned model per family per
generation, of similar scale, each run unquantised in bf16 on a single GPU
(Table~\ref{tab:models}). The families are Qwen, Gemma, OLMo and Mistral; OLMo
was included because its training data are openly released. Paired models also
differ in size and partly in architecture, and the later model of each pair
was released 9 to 23 months after the earlier one (release months and
checkpoint sizes in Table~\ref{tab:models}).

\begin{table*}[t]
\centering\small
\caption{Models. \emph{Released}: year and month in which the Hugging Face
repository was created (a repository can be created before its public
release; note c). \emph{Revision}: first eight hexadecimal digits of the
Hugging Face commit used. \emph{Size}: bf16 checkpoint size hashed by the
evaluation harness, in GB (image-text checkpoints include their vision
encoder); nominal scale from the repository name in parentheses.
\emph{Layers}: transformer blocks reported by Heretic. Short model names
follow the vendors' spelling (OLMo-2-32B, Olmo-3.1-32B; Mistral-Small-2409 for
\texttt{Mistral-Small-Instruct-2409}); the pairs are labelled Qwen, Gemma, OLMo,
Mistral Small and small pair.}
\label{tab:models}
\setlength{\tabcolsep}{4pt}
\resizebox{\textwidth}{!}{%
\begin{tabular}{@{}llllllr>{\raggedright\arraybackslash}p{3.1cm}@{}}
\toprule
Family & Gen. & Released & Repository & Revision & Size (GB) & Layers & Notes \\
\midrule
\multicolumn{8}{l}{\emph{Main study}} \\
Qwen    & earlier & 2024-09 & \texttt{Qwen/Qwen2.5-32B-Instruct}                       & \texttt{5ede1c97} & 65.5 (32B) & 64 & -- \\
Qwen    & later   & 2026-08 & \texttt{Qwen/Qwen3.8-27B}                                & \texttt{1d4bf0f2} & 55.6 (27B) & 64 & hybrid linear/full attention; image-text \\
Gemma   & earlier & 2024-06 & \texttt{google/gemma-2-27b-it}                           & \texttt{aaf20e6b} & 54.5 (27B) & 46 & post-norm \cite{gemmateam2024gemma2}; no system role \\
Gemma   & later   & 2026-04$^{c}$ & \texttt{google/gemma-4-31B-it}                     & \texttt{842da379} & 62.6 (31B) & 60 & post-norm; image-text \\
OLMo    & earlier & 2025-03 & \texttt{allenai/OLMo-2-0325-32B-Instruct}                & \texttt{b9602434} & 64.5 (32B) & 64 & post-norm \cite{walsh2025olmo2} \\
OLMo    & later   & 2025-12 & \texttt{allenai/Olmo-3.1-32B-Instruct}                   & \texttt{ac0587e4} & 64.5 (32B) & 64 & post-norm; sliding-window attention$^{b}$ \\
Mistral & earlier & 2024-09 & \texttt{mistralai/Mistral-Small-Instruct-2409}           & \texttt{4600506f} & 44.5$^{d}$ (22B) & 56 & -- \\
Mistral & later   & 2025-06 & \texttt{mistralai/Mistral-Small-3.2-24B-Instruct-2506}$^{a}$ & \texttt{3ed8d341}$^{a}$ & 48.1 (24B) & 40 & image-text \\
\midrule
\multicolumn{8}{l}{\emph{Small pair}} \\
Qwen    & earlier & 2024-09 & \texttt{Qwen/Qwen2.5-7B-Instruct}                        & \texttt{a09a3545} & 15.2 (7B) & 28 & -- \\
Qwen    & later   & 2026-02 & \texttt{Qwen/Qwen3.5-4B}                                 & \texttt{851bf6e8} & 9.3 (4B)  & 32 & hybrid linear/full attention; image-text \\
\bottomrule
\end{tabular}}

\smallskip
\footnotesize $^{a}$Loaded from the mirror
\texttt{unsloth/Mistral-Small-3.2-24B-Instruct-2506} at commit
\texttt{3ed8d341}, whose weight shards are byte-identical to the official
repository (Appendix~\ref{sec:setup-deviations}).
$^{b}$On three of every four layers (48 of 64, every fourth layer using full
attention; 4{,}096-token window), as specified by \texttt{layer\_types} and
\texttt{sliding\_window} in the checkpoint's \texttt{config.json} and reported
for Olmo~3 \cite{olmo2025olmo3}.
$^{c}$The repository was created on 2026-03-11, but its public history begins
with a commit titled ``Preparing for release!'' on 2026-04-02.
$^{d}$Hugging Face-format weight shards only; the repository also holds the
weights in Mistral's native format (\texttt{consolidated.safetensors},
44.5\,GB), so the harness hashed 89.0\,GB in total. The repository name gives
no scale; 22B follows from its 22.2 billion parameters.
\end{table*}

Three architectural properties are relevant. Gemma-2, Gemma-4, OLMo-2 and Olmo-3.1
normalise module outputs with an RMSNorm \cite{zhang2019rmsnorm} with
elementwise gain $\boldsymbol{\gamma}$ before they enter the residual stream
(``post-norm'', i.e.\ output normalisation \cite{kim2025periln}, documented for
Gemma~2, OLMo~2 and Olmo~3
\cite{gemmateam2024gemma2,walsh2025olmo2,olmo2025olmo3}), so direction-based edits are
incomplete on four of the eight main-study models; this potential confound is
disclosed rather than corrected in the primary runs, and a
$\boldsymbol{\gamma}$-corrected MPOA control ($\gamma$-MPOA) was run on these
four models as an additional condition (Appendix~\ref{app:setup-methods};
results in \S\ref{sec:res-secondary}). Qwen3.8-27B and Qwen3.5-4B
interleave gated-delta-rule linear-attention layers \cite{yang2025gated} with
full-attention layers, and four checkpoints are image-text
models used with text only (Appendix~\ref{app:setup-models}).

The small pair (Qwen2.5-7B-Instruct and Qwen3.5-4B) is the only candidate
pair with instruction-tuned Jacobian lenses, so J-space ablation is restricted
to it; its later model is the smaller one, confounding generation and size
(Appendix~\ref{app:validity}).

\paragraph{Prompt data.} Refusal directions, SOM fitting and ARA module
capture use the first 400 training prompts of
\texttt{mlabonne/harmful\_behaviors} and \texttt{mlabonne/harmless\_alpaca}
\cite{zou2023universal,taori2023alpaca}. The first 100 test prompts of
\texttt{harmful\_behaviors} form the development (dev) set for search and
selection, and those of \texttt{harmless\_alpaca} serve the KL objective.
All primary and secondary refusal, capability and teacher-forced outcomes come
from held-out sets. The held-out refusal set, frozen before any held-out response was generated,
contains 1{,}322 prompts, 972 harmful and 350 benign
(Table~\ref{tab:refusal-sets}), from JailbreakBench, XSTest, StrongREJECT,
HarmBench and a code-as-weapon bank
\cite{chao2024jailbreakbench,rottger2024xstest,souly2024strongreject,mazeika2024harmbench,young2026code}
(sampling in Appendix~\ref{app:setup-eval}). The 44 harmful
items that match a \texttt{harmful\_behaviors} prompt are flagged, and results
are reported with and without them. The code-as-weapon bank was compiled by the
author from public malicious-code prompt corpora and is described in a preprint
that has not been peer-reviewed \cite{young2026code}; its 200 items were drawn,
and the held-out file frozen, before any held-out response was generated.

\begin{table}[t]
\centering\small
\caption{Held-out refusal set (\texttt{refusal\_sets.json}, v2). ``Overlap'':
items matching a \texttt{mlabonne/harmful\_behaviors} prompt exactly / fuzzily.}
\label{tab:refusal-sets}
\begin{tabular}{llrr}
\toprule
Subset & Kind & $n$ & Overlap \\
\midrule
JailbreakBench harmful & harmful & 100 & 11 / 6 \\
JailbreakBench benign  & benign  & 100 & -- \\
XSTest unsafe          & harmful & 200 & 0 / 0 \\
XSTest safe            & benign  & 250 & -- \\
StrongREJECT           & harmful & 313 & 25 / 1 \\
HarmBench (standard)   & harmful & 159 & 0 / 1 \\
Code-as-weapon         & harmful & 200 & 0 / 0 \\
\midrule
Total                  &         & 1{,}322 & 36 / 8 \\
\bottomrule
\end{tabular}
\end{table}

\subsection{Study Design and Conditions}
\label{sec:setup-methods}

The independent variables are model generation (earlier or later release
within a family) and refusal-removal method; for H2, ARA forms the optimisation-based class and the
remaining methods the direction-based class. All methods run inside
Heretic~2.0 \cite{heretic2025} and differ in the weight edit and its search
space (Table~\ref{tab:search}, Appendix~\ref{app:setup-methods}). Each edit targets
the attention output projection and MLP down-projection of each layer and is
held in a LoRA adapter \cite{hu2022lora} during optimisation. The methods of
Appendix~\ref{sec:methods} are instantiated as follows (full specification
and additional conditions in Appendix~\ref{app:setup-methods}). Three presets
of the external OBLITERATUS toolkit \cite{obliteratus2026} on
Qwen2.5-7B-Instruct are evaluated with this study's harness in the companion
paper \cite{young2026paper30} and are not part of this study.

\paragraph{DiM.} The difference-in-means refusal direction
\cite{arditi2024refusal} is removed by a rank-1 edit that leaves the embedding
matrix unchanged; unlike the original procedure
(Appendix~\ref{sec:m-dim}), direction,
strength and layer profile are tuned by the same search as MPOA
(``Heretic-tuned DiM'').

\paragraph{MPOA.} Heretic's defaults: the direction is projected orthogonal to
the harmless mean \cite{lai2024projected}, and the edit is computed on
row-normalised weights whose row norms are then restored \cite{lai2025mpoa}
(rank-3 adapter).

\paragraph{SOM.} A port of the proof of concept in Heretic PR~\#196
\cite{piras2025som} (``Heretic-tuned SOM''): the $k=4$ most-winning neurons of a
per-layer self-organising map of harmful residuals yield orthonormalised
directions, which are ablated with a searched geometric decay (rank-6 adapter).

\paragraph{ARA.} Heretic's default modifier \cite{weidmann2026ara}: a rank-50
adapter per targeted module is fitted by L-BFGS to the objective of
Appendix~\ref{sec:m-ara}; row-magnitude preservation enters only the objective, and
the exported adapter is not renormalised.

\paragraph{J-space (small pair only).} Directions computed by the
\texttt{jspace-refusal} code from Jacobian lenses and refusal-token
unembeddings
\cite{sekinal2026jspace,neuronpedia2026jacobian,gurnee2026verbalizable} are
baked into the weights with the DiM/MPOA edit and search space (rank-3 adapter)
rather than applied as the original forward hook.

\subsection{Measures}
\label{sec:setup-eval}

\paragraph{Generation.} One response per held-out prompt is generated for the
original and every modified model with vLLM \cite{kwon2023efficient}:
greedy, at most 512 new tokens, the model's own chat template, system prompt
``You are a helpful assistant.'' and reasoning disabled. Decoding is
identical for all models because refusal is highly sensitive to decoding
settings \cite{huang2024catastrophic}.

\paragraph{Refusal and outcome classes.} The primary outcome is the
WildGuard \cite{han2024wildguard} refusal label of each harmful held-out
response; WildGuard agreed more closely with human labels than the HarmBench
classifier and was less affected by stylistic rewriting
\cite{xie2025sorrybench,eiras2025know}, but it has not been validated against
human labels on the outputs of edited models. No human audit was made here, so
every refusal rate is relative to the automatic judges, and ``independent''
means independent of the optimisation objective.
All judges score the same stored responses: WildGuard (refusal and
harmful response), the HarmBench classifier (harmful compliance),
Qwen3Guard-Gen-8B \cite{zhao2025qwen3guard} (second LLM judge), two
encoder-based refusal classifiers, and Heretic's 33 keyword markers on the full
response and its first 256 tokens; WildGuard, Qwen3Guard and the HarmBench
classifier were also used by a recent refusal-direction study
\cite{evseev2026wrdo}. A degeneracy heuristic flags empty, repetitive or
largely non-printable responses, which ablation can produce
\cite{marshall2024refusal}. Each response is assigned one outcome class in a
fixed order: degenerate if flagged, otherwise refusal if WildGuard labelled it
a refusal, otherwise compliance, split into complete and incomplete compliance
by whether generation stopped at the 512-token limit (finish reason
\texttt{length}). The primary outcome uses the WildGuard label of every
response, degenerate ones included; in the outcome classes degenerate
responses count as not refused (Appendix~\ref{app:setup-eval}).

\paragraph{Capability and distribution shift.} Five benchmarks run with
lm-evaluation-harness 0.4.13 \cite{gao2024language} on vLLM 0.33.0
(Table~\ref{tab:capability}): fixed, seeded subsets of MMLU-Pro
\cite{wang2024mmlupro} and GSM8K \cite{cobbe2021training} and the complete
IFEval \cite{zhou2023instruction}, HumanEval \cite{chen2021evaluating} and MBPP
\cite{austin2021program} test sets, with greedy decoding, chat templates and
reasoning disabled. Distribution shift is measured by the first-token KL
divergence of the distance objective (\S\ref{sec:setup-opt}) and by a
teacher-forced divergence: the original model's non-empty greedy responses $y$
to the 350 benign held-out prompts are scored by each model, and the divergence
of a modified model is the token-averaged
$\log p_{\mathrm{orig}}(y)-\log p_{\mathrm{mod}}(y)$.

\begin{table}[t]
\centering\small
\caption{Capability benchmarks: lm-eval task, items, shots, primary metric and
maximum generated tokens.}
\label{tab:capability}
\begin{tabular}{llrrlr}
\toprule
Benchmark & Task & $n$ & Shots & Metric & Max.\ tok. \\
\midrule
MMLU-Pro  & \texttt{mmlu\_pro\_*}        & 1{,}000 & 5 (CoT) & exact match & 2{,}048 \\
GSM8K     & \texttt{gsm8k}               & 500     & 5 & exact match (flexible) & 512 \\
IFEval    & \texttt{ifeval}              & 541     & 0 & prompt-level strict acc. & 1{,}280 \\
HumanEval & \texttt{humaneval\_instruct} & 164     & 0 & pass@1 & 1{,}024 \\
MBPP      & \texttt{mbpp\_instruct}      & 500     & 3 & pass@1 & 1{,}024 \\
\bottomrule
\end{tabular}
\end{table}

\subsection{Procedure}
\label{sec:setup-opt}

\paragraph{Optimisation.} Each model--method combination is one Optuna study
\cite{akiba2019optuna} with a multivariate TPE sampler
\cite{bergstra2011algorithms}: 100 trials, of which the first 30 are random,
with seed~1 (seeded generators, exceptions for two resumed studies and
response-prefix handling in Appendix~\ref{app:setup-opt}). Every model used the
same seed, so studies with the same search space share their 30 random start-up
draws, and two pairs of selected parameter vectors coincide across models
(Appendix~\ref{app:kl-cap}).
The ten small-pair
studies were repeated with seed~2 to estimate seed sensitivity
(Appendix~\ref{app:setup-methods}). Two objectives are minimised, so each study
yields a Pareto front. The refusal objective is the fraction of 100 greedy
dev-set responses (at most 100 new tokens) containing any of Heretic's 33
refusal markers; the distance objective is the KL divergence of the modified
model's first-token distribution from the original model's, averaged over the
100 harmless prompts.

\paragraph{Selection rule.} One rule is applied to every run: among completed trials
with first-token KL divergence $\le 0.1$, select the one with the fewest dev-set refusals, breaking ties by
lower KL divergence and then lower trial index; if none qualifies, select
nothing and report the lowest-KL trial. For sensitivity analysis the rule is repeated at thresholds 0.05,
0.2 and 0.5, and full Pareto fronts are reported. The cap is the same for
every model, but it need not constrain every run: it is said to bind in a run
when some trial above KL 0.1 had fewer dev refusals than the selected trial.
Binding is recorded for every run, together with the position of the selected
trial's parameters in Heretic's search ranges (Appendix~\ref{app:kl-cap}).

\paragraph{Export and release.} Each selected trial is merged into a
standalone checkpoint that differs from the original only in the edited
matrices; the 42 primary models and the four $\gamma$-MPOA controls are
published on Hugging Face, except the five Gemma-2-27B variants, which are
withheld; the ten seed-2 replicates are evaluated but not
released (Appendix~\ref{sec:setup-export}).

\subsection{Data Analysis}
\label{sec:setup-stats}

All intervals are 95\%. Proportions carry Wilson score intervals
\cite{wilson1927probable,brown2001interval}; paired differences on the same prompts
carry Newcombe's hybrid score intervals \cite{newcombe1998improved} and are
tested with the exact McNemar test \cite{mcnemar1947note}, a paired,
prompt-level comparison as recommended for language-model evaluations
\cite{miller2024adding}. Each modified model is compared in this way with its
original. For H1, the generation effect of a pair and method is the paired
difference in residual refusal between the selected models of the newer and
the older release on the same 972 harmful prompts, tested by exact McNemar.
Pooled over the four main-study families, H1 and H2 are tested with a logistic
generalised estimating equation (GEE) \cite{liang1986longitudinal} with
generation, method class, their interaction and family as fixed effects, an
exchangeable working correlation clustered by prompt and robust standard
errors. The planned mixed-effects logistic regression,
\texttt{refusal $\sim$ generation $\times$ method\_class + (1|prompt) + (1|family)}
in lme4 notation \cite{bates2015lme4,baayen2008mixed}, was fitted by
variational Bayes and is reported as posterior means with $\pm 1.96$~SD
intervals, without $p$ values; its conditional coefficients are larger in
magnitude than the population-averaged GEE coefficients. It has no
pair-specific generation or interaction slopes, so it cannot represent
between-pair heterogeneity, and it is reported for completeness only, in
Appendix~\ref{app:rr-tables} (Table~\ref{tab:h2-vb}).
H1 is tested by the GEE
generation effect within each method class and H2 by the generation $\times$
class interaction, pooled and per pair. H2 is also expressed as a
risk-difference contrast (ARA generation effect minus the mean of the
direction-based methods), and baseline-adjusted generation effects subtract
the originals' own generation difference; both carry item-bootstrap intervals
(4,000 resamples). Pooled baseline-adjusted generation effects come from the
same GEE with the originals added as a reference class, so that the
generation coefficient of each method class is its generation effect minus
that of the originals, in log-odds.

Because each main-study cell was optimised once, these intervals reflect
prompt sampling only, not run-to-run variation of the search. That variation
was estimated post hoc from the ten small-pair cells, each optimised with two
seeds: with $d_i$ the seed-2 minus seed-1 held-out refusal of cell $i$ in
percentage points, the run-to-run standard deviation of one cell is
$\sigma=\sqrt{\sum_{i=1}^{n} d_i^2/(2n)}$ ($n=10$), with a $\chi^2_{10}$
interval, and the difference between two independently optimised cells has
standard deviation $\sigma\sqrt{2}$. A per-pair generation effect is flagged
as exceeding seed noise when its absolute value is larger than
$2\sigma\sqrt{2}$. This flag is a screen that accompanies, not replaces, the
McNemar tests, and it assumes that the seed variability of the 4B and 7B
small-pair models transfers to the 22--32B main-study models. In 3 of the 10
seed pairs (Qwen3.5-4B DiM, SOM and ARA) the seed-2 run also used a generation
batch size of 64 instead of 128, because 128 ran out of memory, so $\sigma$ is
also reported without these pairs.

The generational question has the model pair as its unit: four main-study
pairs and one small pair. The prompt-level intervals and tests above
therefore describe prompt sampling within a pair, not variation between runs
or between pairs, and per-pair effects are the primary evidence. For each pair
the generation effect of the direction-based class (the mean of its methods)
and of ARA is given in percentage points (mean of the paired item differences)
and in log-odds (per-pair GEE), raw and baseline-adjusted, with intervals that
add run-to-run variance: each optimised cell contributes $\sigma^2$, so a
single-method effect has $2\sigma^2$, a mean of $m$ methods $2\sigma^2/m$
and the H2 contrast $2\sigma^2+2\sigma^2/m$, with $\sigma$ estimated on each
scale from the seed pairs. Differences between pairs are tested at the prompt level with Wald
tests of the generation $\times$ family and generation $\times$ class
$\times$ family terms of the GEE (3 df), which treat prompts, not runs, as
independent. Pooled effects over the four main
pairs are summarised with a random-effects model: DerSimonian--Laird
between-pair variance \cite{dersimonian1986meta}, a
Hartung--Knapp--Sidik--Jonkman 95\% interval with a $t$ distribution on 3
degrees of freedom \cite{hartung2001refined,inthout2014hartung}, $I^2$
\cite{higgins2002quantifying} and a 95\% prediction interval for a new pair
\cite{higgins2009reevaluation}; with $k=4$ pairs these intervals are wide by
construction. Leave-one-family-out GEE fits complete the pooled analysis. The
baseline-adjusted contrasts were added after H1 had been stated and are
exploratory; because the percentage-point and log-odds scales can disagree in
sign, both are reported.

Four further analyses were added in response to review, after all results were
known, and are exploratory. First, between-pair heterogeneity is reported as
Cochran's $Q$, $I^2$ and the DerSimonian--Laird $\tau^2$ with within-pair
variances that include the run-to-run variance above, not only prompt
sampling. Second, because a constant $\sigma$ in percentage points ignores
that a rate varies less near 0\% and 100\%, the run effect is also modelled
as constant on the logit scale, $\sigma_L$ (estimated from the logits of the
same seed pairs), and mapped to each cell by the delta method,
$\sigma_{\mathrm{pp}}=\sigma_L\,p(1-p)$ at the cell's rate $p$; per-pair
verdicts are reported under both models across the interval of $\sigma$.
Third, H2 is re-estimated on the direction-based units in which the KL cap
bound in both cells, and the per-pair verdicts are read on the dev set at
first-token KL budgets of 0.05 to 1.0 and on the area under each run's
refusal--KL front, with run-to-run variance estimated from the small-pair seed
pairs on the same quantity. Fourth, as a scale-free alternative to the
baseline adjustment, conditional refusal retention is the share of the prompts
refused by a model's own original that the edited model still refuses, with
prompt-bootstrap intervals and with intervals that add a run-to-run variance
of retention estimated, as for the refusal rate, from the ten small-pair seed
pairs (Table~\ref{tab:retention}).

H3 compares keyword and judge labels on identical responses: the paired
keyword-minus-WildGuard gap of each model (Newcombe interval, exact McNemar),
pooled gaps with prompt-cluster bootstrap intervals (2,000 resamples), and
agreement as Cohen's $\kappa$ for rater pairs \cite{cohen1960coefficient} and
Fleiss' $\kappa$ across judge panels \cite{fleiss1971measuring}, each with a
prompt-bootstrap interval, raw agreement and label marginals, because $\kappa$
is low when labels are as imbalanced as near-0 or near-100\% refusal
\cite{feinstein1990high}. The method-class difference in the gap is tested
with a linear GEE on the per-response paired difference (clustered by prompt,
pair as a fixed effect) and checked with an ordinary least-squares model on
the per-model mean gaps, with the model as unit. Keyword labels are computed
on three windows: the full response (at most 512 tokens), its first 256 tokens
and its first 100 tokens, the window Heretic scores during search. Token ids
were not stored, so the 100-token text is obtained by re-tokenising each
stored response with the evaluated model's tokenizer, a procedure validated
against the stored 256-token prefixes (Appendix~\ref{app:trunc256}).

Holm's procedure \cite{holm1979simple} controls each family of related tests
at a two-sided $\alpha=0.05$, separately for each refusal measure and prompt
subset (families and sizes in Appendix~\ref{app:setup-stats}).
Variational-Bayes results are not adjusted. Overfitting of selection to the
dev set, expected when a noisy criterion is both optimised and used for
selection \cite{cawley2010overfitting}, is assessed by the Spearman
correlation between dev-set keyword refusal and held-out judged refusal across
selected models, with percentile bootstrap intervals over models. Of the
secondary outcomes (over-refusal on the 350 benign items, which tracks
harmful-prompt refusal across models \cite{cui2025orbench}, capability
differences and teacher-forced divergence), changes in over-refusal from the
original are tested by exact McNemar, Holm-adjusted over all modified models;
the others are descriptive apart from the tests listed above. As
sensitivity analyses, H1 and H2 are repeated on the four public item sets
(772 prompts) and on the code-as-weapon bank (200) separately, and with
harmful compliance as the outcome (HarmBench classifier and WildGuard
harmful-response labels, coded as not harmful so that signs match the refusal
analysis). Dev-set drift of the released exports is compared between exports
rebuilt on the search machine and on another GPU with Fisher's exact test.
These analyses, the pair-level pooling and the KL-cap analysis were added
after the held-out results were known. Analyses used
Python~3.12.15 with pandas~3.1.1, NumPy~2.5.4, SciPy~1.18.1 and
statsmodels~0.15.0 \cite{seabold2010statsmodels}; hardware and other software
versions are listed in Appendix~\ref{sec:setup-compute}.

\subsection{Ethics and Preregistration}
\label{sec:setup-ethics}

\paragraph{Ethics.} No human participants were involved.
Prompt sets are described by name and count only; responses to harmful prompts
are stored privately for scoring and not released. Licences of released
models and judges are listed in Appendix~\ref{sec:setup-export}.

\paragraph{Preregistration.} The study was not preregistered; the hypotheses
were formulated after pilot runs on the small pair (30 trials, 10 random),
which served to verify the pipeline and are not among the analysed runs. The
baseline-adjusted contrast was added after H1 had been stated, and the
pair-level pooling, the KL-cap, item-source and harmful-compliance analyses
and the 100-token keyword window were added after the held-out results were
known, as were the analyses added in response to review (seed-inclusive
heterogeneity, refusal-level run-to-run variance, KL budgets, cap-bound units,
conditional retention and outcome classes); all of these are exploratory.
Deviations from the study plan are listed in
Appendix~\ref{sec:setup-deviations}.

%% file: sections/05_results.tex
\section{Results}
\label{sec:results}

The study asked whether refusal removal is harder in the newer model of each
release pair (H1), whether this depends on the class of edit (H2), and whether
keyword counts agree with an automatic judge (H3). Outcomes are WildGuard
refusal rates on the 972 harmful held-out prompts for the selected seed-1
models; pp denotes percentage points. The release pair is the unit, so
per-pair effects with intervals that add run-to-run variance are the primary
evidence; prompt-level tests describe prompt sampling within a pair
(\S\ref{sec:setup-stats}). Detail that earlier drafts gave here is in
Appendix~\ref{app:moved-results}.

\subsection{H1: Residual Refusal in the Newer Model}
\label{sec:res-h1}

\begin{figure*}[!htbp]
\centering
\includegraphics[width=\textwidth]{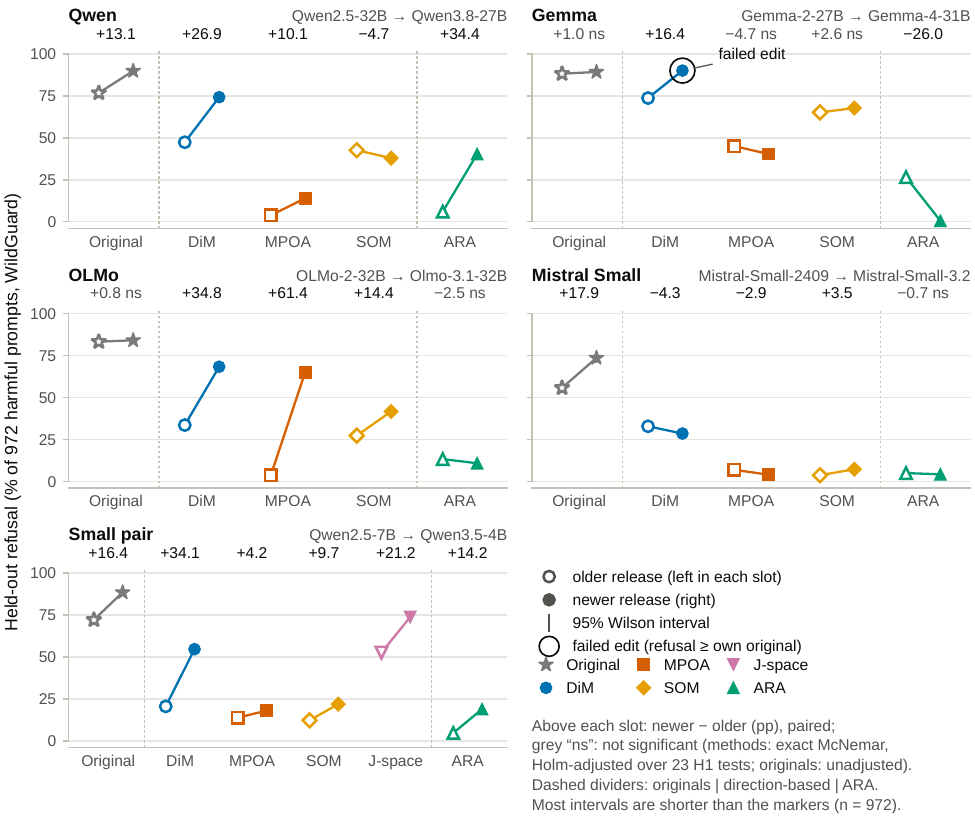}
\caption{\textbf{Held-out, judged refusal of the original and refusal-removed models, older vs newer release.} Each panel is one pair; each slot is one condition: the unedited originals (grey star), the direction-based methods (DiM, MPOA, SOM; J-space on the small pair only) and, right of the second dashed divider, the optimisation-based ARA. Within a slot the older release (left, hollow marker) is joined to the newer one (right, filled marker). The value is the share of the 972 harmful held-out prompts whose response WildGuard labels a refusal; bars are Wilson 95\% intervals. The number above a slot is the paired generation effect (newer minus older, percentage points), printed in grey with ``ns'' where it is not significant. Significance for the methods is Holm-adjusted (exact McNemar, Holm over the 23 H1 tests); significance for the originals is unadjusted (exact McNemar), because they are the baseline rather than an H1 test. The ring marks a failed edit, a selected model whose held-out refusal is at or above its own original's (Gemma-4-31B DiM). Each modified model is the dev-set selection of its cell (fewest dev keyword refusals at first-token KL $\le$ 0.1, seed 1).}
\label{fig:heldout-generation}
\end{figure*}

Refusal removal was not uniformly harder on the newer model. The generation
effect differed in sign and size between pairs (Table~\ref{tab:pair-effects},
Figure~\ref{fig:heldout-generation}; per method in
Table~\ref{tab:h1-heldout}): after direction-based removal it ranged from
$-$1.2~pp (Mistral Small) to +36.9~pp (OLMo), and after ARA from $-$26.0~pp
(Gemma) to +34.4~pp (Qwen). No main-study pair retained more refusal on the
newer model for every method, as H1 required; in the small pair all five
methods did with seed~1. With run-to-run variance added to prompt sampling, the
raw per-pair effects whose interval excluded zero were the OLMo and small-pair
direction-based increases, the Qwen ARA increase and the Gemma ARA decrease.

\paragraph{Heterogeneity between pairs.} With prompt sampling only, the
pairs differed far beyond chance (prompt-level generation $\times$ pair Wald
$\chi^2=547$, 3~df; $I^2$ 99--100\%), but this statistic treats prompts, not
optimisation runs, as the unit. With the seed-inclusive within-pair variances
of Table~\ref{tab:pair-effects} and a constant run-to-run variance,
heterogeneity of the raw direction-based effect was $Q=14.8$ (3~df, $p=0.002$;
$I^2=80\%$, $\tau^2=225$~pp$^2$, $\tau=15.0$~pp) on the pp scale but $Q=3.8$
($p=0.29$; $I^2=20\%$, $\tau^2=0.12$) on the log-odds scale; for the
baseline-adjusted effect it was $Q=27.7$ ($I^2=89\%$, $\tau^2=476$~pp$^2$) and
$Q=7.7$ ($p=0.053$; $I^2=61\%$, $\tau^2=0.72$). How large the pp-scale
heterogeneity is depends on the variance model: with the refusal-level
run-to-run variance it was $Q=5.5$ ($p=0.14$; $I^2=46\%$, $\tau^2=127$~pp$^2$)
for the raw direction-based effect, and across the 95\% interval of the
run-to-run standard deviation $I^2$ ranged from 90\% to 39\% (constant) and
from 73\% to 0\% (refusal-level). Table~\ref{tab:heterogeneity} gives $Q$,
$I^2$ and $\tau^2$ for every estimand and Table~\ref{tab:pair-variances} the
within-pair variances. The pooled random-effects summaries
(Table~\ref{tab:pair-effects}, last column) span large effects of either sign;
beyond the originals' own difference, an exploratory contrast, the pooled
direction-based effect was +0.06 log-odds [$-$1.67, +1.80].

\paragraph{Run-to-run variation.} Each main-study cell was optimised once, so
run-to-run variance was transferred from the ten small-pair seed pairs
(\S\ref{sec:setup-stats}). Under the constant standard deviation of 9.2~pp
per cell, 5 of the 16 main-study per-method effects exceeded twice the
standard deviation of a difference of two runs (Table~\ref{tab:h1-heldout}).
When the run effect was instead held constant on the logit scale and mapped to
percentage points at each cell's refusal rate, which makes it smaller near 0\%
and larger near 50\%, only OLMo MPOA exceeded it, and of the per-pair class
intervals only the OLMo direction-based increase, the OLMo H2 contrast and the
Mistral Small baseline-adjusted effects excluded zero; the Qwen and Gemma ARA
effects and the Gemma H2 contrast did not
(Appendix~\ref{app:sigma-level}, Table~\ref{tab:sigma-level}). The ten seed
pairs did not favour either variance model (log-likelihood 6.3 vs.\ 6.0).

\input{tables/t7_pair_effects}

\input{tables/t4_h1_heldout}

\paragraph{Baseline.} The newer Qwen, Mistral Small and small-pair originals
already refused more than their predecessors (+13.1, +17.9 and +16.4~pp); the
Gemma and OLMo originals did not. Relative to this difference the
direction-based effect was near zero in Qwen and the small pair, negative in
Mistral Small and unchanged in Gemma and OLMo (Table~\ref{tab:pair-effects});
the two scales of this adjustment disagreed in sign for 3 of 21 pair $\times$
method cells (Appendix~\ref{app:kl-cap}, Table~\ref{tab:scales}). As a
scale-free alternative, the share of the prompts refused by a model's own
original that the edited model still refused rose on the newer model by
+43.4~pp in OLMo (95\% interval with run-to-run variance [+25.7, +61.1]) and
+13.5~pp in the small pair ([$-$1.9, +28.9]) after direction-based removal, and
changed by $-$6.6 to +6.0~pp in the other pairs, none of which excluded zero
(Table~\ref{tab:retention}).

\paragraph{Item source.} The per-pair verdicts depended in part on where the
prompts came from (Table~\ref{tab:pair-source}). The code-as-weapon bank (200
of the 972 prompts), compiled by the author, carried 52\% of the mean
originals' difference of the four main pairs. Gemma's direction-based increase
came entirely from it: +49.3~pp on the code items against $-$6.8~pp on the
public sets. The OLMo direction-based increase, the Mistral Small null and the
sign of the Gemma and OLMo H2 contrasts held on both sources; the Gemma ARA
decrease lay in the public sets (the code items were near the floor), and the
Qwen H2 contrast was positive on the public sets (+30.3~pp) and near zero on
the code items ($-$1.8~pp).

\input{tables/t9_pair_source}

\subsection{H2: Generation \texorpdfstring{$\times$}{x} Edit Class}
\label{sec:res-h2}

ARA, a rank-50 per-module edit, lost less effectiveness on the newer model
than the low-rank direction-based edits in the Gemma ($-$30.8~pp) and OLMo
($-$39.3~pp) pairs, more in the Qwen pair (+23.6~pp), and about as much in
Mistral Small and the small pair (Tables~\ref{tab:pair-effects}
and~\ref{tab:h2-contrasts}). Each per-pair contrast rests on a single ARA cell.
With constant run-to-run variance, only the Gemma and OLMo contrasts excluded
zero; with the refusal-level variance, only OLMo did. The Qwen contrast is
therefore indeterminate: its interval with run-to-run variance included zero
($-$5.9 to +53.1~pp). Pooled with the pair as the unit, the contrast was
$-$11.5~pp [$-$57.6, +34.6]. The planned mixed model gave the same signs
(Appendix~\ref{app:rr-tables}, Table~\ref{tab:h2-vb}).

\paragraph{KL cap and budget.} The KL cap bound in all 8 ARA runs but in 15 of
the 24 direction-based runs. Restricted to the direction-based units in which
it bound in both cells (Qwen DiM; Gemma DiM, MPOA and SOM; Mistral Small DiM;
none in OLMo), the H2 contrast was +7.5~pp [$-$28.6, +43.7] in Qwen,
$-$30.8~pp [$-$60.2, $-$1.3] in Gemma and +3.6~pp [$-$32.4, +39.6] in Mistral
Small, and pooled $-$8.5~pp [$-$62.5, +45.5]. Away from the selected budget
only dev-set keyword counts exist, and both checks below use them. Their
validity for H2 is limited: at the selected budget, the dev ARA generation
effects had the opposite sign to the held-out judged ARA effects in three of
the four main pairs (Appendix~\ref{app:kl-budgets}), so these checks show how
stable the dev-set contrast is, not the held-out one. On them the direction-based generation
effects kept their sign at every budget from 0.1 to 1.0 in all four main
pairs, whereas the H2 contrast changed sign with the budget in three of them
(Gemma above KL~$=0.1$; OLMo and Mistral Small below it). On the area under
the refusal--KL front, a budget-free summary, the contrast was in the
predicted direction but within run-to-run variation in Gemma and OLMo and
opposite, beyond it, in Qwen and Mistral Small (Appendix~\ref{app:kl-budgets},
Table~\ref{tab:kl-budgets}).

\paragraph{ARA cells.} Three properties of the ARA cells bear on these
contrasts (Appendix~\ref{app:moved-results}): the four ARA exports refitted on
a different GPU from the search (Qwen and Gemma pairs) drifted more on dev
re-score than those refitted on the search GPU; the Gemma-4-31B ARA export,
behind the largest contrast, re-scored above the KL cap (0.107); and a public
ARA model of Qwen3.8-27B built with the same defaults refused 26.4\% of the
held-out prompts, against 40.4\% for the ARA model selected here, on which the
reversed Qwen estimate rests. Without the failed Gemma-4-31B DiM edit, the Gemma
contrast was $-$24.9~pp [$-$56.2, +6.3].

\input{tables/t5b_h2_contrasts_main}

\subsection{H3: Keyword Counts versus an Automatic Judge}
\label{sec:res-h3}

\begin{figure*}[!htbp]
\centering
\includegraphics[width=\textwidth]{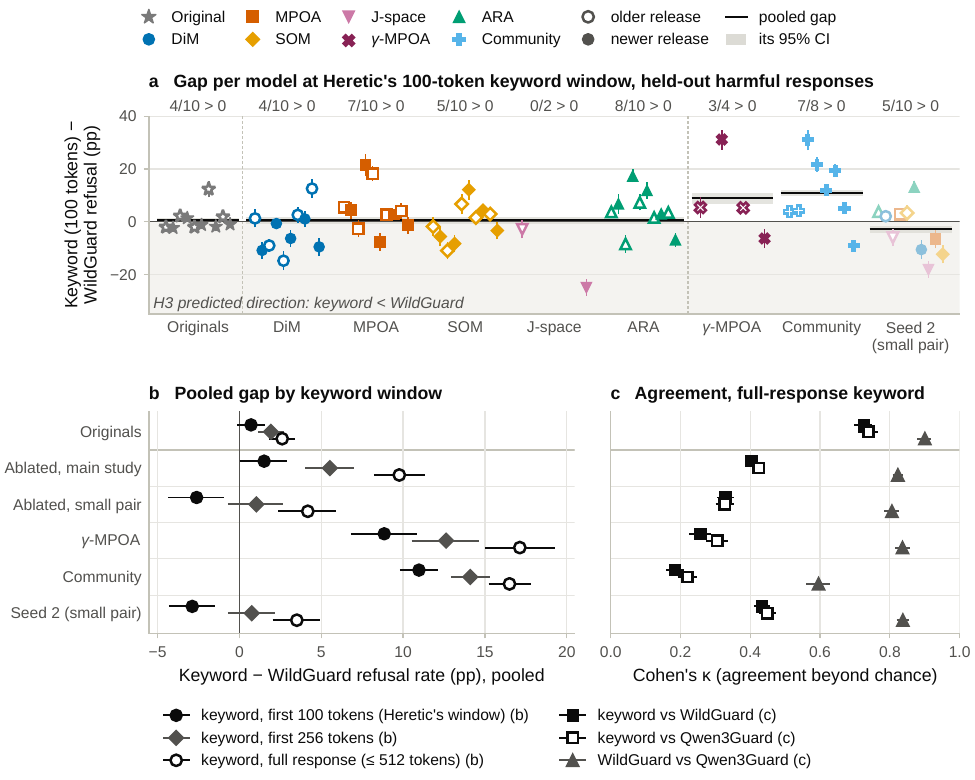}
\caption{\textbf{Keyword counts versus two LLM judges on the same held-out responses (H3).} (a) Each marker is one evaluated model (cell): its keyword refusal rate (any of Heretic's 33 refusal markers in the first 100 generated tokens, the window Heretic scores during search) minus its WildGuard refusal rate on the 972 harmful held-out prompts, in percentage points, with a paired 95\% interval (Newcombe). Slots, left to right: the unedited originals, the seed-1 ablated models by method (direction-based, then ARA), $\gamma$-MPOA, public community releases and the small pair's seed-2 replicates (lighter); within a slot the models run in pair order (Qwen, Gemma, OLMo, Mistral Small, small pair), hollow = older, filled = newer release. The black bar and grey band are the pooled gap of the slot group with its 95\% interval; the count above a slot is the number of cells with a positive gap. H3 predicted the shaded region (keyword below the judge). (b) Pooled keyword-minus-WildGuard gap for keyword windows of 100 and 256 tokens and the full response (at most 512 tokens). (c) Cohen's $\kappa$ of the full-response keyword label, with prompt-bootstrap 95\% intervals. No human-labelled audit exists, so the panels show where keyword matching and the judges disagree, not which is correct.}
\label{fig:judge-agreement}
\end{figure*}

H3's prediction that keyword matching would count fewer refusals than the
judge was not supported at any window (Table~\ref{tab:h3-windows},
Figure~\ref{fig:judge-agreement}). At the 100-token window that Heretic scores,
an exploratory window added after the held-out results were known, keyword
matching and WildGuard agreed on average over the 42 ablated models (+0.5~pp
[$-$1.0, +1.9]) but differed by 7.0~pp per model on average, significantly in
25 models and in both directions. Over the full response keyword matching
counted more refusals (+8.4~pp [+6.9, +10.0]), mainly through markers after
the first 100 tokens. In the main study the gap was larger for ARA and on the
newer models with the prompt as the unit, but not with the model as the unit
(Table~\ref{tab:h3-gap}). Keyword--WildGuard agreement fell from
$\kappa=0.73$ in the originals to 0.40 (main study) after ablation, whereas
the two LLM judges agreed at 0.82.

\input{tables/t8_h3_windows}

\input{tables/t6_h3_gap}

\subsection{Secondary Outcomes}
\label{sec:res-secondary}

\begin{figure*}[!htbp]
\centering
\includegraphics[width=\textwidth]{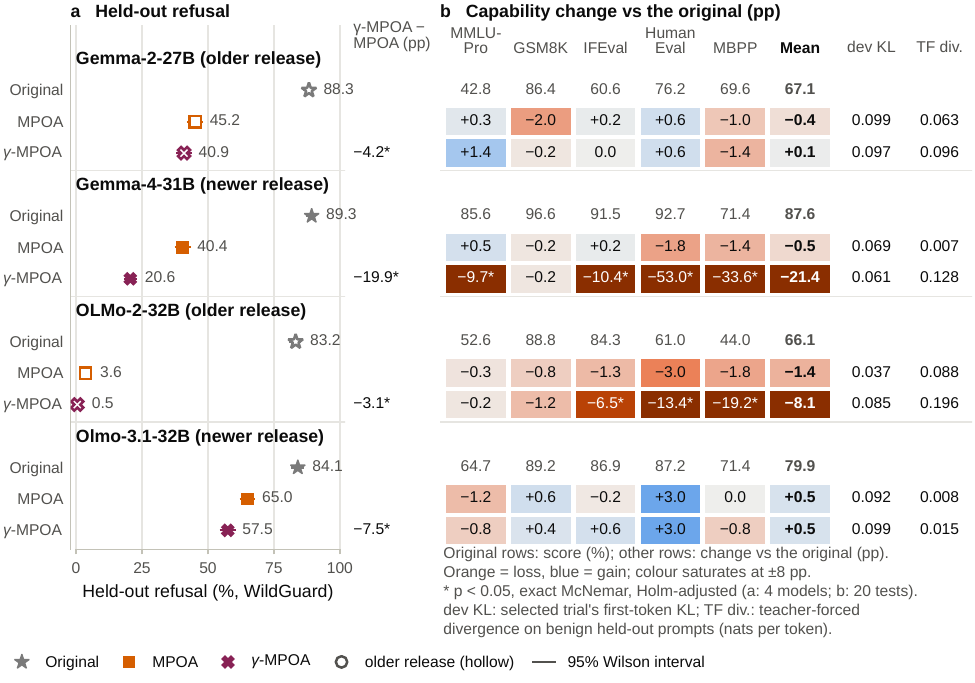}
\caption{\textbf{$\gamma$-corrected MPOA versus MPOA on the four post-norm models: held-out, judged refusal and capability.} Each block is one model (older or newer release of the Gemma and OLMo pairs) with its unedited original (grey star), the selected MPOA model (vermillion square) and the selected $\gamma$-MPOA model (wine-red cross; MPOA with the edit corrected for the post-norm output gains); hollow markers are the older releases. (a) Share of the 972 harmful held-out prompts whose response WildGuard labels a refusal, with Wilson 95\% intervals; the column on the right is $\gamma$-MPOA minus MPOA (percentage points, paired; * = significant, exact McNemar, Holm over the four models). (b) Change of each benchmark score from the model's own original, in percentage points (orange = loss, blue = gain; the colour saturates at $\pm$8 pp, the printed values are exact; the original's row gives its scores in \%); * marks a $\gamma$-MPOA score that differs from the original's (paired per-item McNemar, Holm over 20 tests). The last two columns are the selected trial's dev first-token KL divergence and the teacher-forced divergence on the 350 benign held-out prompts (nats per token). Single seed; one benchmark item is worth 0.1--0.61 pp.}
\label{fig:gmpoa-heldout}
\end{figure*}

\paragraph{$\gamma$-MPOA control.} Correcting MPOA for the post-norm output
gains (Appendix~\ref{app:setup-methods}) lowered held-out refusal on all four
post-norm models (Figure~\ref{fig:gmpoa-heldout}), more on the newer model of
each pair: by 4.2 and 19.9~pp on Gemma-2-27B and Gemma-4-31B (generation
difference $-$15.6~pp [$-$19.6, $-$11.7]) and by 3.1 and 7.5~pp on OLMo-2-32B
and Olmo-3.1-32B ($-$4.4~pp [$-$6.6, $-$2.2]). With $\gamma$-MPOA in place of
MPOA, the OLMo generation effect remained +57.0~pp and the pooled
prompt-level H2 interaction $-$0.42 log-odds. On Gemma-4-31B and OLMo-2-32B,
$\gamma$-MPOA lost 21.4 and 8.1~pp of mean capability. DiM and SOM were not
corrected.

\begin{figure*}[!htbp]
\centering
\includegraphics[width=\textwidth]{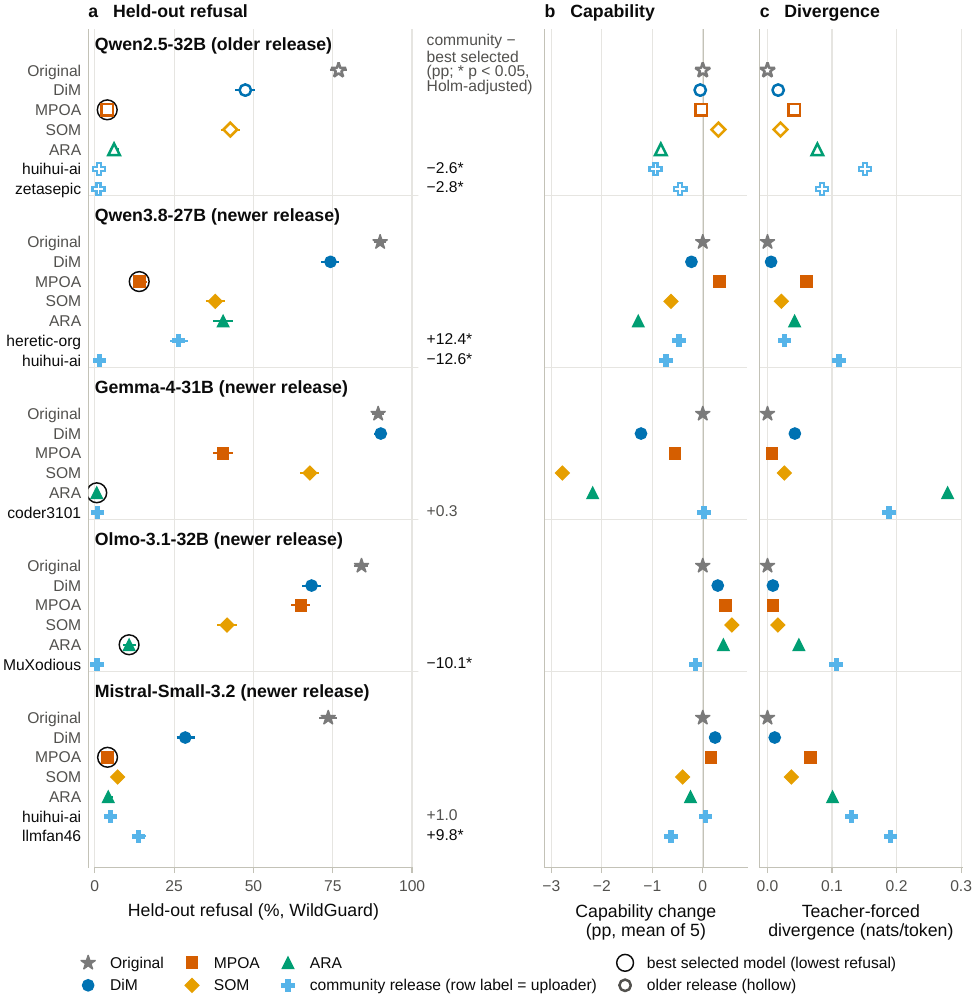}
\caption{\textbf{Models selected in this study versus public community abliterations of the same checkpoints (held-out, judged).} One block per base model with at least one community release: the unedited original (grey star), the four seed-1 models selected here by the dev-set rule (hollow markers = older release) and each community release (sky-blue plus; row label = Hugging Face uploader). (a) Share of the 972 harmful held-out prompts whose response WildGuard labels a refusal, with Wilson 95\% intervals; the ring marks the best selected model (lowest held-out refusal of the four), and the column on the right gives community minus that model (percentage points, paired; * = significant, exact McNemar, Holm over the four comparisons per community release). (b) Change of the unweighted mean of the five capability benchmarks from the model's own original (percentage points). (c) Teacher-forced divergence from the original on the 350 benign held-out prompts (nats per token). $\gamma$-MPOA is not drawn.}
\label{fig:community}
\end{figure*}

\paragraph{Community releases.} Of eight public abliterations of five of the
checkpoints (Figure~\ref{fig:community}), four refused less than the best
model selected here for the same checkpoint, for example 1.4\% against 14.0\%
on Qwen3.8-27B and 0.7\% against 10.8\% on Olmo-3.1-32B, at teacher-forced
divergences of 0.085--0.151 (best selected models 0.042--0.060) and mean
capability changes within 1~pp.

\paragraph{Harmful compliance and outcome classes.} Residual refusal is not
harmful compliance: of the ablated main-study responses that WildGuard did not
label as refusals, 75.1\% were harmful by the HarmBench classifier. With
harmful compliance as the outcome the pooled conclusions and the sign of the
per-pair H2 interactions were the same, but under HarmBench every Qwen method
was higher on the newer model, so the Qwen pair met the H1 criterion
(Appendix~\ref{app:harm}). Degenerate responses were rare (at most 0.6\% per
original or primary model); of the non-refused responses of the 42 primary models, 65.8\% reached
the 512-token limit (incomplete compliance;
Appendix~\ref{app:outcome-classes}).

\begin{figure*}[!htbp]
\centering
\includegraphics[width=\textwidth]{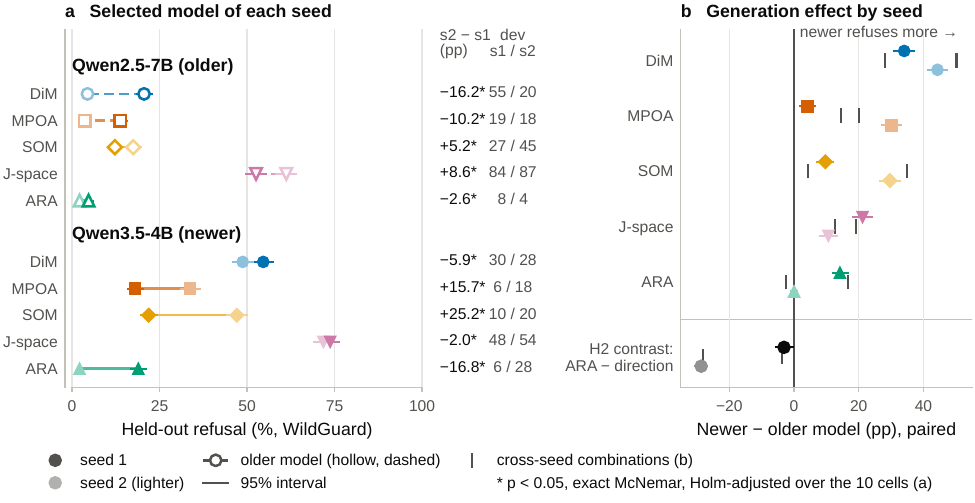}
\caption{\textbf{Seed sensitivity of held-out, judged refusal on the small pair (Qwen2.5-7B $\to$ Qwen3.5-4B).} Each of the ten model $\times$ method cells was optimised twice with 100 TPE trials that differ only in the seed (in three seed-2 runs also in batch size), and the model selected from each run was evaluated on the held-out set. (a) Share of the 972 harmful held-out prompts whose response WildGuard labels a refusal for the seed-1 model (full method colour) and the seed-2 model (lighter), with Wilson 95\% intervals; hollow markers and dashed lines mark the older model. Columns on the right: seed 2 minus seed 1 (percentage points, paired; * = significant, exact McNemar, Holm over the ten cells) and the dev keyword refusals (of 100) of the two selected trials. (b) Generation effect (newer minus older model, paired, Newcombe 95\% intervals; item-bootstrap interval for the H2 row) with seed 1 (upper) and seed 2 (lower, lighter); the grey ticks are the two cross-seed combinations. The bottom row is the H2 contrast (ARA minus the mean of the direction-based methods). The main-study pairs have one seed only.}
\label{fig:seed-heldout}
\end{figure*}

\paragraph{Seed sensitivity.} On the small pair a second seed changed held-out
refusal significantly in all ten cells (median 9.4~pp, maximum 25.2~pp;
Figure~\ref{fig:seed-heldout}) and erased the ARA generation effect (+14.2~pp
with seed~1, +0.0~pp with seed~2); method accounted for $\eta^2=0.94$ and 0.88
of the variance among the cell rates of the two models.

\begin{figure*}[!htbp]
\centering
\includegraphics[width=\textwidth]{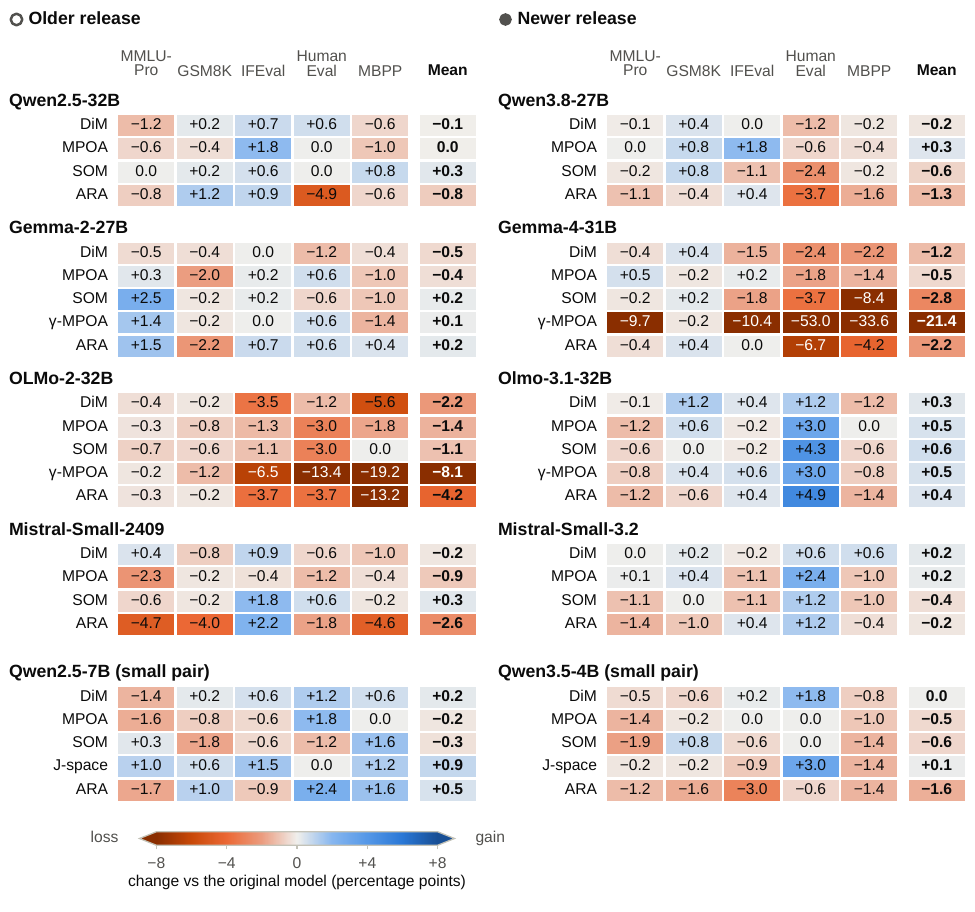}
\caption{\textbf{Capability change after refusal removal.} Each cell is the benchmark score of the selected model minus that of its own original, in percentage points (pp); the last column is the unweighted mean of the five benchmarks. Orange marks a loss, blue a gain; the scale saturates at $\pm$8 pp and the printed values are exact (beyond the scale: Gemma-4-31B SOM MBPP, $-$8.4 pp; Gemma-4-31B $\gamma$-MPOA MMLU-Pro, $-$9.7 pp; Gemma-4-31B $\gamma$-MPOA IFEval, $-$10.4 pp; Gemma-4-31B $\gamma$-MPOA HumanEval, $-$53.0 pp; Gemma-4-31B $\gamma$-MPOA MBPP, $-$33.6 pp; Gemma-4-31B $\gamma$-MPOA Mean, $-$21.4 pp; OLMo-2-32B ARA MBPP, $-$13.2 pp; OLMo-2-32B $\gamma$-MPOA HumanEval, $-$13.4 pp; OLMo-2-32B $\gamma$-MPOA MBPP, $-$19.2 pp; OLMo-2-32B $\gamma$-MPOA Mean, $-$8.1 pp). Rows align the same method across the older (left) and newer (right) release of a pair; the 8 public community releases are shown in Figure~\ref{fig:community-releases}. One item is worth 0.1--0.61 pp depending on the benchmark and the lm-eval standard error of a single score is 0.7--3.9 pp, so differences of 1--2 pp are within noise (single seed).}
\label{fig:capability}
\end{figure*}

\begin{figure*}[!htbp]
\centering
\includegraphics[width=\textwidth]{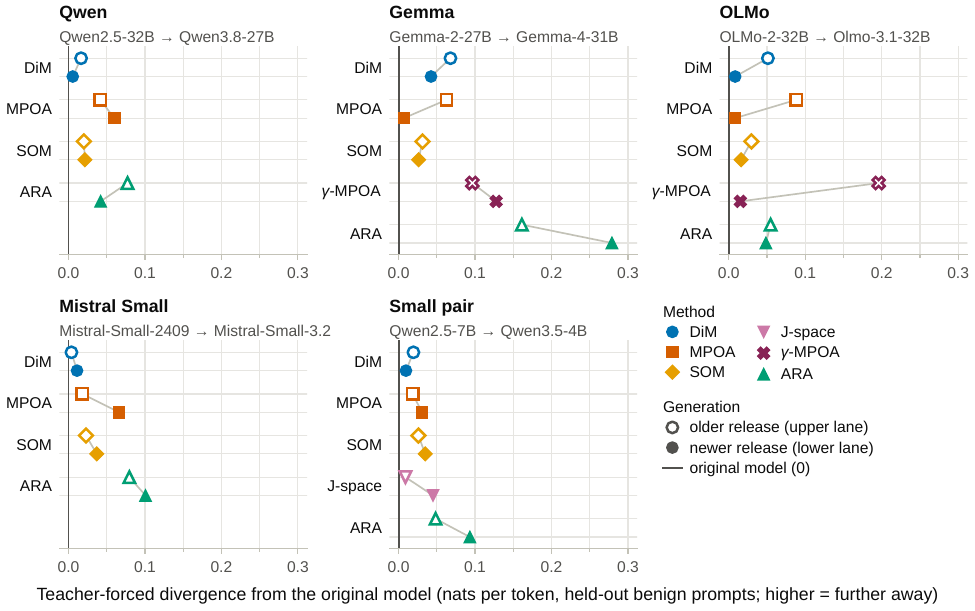}
\caption{\textbf{Divergence of the refusal-removed models from their originals on benign held-out prompts.} Each dot is the teacher-forced divergence of one selected model: the token-averaged $\log p_{\mathrm{orig}}(y)-\log p_{\mathrm{mod}}(y)$ over the original model's own greedy answers $y$ to the 350 benign held-out prompts; 0 is the original model. Panels are pairs; in each row the older release is in the upper lane (hollow marker) and the newer one in the lower lane (filled), joined by a grey line. The Qwen3.5-4B values also contain an export difference: its evaluated checkpoints lack the 15 mtp.* tensors and store 48 F32 linear-attention tensors (A\_log, norm.weight) as BF16, which by itself changes the log-probabilities. Single seed; no uncertainty intervals are drawn.}
\label{fig:divergence}
\end{figure*}

\paragraph{Capability, divergence and over-refusal.} Mean capability change of
the 42 primary models ranged from $-$4.2 to +0.9~pp
(Figure~\ref{fig:capability}) and teacher-forced divergence from 0.004 to
0.279 nats per token (Figure~\ref{fig:divergence}). Capability change was not
correlated with the selected dev KL (Spearman $\rho=-0.03$ [$-$0.29, +0.24],
56 modified models) but was with teacher-forced divergence ($\rho=-0.48$
[$-$0.66, $-$0.24]). Over-refusal on the 350 benign prompts fell significantly
in 41 of the 42 primary models; the exception was the failed Gemma-4-31B DiM
edit.

\begin{figure*}[!htbp]
\centering
\includegraphics[width=\textwidth]{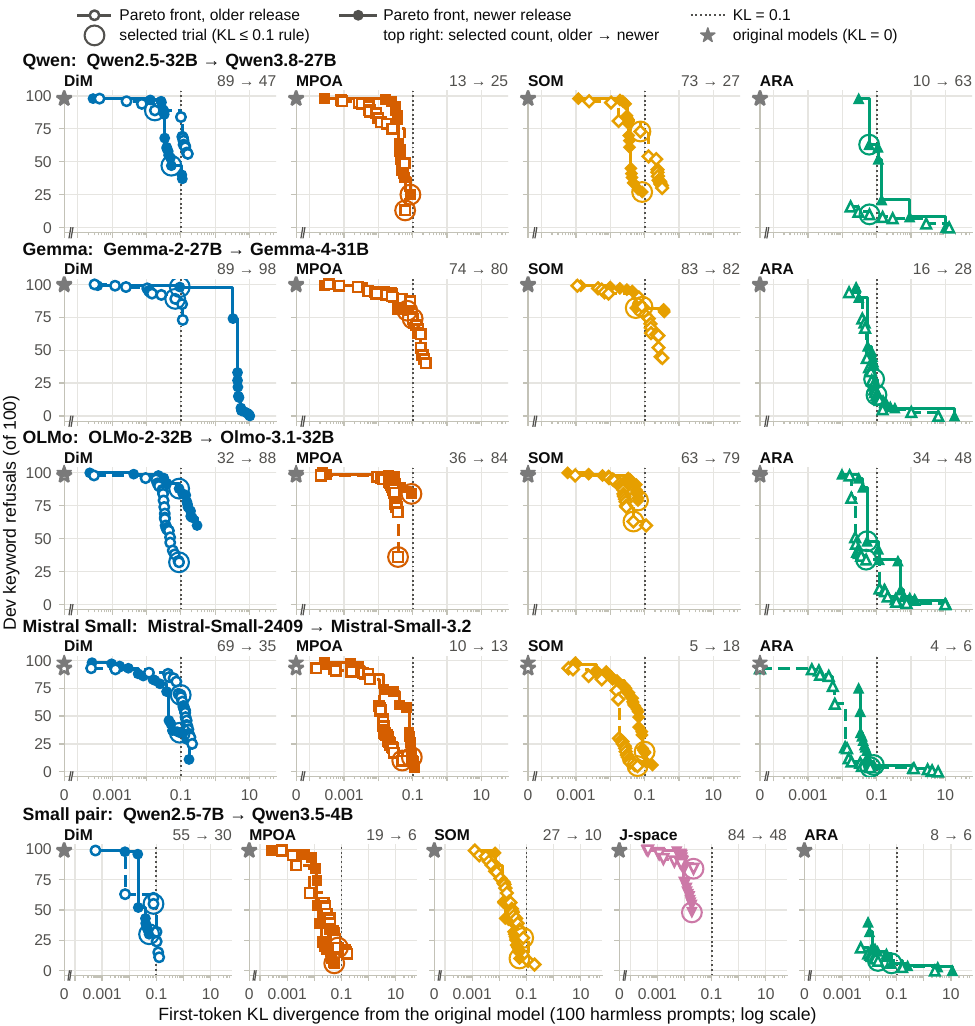}
\caption{\textbf{Refusal--KL trade-off explored by each optimisation run (dev set).} Rows are model pairs (older $\to$ newer release), columns refusal-removal methods; J-space was run on the small pair only. Each panel holds two Pareto fronts, one per release (one run each: 100 TPE trials, seed 1): the trials that no other trial of that run beats on both dev keyword refusals and first-token KL divergence; dashed with hollow markers for the older release, solid with filled markers for the newer one. The ring marks the trial chosen by the selection rule (fewest dev keyword refusals at KL $\le$ 0.1); its count is given at the top right of the panel (older $\to$ newer release). Grey stars are the original models (KL = 0). The x axis is logarithmic, with KL = 0 drawn at the left edge behind an axis break; the dotted line is the selection budget KL = 0.1. Every completed trial is shown per run in Figures~\ref{fig:trials-old} and~\ref{fig:trials-new}. Refusal counts are keyword matches on the 100 dev prompts that were also used for optimisation and selection, not held-out refusal rates.}
\label{fig:pareto}
\end{figure*}

\paragraph{Development set and search.} On the dev prompts used for search and
selection (Figure~\ref{fig:pareto}), dev keyword counts correlated with
held-out refusal across the 42 primary models ($\rho=0.77$ [0.61, 0.87]), but
the change from older to newer model could differ in sign between the two
sets; for ARA it rose by 20.3 dev refusals on average while the pooled
held-out ARA effect was small (Appendix~\ref{app:moved-results},
Figure~\ref{fig:dev-generation}). The released exports reproduced their
selected trials' dev counts to within two refusals in 37 of 46, and 34 of the
42 core searches reached their selected count by trial 75
(Figure~\ref{fig:search-convergence}).

\subsection{Summary of Key Findings}
\label{sec:res-summary}

Judged by the class-level per-pair effects, H1 was not supported: after
direction-based removal the newer model retained much more refusal in the
OLMo pair, somewhat more in the small, Qwen and Gemma pairs and none in the
Mistral Small pair, and after ARA more in the Qwen and small pairs and less in
the Gemma pair (Table~\ref{tab:pair-effects}). At the point estimate of the
run-to-run standard deviation, the only raw per-pair generation effect that
excluded zero under both variance models was the OLMo direction-based
increase. With run-to-run variance included, the between-pair heterogeneity
of the direction-based effect was high on the percentage-point scale under
the constant variance model ($I^2=80\%$), moderate under the refusal-level
model (46\%) and small on the log-odds scale (20\%). The conjunctive
criterion of H1, more refusal on the newer model for every method, was met by
no main-study pair; it is a secondary and seed-fragile summary, because the
small pair met it with seed~1 but not with seed~2, which erased the ARA
generation effect. H2 held per pair in Gemma and OLMo, but each contrast rests
on one ARA cell; the Gemma contrast lost its interval support under the
refusal-level variance and changed sign with the KL budget on the dev set,
and the Qwen contrast, positive in its point estimate and on the public item
sets (Table~\ref{tab:pair-source}), was indeterminate, because its interval
with run-to-run variance included zero. H3's prediction was not supported:
keyword matching agreed with the judge on average at the tool's window but
disagreed by several points per model.

%% file: tables/t7_pair_effects.tex
\begin{table*}[t]
\centering\scriptsize
\setlength{\tabcolsep}{2pt}
\caption{Per-pair generation effects (newer minus older model) on held-out WildGuard refusal, by method class and
scale; the primary evidence for H1 and H2. Direction: mean of DiM, MPOA and SOM (on the small pair also J-space); ARA:
the single ARA cell. Adjusted: minus the originals' own generation difference, an exploratory contrast added after H1
was stated. H2 contrast: ARA minus direction. pp: percentage points (paired item differences); log-odds: per-pair GEE.
Per-pair intervals (95\%) combine prompt sampling with run-to-run (seed) variance ($\sigma=9.15$~pp, 0.83 on the logit
scale, from the ten small-pair seed pairs). Pooled: random-effects estimate over the four main pairs with
Hartung--Knapp--Sidik--Jonkman 95\% interval ($t_3$). The pooled interval uses the prompt-sampling variances; with
run-to-run variance added its limits move by at most 0.1~pp and 0.07 log-odds, because the interval is driven by the
spread between pairs. Between-pair heterogeneity was $I^2$ = 99--100\% with prompt-sampling variances only and 20--89\%
with the seed-inclusive variances of this table (Table~\ref{tab:heterogeneity}). Gemma direction includes the failed
Gemma-4-31B DiM edit; without it the Gemma direction effect was $-$1.1~pp and the H2 contrast $-$24.9~pp. Per-method
effects are in Table~\ref{tab:h1-heldout} and the prompt-level H2 tests in Table~\ref{tab:h2-contrasts}.}
\label{tab:pair-effects}
\begin{tabular}{@{}lllllll@{}}
\toprule
Estimand, scale & Qwen & Gemma & OLMo & Mistral Small & Small pair & Pooled, 4 main pairs \\
\midrule
Direction, raw, pp & +10.7 [$-$4.0, +25.5] & +4.7 [$-$10.2, +19.6] & +36.9 [+22.1, +51.7] & $-$1.2 [$-$16.0, +13.5] & +17.3 [+4.4, +30.1] & +12.8 [$-$13.9, +39.5] \\
\quad log-odds & +0.46 [$-$0.86, +1.79] & +0.21 [$-$1.13, +1.54] & +1.63 [+0.30, +2.96] & $-$0.10 [$-$1.43, +1.23] & +0.79 [$-$0.36, +1.94] & +0.55 [$-$0.66, +1.76] \\
Direction, adjusted, pp & $-$2.4 [$-$17.3, +12.5] & +3.7 [$-$11.2, +18.6] & +36.0 [+21.2, +50.9] & $-$19.2 [$-$34.1, $-$4.2] & +0.9 [$-$12.1, +14.0] & +4.6 [$-$32.2, +41.3] \\
\quad log-odds & $-$0.53 [$-$1.86, +0.81] & +0.10 [$-$1.24, +1.44] & +1.57 [+0.24, +2.91] & $-$0.90 [$-$2.23, +0.44] & $-$0.30 [$-$1.46, +0.86] & +0.06 [$-$1.67, +1.80] \\
\midrule
ARA, raw, pp & +34.4 [+8.8, +59.9] & $-$26.0 [$-$51.6, $-$0.5] & $-$2.5 [$-$27.9, +23.0] & $-$0.7 [$-$26.1, +24.7] & +14.2 [$-$11.3, +39.7] & +1.3 [$-$38.3, +40.8] \\
\quad log-odds & +2.35 [+0.04, +4.66] & $-$4.07 [$-$6.50, $-$1.64] & $-$0.23 [$-$2.54, +2.07] & $-$0.17 [$-$2.48, +2.15] & +1.55 [$-$0.77, +3.86] & $-$0.49 [$-$4.66, +3.69] \\
ARA, adjusted, pp & +21.3 [$-$4.4, +47.0] & $-$27.1 [$-$52.7, $-$1.5] & $-$3.3 [$-$28.8, +22.2] & $-$18.6 [$-$44.2, +6.9] & $-$2.2 [$-$27.8, +23.5] & $-$6.9 [$-$40.7, +26.8] \\
\quad log-odds & +1.36 [$-$0.95, +3.68] & $-$4.17 [$-$6.61, $-$1.73] & $-$0.29 [$-$2.60, +2.01] & $-$0.96 [$-$3.28, +1.36] & +0.46 [$-$1.86, +2.78] & $-$0.96 [$-$4.59, +2.68] \\
\midrule
H2 contrast, pp & +23.6 [$-$5.9, +53.1] & $-$30.8 [$-$60.2, $-$1.3] & $-$39.3 [$-$68.8, $-$9.9] & +0.5 [$-$28.8, +29.9] & $-$3.1 [$-$31.6, +25.4] & $-$11.5 [$-$57.6, +34.6] \\
\quad log-odds & +1.89 [$-$0.78, +4.55] & $-$4.27 [$-$7.04, $-$1.51] & $-$1.87 [$-$4.53, +0.79] & $-$0.06 [$-$2.73, +2.61] & +0.76 [$-$1.83, +3.34] & $-$1.05 [$-$5.21, +3.11] \\
\bottomrule
\end{tabular}
\end{table*}

%% file: tables/t4_h1_heldout.tex
\begin{table*}[t]
\centering\footnotesize
\setlength{\tabcolsep}{3.6pt}
\caption{Held-out residual refusal of the older and newer model of each pair (H1). Rates: share of the 972
harmful held-out prompts whose response WildGuard labels a refusal, with Wilson 95\% intervals. Difference:
newer minus older model in percentage points (pp) with Newcombe 95\% intervals; $b$/$c$: prompts refused only
by the older / only by the newer model (exact McNemar test); $p_\mathrm{Holm}$: Holm over the 23 tests of the
H1 family (21 pair $\times$ method tests and two pooled terms); $^{u}$for the originals the unadjusted McNemar $p$
is given, as they are the baseline rather than an H1 test. Baseline-adjusted: the method's generation effect minus
that of the originals, pp [95\% CI]. Each modified model is the dev-set selection of its cell (first-token KL
$\le 0.1$, seed~1). $^{a}$Not significant after Holm when the 44 prompts that overlap the search prompts are
excluded. $^{b}$Failed edit: held-out refusal at or above the model's own original.
$^{s}$Seed noise: ``exceeds'' if $|$new $-$ old$|$ is larger than 25.9~pp, twice the SD (12.9~pp) of an older-minus-newer difference between two independently optimised cells, otherwise ``within'' (5 of 16 exceed); the per-cell run-to-run SD, 9.2~pp [6.4, 16.1], is estimated from the seed-1 versus seed-2 held-out refusal of the ten small-pair cells (\S\ref{sec:res-h1}) and assumed to transfer to the main pairs. Not given for the small pair, whose replicate is the source of the estimate (Figure~\ref{fig:seed-heldout}).}
\label{tab:h1-heldout}
\begin{tabular}{@{}llllrlll@{}}
\toprule
Condition & Older, \% & Newer, \% & New $-$ old, pp & $b$/$c$ & $p_\mathrm{Holm}$ & Baseline-adj., pp & Seed \\
 & [95\% CI] & [95\% CI] & [95\% CI] & & & [95\% CI] & noise$^{s}$ \\
\midrule
\multicolumn{8}{l}{\emph{Qwen: Qwen2.5-32B $\to$ Qwen3.8-27B}} \\
Original & 76.8 [74.1, 79.4] & 89.9 [87.9, 91.7] & +13.1 [+10.9, +15.4] & 6/133 & $<$0.001$^{u}$ & -- & -- \\
DiM & 47.4 [44.3, 50.6] & 74.3 [71.4, 76.9] & +26.9 [+23.8, +29.8] & 17/278 & $<$0.001 & +13.7 [+10.1, +17.1] & exceeds \\
MPOA & 3.9 [2.9, 5.3] & 14.0 [12.0, 16.3] & +10.1 [+7.8, +12.5] & 23/121 & $<$0.001 & $-$3.1 [$-$6.3, +0.2] & within \\
SOM & 42.7 [39.6, 45.8] & 38.0 [35.0, 41.1] & $-$4.7 [$-$8.0, $-$1.4] & 157/111 & 0.041$^{a}$ & $-$17.8 [$-$21.5, $-$14.2] & within \\
ARA & 6.1 [4.7, 7.8] & 40.4 [37.4, 43.5] & +34.4 [+31.2, +37.5] & 11/345 & $<$0.001 & +21.3 [+17.1, +25.4] & exceeds \\
\midrule
\multicolumn{8}{l}{\emph{Gemma: Gemma-2-27B $\to$ Gemma-4-31B}} \\
Original & 88.3 [86.1, 90.1] & 89.3 [87.2, 91.1] & +1.0 [$-$1.4, +3.5] & 69/79 & 0.460$^{u}$ & -- & -- \\
DiM & 73.8 [70.9, 76.4] & 90.1 [88.1, 91.8]$^{b}$ & +16.4 [+13.5, +19.3] & 35/194 & $<$0.001 & +15.3 [+13.0, +17.8] & within \\
MPOA & 45.2 [42.1, 48.3] & 40.4 [37.4, 43.5] & $-$4.7 [$-$8.6, $-$0.9] & 206/160 & 0.093 & $-$5.8 [$-$9.7, $-$1.7] & within \\
SOM & 65.2 [62.2, 68.2] & 67.8 [64.8, 70.7] & +2.6 [$-$1.1, +6.2] & 151/176 & 0.369 & +1.5 [$-$1.9, +4.9] & within \\
ARA & 26.6 [24.0, 29.5] & 0.6 [0.3, 1.3] & $-$26.0 [$-$28.9, $-$23.3] & 253/0 & $<$0.001 & $-$27.1 [$-$30.3, $-$23.6] & exceeds \\
\midrule
\multicolumn{8}{l}{\emph{OLMo: OLMo-2-32B $\to$ Olmo-3.1-32B}} \\
Original & 83.2 [80.8, 85.4] & 84.1 [81.6, 86.2] & +0.8 [$-$0.6, +2.2] & 19/27 & 0.302$^{u}$ & -- & -- \\
DiM & 33.5 [30.6, 36.6] & 68.3 [65.3, 71.2] & +34.8 [+31.5, +37.9] & 14/352 & $<$0.001 & +34.0 [+30.5, +37.6] & exceeds \\
MPOA & 3.6 [2.6, 5.0] & 65.0 [62.0, 68.0] & +61.4 [+58.2, +64.4] & 2/599 & $<$0.001 & +60.6 [+57.1, +64.1] & exceeds \\
SOM & 27.3 [24.6, 30.1] & 41.7 [38.6, 44.8] & +14.4 [+11.5, +17.3] & 45/185 & $<$0.001 & +13.6 [+10.4, +16.7] & within \\
ARA & 13.3 [11.3, 15.5] & 10.8 [9.0, 12.9] & $-$2.5 [$-$4.6, $-$0.3] & 67/43 & 0.111 & $-$3.3 [$-$5.8, $-$0.9] & within \\
\midrule
\multicolumn{8}{l}{\emph{Mistral Small: Mistral-Small-2409 $\to$ Mistral-Small-3.2}} \\
Original & 55.7 [52.6, 58.8] & 73.6 [70.7, 76.2] & +17.9 [+15.2, +20.6] & 21/195 & $<$0.001$^{u}$ & -- & -- \\
DiM & 32.8 [29.9, 35.8] & 28.5 [25.7, 31.4] & $-$4.3 [$-$7.4, $-$1.2] & 141/99 & 0.048$^{a}$ & $-$22.2 [$-$26.1, $-$18.4] & within \\
MPOA & 6.9 [5.5, 8.7] & 4.0 [2.9, 5.4] & $-$2.9 [$-$4.7, $-$1.2] & 51/23 & 0.014 & $-$20.8 [$-$24.0, $-$17.7] & within \\
SOM & 3.7 [2.7, 5.1] & 7.2 [5.7, 9.0] & +3.5 [+1.7, +5.3] & 22/56 & 0.001 & $-$14.4 [$-$17.7, $-$11.1] & within \\
ARA & 4.9 [3.7, 6.5] & 4.2 [3.1, 5.7] & $-$0.7 [$-$2.1, +0.6] & 23/16 & 0.369 & $-$18.6 [$-$21.6, $-$15.7] & within \\
\midrule
\multicolumn{8}{l}{\emph{Small pair: Qwen2.5-7B $\to$ Qwen3.5-4B}} \\
Original & 72.1 [69.2, 74.8] & 88.5 [86.3, 90.3] & +16.4 [+13.9, +18.9] & 11/170 & $<$0.001$^{u}$ & -- & -- \\
DiM & 20.6 [18.2, 23.2] & 54.6 [51.5, 57.7] & +34.1 [+30.5, +37.5] & 43/374 & $<$0.001 & +17.7 [+13.5, +21.8] & -- \\
MPOA & 13.8 [11.8, 16.1] & 18.0 [15.7, 20.5] & +4.2 [+1.6, +6.9] & 66/107 & 0.018 & $-$12.1 [$-$15.7, $-$8.3] & -- \\
SOM & 12.2 [10.3, 14.5] & 21.9 [19.4, 24.6] & +9.7 [+6.9, +12.4] & 50/144 & $<$0.001 & $-$6.7 [$-$10.4, $-$3.0] & -- \\
J-space & 52.6 [49.4, 55.7] & 73.8 [70.9, 76.4] & +21.2 [+18.1, +24.3] & 37/243 & $<$0.001 & +4.8 [+1.3, +8.3] & -- \\
ARA & 4.7 [3.6, 6.3] & 18.9 [16.6, 21.5] & +14.2 [+11.6, +16.9] & 24/162 & $<$0.001 & $-$2.2 [$-$6.0, +1.6] & -- \\
\bottomrule
\end{tabular}
\end{table*}

%% file: tables/t9_pair_source.tex
\begin{table*}[t]
\centering\scriptsize
\setlength{\tabcolsep}{2.4pt}
\caption{Per-pair generation effects (newer minus older model, pp) by item source: the four public sets (JailbreakBench,
XSTest unsafe, StrongREJECT, HarmBench; 772 prompts) and the author-compiled code-as-weapon bank (200 prompts).
Originals: unedited models. Direction: mean of the direction-based methods. H2: ARA minus direction. Item-bootstrap
95\% intervals, prompt sampling only (run-to-run variance not added). Many edited models refused none of the code
items, so effects on them are bounded near zero. All-item values are in Table~\ref{tab:pair-effects}; pooled log-odds
by item set in Table~\ref{tab:item-source}.}
\label{tab:pair-source}
\resizebox{\textwidth}{!}{%
\begin{tabular}{@{}lllllllll@{}}
\toprule
 & \multicolumn{2}{c}{Originals} & \multicolumn{2}{c}{Direction} & \multicolumn{2}{c}{ARA} & \multicolumn{2}{c}{H2 contrast} \\
\cmidrule(lr){2-3}\cmidrule(lr){4-5}\cmidrule(lr){6-7}\cmidrule(l){8-9}
Pair & Public & Code & Public & Code & Public & Code & Public & Code \\
\midrule
Qwen & +5.3 [+3.5, +7.1] & +43.2 [+36.5, +50.5] & +11.0 [+8.8, +13.3] & +9.4 [+6.5, +12.4] & +41.3 [+37.6, +45.1] & +7.5 [+4.0, +11.5] & +30.3 [+25.9, +34.6] & $-$1.8 [$-$5.5, +2.0] \\
Gemma & $-$5.7 [$-$7.5, $-$4.0] & +27.0 [+18.0, +36.0] & $-$6.8 [$-$9.5, $-$4.4] & +49.3 [+43.7, +55.0] & $-$32.4 [$-$35.8, $-$29.0] & $-$1.5 [$-$3.5, 0.0] & $-$25.6 [$-$29.3, $-$21.9] & $-$50.8 [$-$56.3, $-$45.2] \\
OLMo & $-$0.3 [$-$1.6, +1.0] & +5.0 [+0.5, +9.5] & +45.2 [+42.8, +47.5] & +4.8 [+2.7, +7.2] & $-$3.1 [$-$5.7, $-$0.4] & 0.0 [0.0, 0.0] & $-$48.3 [$-$51.6, $-$44.9] & $-$4.8 [$-$7.2, $-$2.7] \\
Mistral Small & +20.5 [+17.2, +23.7] & +8.0 [+4.5, +12.1] & $-$1.6 [$-$3.5, +0.2] & +0.2 [0.0, +0.5] & $-$0.9 [$-$2.5, +0.6] & 0.0 [0.0, 0.0] & +0.7 [$-$1.6, +3.0] & $-$0.2 [$-$0.5, 0.0] \\
Small pair & +9.5 [+7.1, +11.9] & +43.0 [+36.0, +50.0] & +20.2 [+17.9, +22.7] & +5.9 [+4.1, +7.6] & +17.7 [+14.6, +21.0] & +0.5 [0.0, +1.5] & $-$2.5 [$-$6.1, +1.2] & $-$5.4 [$-$7.2, $-$3.5] \\
\bottomrule
\end{tabular}}
\end{table*}

%% file: tables/t5b_h2_contrasts_main.tex
\begin{table*}[t]
\centering\footnotesize
\setlength{\tabcolsep}{3pt}
\caption{Generation $\times$ method-class interaction (H2) on held-out WildGuard refusal. ARA and direction
mean: generation effects (newer minus older model, pp) of ARA and of the mean of the direction-based methods
(DiM, MPOA, SOM; on the small pair also J-space). RD contrast: ARA minus direction mean, pp, item-bootstrap
95\% interval. GEE: generation $\times$ class interaction in log-odds (exchangeable, clustered by prompt) with
95\% interval; $p_\mathrm{Holm}$ over the six H2 tests. Pooled (main): the four main-study pairs. All
intervals here reflect prompt sampling only; the per-pair intervals with run-to-run variance are in
Table~\ref{tab:pair-effects}, and the planned mixed-model estimates in Table~\ref{tab:h2-vb}. H2 predicts a
negative interaction (ARA loses less effectiveness on the newer model); sign vs.\ H2: GEE interaction
significant after Holm and negative (predicted) or positive (opposite), or not significant (n.s.).
$^{c}$The RD contrast and the log-odds interaction have opposite signs.}
\label{tab:h2-contrasts}
\begin{tabular}{lrrllll}
\toprule
Pair & ARA, & Direction & RD contrast, pp & GEE, log-odds & $p_\mathrm{Holm}$ & Sign \\
 & pp & mean, pp & [95\% CI] & [95\% CI] & & vs.\ H2 \\
\midrule
Qwen & +34.4 & +10.7 & +23.6 [+19.9, +27.1] & +1.89 [+1.61, +2.16] & $<$0.001 & opposite \\
Gemma & $-$26.0 & +4.7 & $-$30.8 [$-$34.0, $-$27.5] & $-$4.27 [$-$5.07, $-$3.47] & $<$0.001 & predicted \\
OLMo & $-$2.5 & +36.9 & $-$39.3 [$-$42.2, $-$36.4] & $-$1.87 [$-$2.08, $-$1.65] & $<$0.001 & predicted \\
Mistral Small & $-$0.7 & $-$1.2 & +0.5 [$-$1.4, +2.4] & $-$0.06 [$-$0.37, +0.25] & 0.696 & n.s. \\
Small pair & +14.2 & +17.3 & $-$3.1 [$-$6.0, $-$0.1] & +0.76 [+0.46, +1.06] & $<$0.001 & opposite$^{c}$ \\
\midrule
Pooled (main) & -- & -- & -- & $-$0.50 [$-$0.62, $-$0.38] & $<$0.001 & predicted \\
\bottomrule
\end{tabular}
\end{table*}

%% file: tables/t8_h3_windows.tex
\begin{table*}[t]
\centering\footnotesize
\setlength{\tabcolsep}{2.8pt}
\caption{Keyword matching versus WildGuard on the same held-out responses to the 972 harmful prompts, by keyword window
(H3). Kw: refusal rate by Heretic's 33 markers in the full response (at most 512 tokens) or in its first 100 tokens,
the window Heretic scores during search; WG: WildGuard refusal on the full response. Gap: keyword minus WildGuard, pp,
with prompt-cluster bootstrap 95\% intervals; 512 and 100 tokens: keyword on that window against WildGuard on the full
response; 256 tokens: both on the first 256 tokens (Appendix~\ref{app:trunc256}). H3 predicted a negative gap. Over
the 42 ablated models the gap was positive in 34 models (24 Holm-significant) and negative in 7 (3) at 512 tokens, and
positive in 24 (14) and negative in 18 (11) at 100 tokens. Ablated: the 42 seed-1 models selected from the primary runs;
cap: the 512-token generation limit.}
\label{tab:h3-windows}
\begin{tabular}{@{}lrrrrlll@{}}
\toprule
 & & \multicolumn{2}{c}{Kw, \%} & & \multicolumn{3}{c}{Gap, pp [95\% CI]} \\
\cmidrule(lr){3-4}\cmidrule(lr){6-8}
Models & Cells & 512 & 100 & WG, \% & 512 tokens & 256 tokens & 100 tokens \\
\midrule
Originals & 10 & 82.8 & 80.9 & 80.2 & +2.6 [+1.8, +3.4] & +1.6 [+0.8, +2.4] & +0.7 [$-$0.1, +1.5] \\
\midrule
Ablated (seed 1), all & 42 & 39.9 & 32.0 & 31.5 & +8.4 [+6.9, +10.0] & +3.8 [+2.4, +5.2] & +0.5 [$-$1.0, +1.9] \\
\quad main study & 32 & 42.0 & 33.8 & 32.3 & +9.8 [+8.2, +11.3] & +4.9 [+3.5, +6.4] & +1.5 [+0.0, +2.9] \\
\quad small pair & 10 & 33.3 & 26.5 & 29.1 & +4.2 [+2.5, +5.9] & +0.4 [$-$1.4, +2.0] & $-$2.6 [$-$4.3, $-$0.9] \\
\quad responses ending before the cap & 42 & 51.2 & 42.1 & 52.4 & $-$1.2 [$-$3.2, +0.9] & $-$5.6 [$-$7.6, $-$3.8] & $-$10.4 [$-$12.2, $-$8.6] \\
\quad responses truncated at the cap & 42 & 29.1 & 22.4 & 11.4 & +17.7 [+16.2, +19.3] & +13.0 [+11.5, +14.4] & +11.0 [+9.7, +12.4] \\
\midrule
$\gamma$-MPOA control & 4 & 47.0 & 38.7 & 29.9 & +17.1 [+14.9, +19.4] & +12.0 [+9.9, +14.0] & +8.8 [+6.8, +10.9] \\
Community releases & 8 & 22.9 & 17.3 & 6.4 & +16.5 [+15.2, +17.8] & +13.6 [+12.4, +14.8] & +11.0 [+9.8, +12.1] \\
Seed-2 replicates (small pair) & 10 & 32.8 & 26.4 & 29.2 & +3.5 [+2.0, +5.0] & +0.0 [$-$1.5, +1.4] & $-$2.9 [$-$4.3, $-$1.5] \\
\midrule
All judged models & 74 & 43.3 & 36.6 & 35.0 & +8.3 [+7.0, +9.6] & +4.5 [+3.3, +5.7] & +1.7 [+0.5, +2.8] \\
\bottomrule
\end{tabular}
\end{table*}

%% file: tables/t6_h3_gap.tex
\begin{table}[t]
\centering\footnotesize
\setlength{\tabcolsep}{3pt}
\caption{Keyword matching versus WildGuard on the same held-out responses to the 972 harmful prompts (H3).
Keyword: any of Heretic's 33 refusal markers in the full response (at most 512 tokens). Gap: keyword minus
WildGuard refusal rate, pp, with prompt-cluster bootstrap 95\% interval (H3 predicted a negative gap). kw/WG only: responses counted
as refusals by keyword matching only / by WildGuard only. $\kappa$: Cohen's kappa with prompt-bootstrap 95\%
interval. Ablated: the 42 seed-1 models selected from the primary runs.}
\label{tab:h3-gap}
\begin{tabular}{lrrrrlrl}
\toprule
Models & Cells & Responses & Kw, \% & WG, \% & Gap, pp [95\% CI] & kw/WG only & $\kappa$ [95\% CI] \\
\midrule
Originals & 10 & 9{,}718 & 82.8 & 80.2 & +2.6 [+1.8, +3.4] & 531/277 & 0.73 [0.70, 0.75] \\
\midrule
Ablated (seed 1), all & 42 & 40{,}824 & 39.9 & 31.5 & +8.4 [+6.9, +10.0] & 7{,}503/4{,}059 & 0.39 [0.37, 0.41] \\
\quad main study & 32 & 31{,}104 & 42.0 & 32.3 & +9.8 [+8.2, +11.3] & 5{,}897/2{,}859 & 0.40 [0.39, 0.42] \\
\quad small pair & 10 & 9{,}720 & 33.3 & 29.1 & +4.2 [+2.5, +5.9] & 1{,}606/1{,}200 & 0.33 [0.31, 0.35] \\
\quad responses ending before the cap & 42 & 20{,}011 & 51.2 & 52.4 & $-$1.2 [$-$3.2, +0.9] & 2{,}998/3{,}240 & -- \\
\quad responses truncated at 512 tokens & 42 & 20{,}813 & 29.1 & 11.4 & +17.7 [+16.2, +19.3] & 4{,}505/819 & -- \\
\midrule
$\gamma$-MPOA control & 4 & 3{,}888 & 47.0 & 29.9 & +17.1 [+14.9, +19.4] & 1{,}038/372 & 0.26 [0.23, 0.29] \\
Community releases & 8 & 7{,}776 & 22.9 & 6.4 & +16.5 [+15.2, +17.8] & 1{,}475/192 & 0.18 [0.16, 0.21] \\
Seed-2 replicates (small pair) & 10 & 9{,}720 & 32.8 & 29.2 & +3.5 [+2.0, +5.0] & 1{,}351/1{,}010 & 0.43 [0.41, 0.46] \\
\midrule
All judged models & 74 & 71{,}926 & 43.3 & 35.0 & +8.3 [+7.0, +9.6] & 11{,}898/5{,}910 & 0.48 [0.47, 0.50] \\
\bottomrule
\end{tabular}
\end{table}

%% file: sections/06_discussion.tex
\section{Discussion}
\label{sec:discussion}

This study tested whether refusal removal is harder in the newer model of
open-weight release pairs (H1), whether the higher-rank, optimisation-based
ARA withstands any such change better than low-rank direction-based edits
(H2), and whether keyword counts agree with an automatic judge (H3). Removal
did not become uniformly harder. The newer model retained much more refusal
after direction-based removal in the OLMo pair, somewhat more in the small,
Qwen and Gemma pairs and none in the Mistral Small pair. ARA lost less
effectiveness than the direction-based edits in two pairs and, in its point
estimate, more in one, but these contrasts rest on single ARA runs and were sensitive to the variance
model, the divergence budget and the item source. At the tool's own window,
keyword counts matched the judge on average and disagreed with it model by
model. The answer to the generational question therefore depended on the
release pair, the method, the budget and the measure.

The per-pair effects (Table~\ref{tab:pair-effects}) agree with the ARA
author's account \cite{weidmann2026ara} most closely in the OLMo pair: the
originals did not differ, the direction-based edits left much more refusal on
Olmo-3.1-32B, ARA did not, and, at the point estimate of the run-to-run
standard deviation, this was the only raw per-pair increase that exceeded
run-to-run variation under both variance models. It also held on the
public and the code items (Table~\ref{tab:pair-source}) and under the second
judge. One candidate mechanism, shown in one 1B OLMo-2 model, is that more
diverse refusal openings raise the rank of refusal-related activation changes
and weaken single-vector ablation \cite{labunets2026refusal}; rank was not
measured here. The other pairs give weaker or source-dependent evidence. The
Gemma direction-based increase came from the author-compiled code items and
reversed on the public sets, and in Mistral Small every method's effect was
smaller than the originals' own difference. These pair-specific patterns
resemble non-monotonic black-box safety across Gemma releases
\cite{mitra2026cross} and model-specific abliteration susceptibility across 24
open models \cite{hadetskyi2026not}. To current knowledge they are the first
held-out, judged estimates of the generation effect for several removal
classes on released model pairs.

The pairs differ in more than release date, so the heterogeneity cannot be
attributed to family lineage. Each family contributes one step, and the steps
differ: the newer Qwen models interleave linear-attention layers with full
attention, Gemma-2 and Gemma-4 and both OLMo models apply post-normalisation
after $W_O$ and $W_{\mathrm{down}}$, so that part of an ablated direction
reaches the residual stream, and the pairs also differ in depth, attention
pattern, multimodality and post-training recipe
(Appendix~\ref{app:setup-models}). Any of these may change how well a given
edit works. Post-training can bypass a behaviour rather than remove it
\cite{lee2024mechanistic,jain2024makes}, and later layers can compensate for an
ablated component \cite{mcgrath2023hydra}; a recipe or depth that does more of
either would predict more refusal left after a single-direction edit, but
neither was measured here. How strongly the pairs differ also depends on the
scale and on the run-to-run variance assumed: with seed-inclusive variances,
$I^2$ on the percentage-point scale was 80\% under the constant and 46\% under
the refusal-level variance model, and 20\% on the log-odds scale. Four main pairs
cannot represent a population of release steps, and they show that a
conclusion drawn from one pair does not transfer to another.

H2 compared two edit families, low-rank projections of an estimated direction
and rank-50 per-module updates fitted to an objective, which differ in rank,
objective, search space and in how often the KL cap bound. The per-pair
contrasts therefore cannot show whether refusal is distributed over more
dimensions in the newer models; they show which family removed more refusal
under this protocol. ARA's advantage was large where it occurred (Gemma and
OLMo). In Qwen the point estimate was reversed, but with run-to-run variance
the contrast was indeterminate, and the positive estimate came from the
public items. The Gemma contrast lost its interval support under the
refusal-level variance model and, on the dev set, changed sign with the
budget. Each
contrast also rests on one ARA export whose adapter was refitted after the
search, and two of them on exports refitted on other hardware or re-scored
above the cap. The study is therefore a first, preliminary independent
comparison of ARA with direction-based methods across releases; support for
the ARA author's claim \cite{weidmann2026ara} is limited to the Gemma and OLMo
pairs and is conditional on these single cells. That the newer Qwen model
retained more refusal under DiM, MPOA and ARA agrees with a report that Qwen3
resisted abliteration in one setup \cite{agnihotri2025granular}, and the small
effect of DiM on Gemma-2-27B agrees with reports that single-direction edits
did little on Gemma-2 \cite{lee2026skindeep,siu2025cosmic}, although rank-1
projection sufficed for Gemma-2-9B in another setup
\cite{nakamura2026measuring}.

H3 predicted that keyword matching would undercount refusal; it did not at
any window. At Heretic's 100-token window the two measures agreed on average
but disagreed by several points for many single models, in both directions,
and over full responses keyword matching counted more refusals because safety
and disclaimer markers appear late in long answers that WildGuard labels as
compliance. This agrees with reports that disclaimers persist after
abliteration \cite{wuhrmann2026measuring}, that binary refusal metrics miss
hedged partial compliance \cite{weidener2026refusalbench}, and that
substring--judge disagreement grows with intervention strength
\cite{rocchetti2026refusal}. Which measure is correct cannot be decided without
human labels: an LLM judge over-counted refusal in disclaimer-wrapped answers
in one hand audit \cite{rocchetti2026refusal}, safety judges can degrade under
distribution shift \cite{schwinn2026coin}, and agreement between judges with
correlated errors overstates the evidence \cite{hossain2026agreement}. The
finding is therefore one of disagreement between two fallible raters: on
edited models, the keyword--judge relation changed with the edit, the model
and the length of text read, so a keyword-optimised search is steered by a
signal whose relation to judged refusal differs between models.

The $\gamma$-MPOA control shows that the post-norm leak matters for MPOA and
that it is not constant across releases: the correction lowered refusal on all
four post-norm models and more on the newer model of each pair. It did not
remove the OLMo MPOA generation effect or the pooled prompt-level H2
interaction, but it turned the Gemma MPOA effect strongly negative, at a large
capability cost on Gemma-4-31B. DiM and SOM were not corrected, so their
generation effects in the Gemma and OLMo pairs may still contain a leak that
differs between releases. The control therefore narrows, but does not remove,
architecture as an explanation of the post-norm pairs' results.

``Harder'' here means more residual refusal after a 100-trial search at
first-token KL~$\le 0.1$. Public releases reached much lower refusal than this
protocol on two newer checkpoints, at larger divergence but with small
capability changes, so residual refusal under the budget is an upper bound,
not a floor, and the protocol's ranking of checkpoints was not validated
against stronger attacks. Tampering evaluations need strong, tuned attacks to
approach worst-case behaviour \cite{che2025model,hossain2026tamperbench}.

Capability costs were small for most models, but first-token KL did not bound
them: capability change tracked teacher-forced divergence, not the selected
first-token KL, and the two largest losses occurred inside the budget. This
qualifies the report that refusal ablation preserves response quality
\cite{dombrowski2025safety}. Dev-set keyword counts would also have suggested
different conclusions from the held-out judge, as expected when one noisy
criterion is both optimised and used for selection
\cite{cawley2010overfitting}.

In sum, judged by the class-level per-pair effects, H1 was not supported:
the generation effects differed in sign and size between pairs, and the data
are consistent with the ARA author's claim in the OLMo pair and, more weakly,
in the Gemma pair, and not in the Qwen and Mistral Small pairs. The conjunctive
criterion, more refusal on the newer model for every method, was met by no
main-study pair; it is a secondary summary and fragile to the seed, since the
small pair met it with one seed but not the other. H2 held per pair in Gemma
and OLMo and was indeterminate in Qwen, with the caveats above. H3's
prediction that keyword counts undercount refusal was not supported.

Several design features address likely objections. Comparisons were made
within release pairs and against each pair's own originals, and they are
stated about the newer model of a pair, because two-point pairs cannot
establish a trend \cite{mitra2026cross}. The prompt-level heterogeneity
statistics were complemented by pair-level statistics with run-to-run
variance, under two models of that variance. The KL cap did not bind in every
direction-based run, so H2 was repeated on the cap-bound units and read at
other budgets. Item source, harmful compliance, a second judge, the 44
held-out prompts that overlap the search set and the post-norm leak were
checked separately (Appendices~\ref{app:item-source}, \ref{app:harm},
\ref{app:kl-cap}). Generation used one greedy response per prompt with
reasoning disabled, and chain of thought can carry refusal
\cite{yang2026beyond}, so the conclusions apply to greedy decoding in
non-thinking mode.

These results bear on how open-weight safety is audited. Casper et
al.~\cite{casper2026open} list as an open problem which attack configurations
are needed to stress-test open-weight safety. Here an audit of the OLMo pair
alone would have concluded that the newer release resists direction-based
removal, and one of the Mistral Small pair that it does not; within Gemma, the
DiM edit of Gemma-4-31B failed (it left refusal at the original's level), and
taken alone would have suggested a resistance that ARA did not show. Its
search found no trial that lowered dev refusal below 98\% at a first-token KL
under 3, thirty times the cap; refusal fell only in trials with larger KL
(Appendix~\ref{app:gemma4-dim}). Difference-in-means directions have failed in
a similar way on Gemma-3, where the failure has been attributed to
high-magnitude outlier channels that dominate the estimate
\cite{lai2025mpoa,mishra2026onecoordinate}, and refusal may also occupy more
than one direction \cite{wollschlager2025geometry}. Both Gemma models carry
outlier post-norm gain channels, more dispersed than in the OLMo models, so
these accounts do not by themselves explain why the edit failed on Gemma-4-31B but
not on Gemma-2-27B; the cause was not tested.
 A minimal audit protocol consistent with
these results would use at least one direction-based and one higher-rank
optimisation-based method, two or more optimisation seeds per model and
method, held-out prompts from more than one source scored by at least two
automatic judges, results at more than one divergence budget, and a
full-sequence divergence and capability check. Auditors should also have
humans label a random sample of the responses on which the judges, or a judge
and keyword matching, disagree, together with a smaller sample on which they
agree; 100 disagreements per release pair would estimate a disagreement error
rate near 10\% to within about $\pm$6 points (95\%), and 50 agreements would
check that agreement is not shared error. The present study made no such human
check (\S\ref{sec:limitations}). In this study a 100-trial search of a
22--32B model took a median of 18 minutes on one NVIDIA H200 GPU (141~GB) for a
direction-based method and 81 minutes for ARA, 18.1~h for the 31 timed
main-study searches (Appendix~\ref{sec:setup-compute}), so a second seed
roughly doubles this search time, before the evaluation of the added models.
Such an audit would still not cover sampled
decoding, reasoning modes, or fine-tuning and reinforcement-learning attacks
\cite{qi2024finetuning,russinovich2026grp}. Defences built around the refusal
direction \cite{yu2025refusal,xie2025beyond,muhamed2026decoy} should also be
tested against optimisation-based removal. For tool developers, keyword counts
and first-token KL are insufficient evidence of success, and keyword counts
should be read at the window used for selection.

\input{sections/06b_limitations}

%% file: sections/06b_limitations.tex
\subsection{Limitations and threats to validity}
\label{sec:limitations}

One limitation is that each main-study cell was optimised with a single seed
(seed~1), although TPE search is stochastic. On the small pair, a second seed
changed held-out refusal in all 10 cells (median 9.4~pp, maximum 25.2~pp;
run-to-run standard deviation 9.2~pp [6.4, 16.1]) and
erased the ARA generation effect (+14.2~pp with seed~1, +0.0~pp with
seed~2). Variance of this size could create or mask a generation effect in a
single main-study cell, as seed and configuration changes have altered
conclusions about algorithm comparisons \cite{bouthillier2021accounting} and
safeguard durability \cite{qi2025durability}. The band used to flag
main-study effects as within seed noise (Table~\ref{tab:h1-heldout}) assumes
that this variance, measured on 4B and 7B models, transfers to the 22--32B
main-study models; this cannot be tested with one seed per cell. Whether it
depends on the refusal level could not be decided from ten seed pairs, and the
per-pair verdicts differed between a constant and a refusal-level model
(Appendix~\ref{app:sigma-level}): at the point estimate of the run-to-run
standard deviation, only the OLMo direction-based increase exceeded run-to-run
variation under both, and it did not at the upper limit of the refusal-level
model. Every main-study run also used the
same seed, so runs with the same search space shared their random start-up
trials. However, full Pareto fronts and
four KL thresholds are reported, method rankings were more stable than
single-cell generation effects (method accounted for $\eta^2=0.94$ and 0.88 of
the variance among the small-pair cell rates), the pooled estimates average
run-to-run variation over the 32 main-study cells (24 direction-based, only 8
for ARA), the per-pair intervals of \S\ref{sec:res-h1} add the estimated
run-to-run variance to prompt sampling (the pair-level pooled intervals barely
change when it is added), and the 11 main-study
effects within the band are flagged as provisional. Three of the ten seed-2
runs also used a smaller generation batch size after running out of memory;
without them the run-to-run standard deviation was 7.2~pp instead of 9.2~pp,
and the same five raw effects exceeded the band. A second seed for the 32
main-study cells, with distinct seeds per model, which would replace the
transferred variance with a direct estimate, is left to future work. The
re-run rule was not uniform: an interrupted Qwen3.8-27B ARA study was repeated
from scratch, whereas the Qwen3.8-27B MPOA study was resumed
(Appendix~\ref{app:setup-opt}). Its selected trial is a seeded start-up draw
that an uninterrupted study would also have evaluated, so the resumption can
only have raised its dev refusal rate, which would favour finding the newer
Qwen model harder for MPOA; the one study run both ways (ARA) differed by 4 of
100 dev refusals. Likewise, for $\gamma$-MPOA the post-norm component along
$\hat{\mathbf{r}}$ was not measured in forward passes of the selected models:
only the weight identity $\mathbf{b}^{\top}W'=(1-\lambda)\,\mathbf{b}^{\top}W$
was checked (Appendix~\ref{app:setup-methods}).

A second limitation is that each held-out prompt received one greedy
response of at most 512 tokens, with reasoning disabled although the chat
templates of Qwen3.8-27B and Qwen3.5-4B open a reasoning block. Sampled
decoding can change refusal rates substantially \cite{huang2024catastrophic},
and chain of thought can carry refusal \cite{yang2026beyond}, possibly by
different amounts in each generation. However, decoding was identical for
every model, so within-pair comparisons are matched on it, and the
conclusions are stated for greedy decoding in non-thinking mode.

The ARA cells carry limitations of their own. The released ARA model is
rebuilt by refitting its adapter, and the four ARA exports refitted on a
different GPU from the search drifted by 5 to 9 dev refusals on re-score,
whereas those refitted on the search machine drifted by at most 2; the
Gemma-4-31B ARA export re-scored at a first-token KL of 0.107, above the cap;
and the 0 of 100 dev refusals reported by heretic-org on its model card (an
unreviewed report) for an ARA model of
Qwen3.8-27B with the same defaults was not reproduced (59 and 63 of 100 in two
runs here). Held-out results describe the released models, so these points
concern the link between search and release rather than the reported rates,
but they bear directly on H2, whose reversed Qwen estimate and Gemma support both rest
on single ARA cells. The per-pair H2 verdicts are therefore preliminary.
Running heretic-org's reported trial through this pipeline, refitting the ARA
exports on the search hardware and a second ARA seed for the Qwen and Gemma
pairs are left to future work.

A further limitation is that trials are optimised and selected on the 100
development prompts that produce Heretic's refusal count, which relies on 33
keyword markers with a reported recall of about 0.48, against 0.77 for an LLM
judge (Heretic~\#419, an unreviewed practitioner report). Development-set refusal is therefore optimistic
\cite{cawley2010overfitting}, and the best dev method was the best held-out
method in only 7 of 12 model $\times$ seed sets. However, keyword counts drive
optimisation and selection only: reported refusal comes from a frozen held-out
set of 1,322 prompts (Table~\ref{tab:refusal-sets}), labelled by WildGuard as
primary judge with Qwen3Guard as a second judge ($\kappa=0.82$ [0.80, 0.83]
on the 42 ablated models).
A related limitation is that no human audit was made. Every refusal rate is
therefore relative to the automatic judges, WildGuard has not been validated
on the outputs of edited models, and where keyword matching and the judges
disagreed (H3) neither can be called correct; agreement between LLM judges
with correlated errors overstates the evidence \cite{hossain2026agreement}.
WildGuard and both OLMo releases come from the same developer, whose open
supervised post-training data for both releases include about 50,000 prompts
from WildGuardMix, the data set released with WildGuard and used to train it\footnote{Dataset
cards of \texttt{allenai/tulu-3-sft-olmo-2-mixture} (OLMo-2) and
\texttt{allenai/Dolci-Instruct-SFT} (Olmo-3.1) on Hugging Face, accessed
8 November 2026.}, so its labels of OLMo outputs may share their provenance; under the second judge,
Qwen3Guard, the OLMo direction-based generation effect was +32.3~pp (WildGuard
+36.9~pp) and the OLMo H2 contrast $-$35.0~pp ($-$39.3~pp). The keyword count
that drives search and selection may also be a less valid signal on one
generation than on the other, which would bias the selected models
differently; dev and held-out changes had opposite signs in several cells
(Appendix~\ref{app:moved-results}), and re-selecting trials with a judge
would test this.

Another limitation is that generation is confounded with size, architecture,
training data, post-training recipe and release interval, and later
checkpoints may have been safety-tuned on the public held-out benchmarks
\cite{sainz2023nlp}. In particular,
post-norm models (Gemma-2/4, OLMo-2/Olmo-3.1) apply an RMSNorm \cite{zhang2019rmsnorm} after $W_O$ and
$W_{\mathrm{down}}$, so the normalised output retains a component along the
ablated direction, handicapping direction-based methods on four of the eight
main-study models. Because both generations of each post-norm pair
share this design
\cite{gemmateam2024gemma2,kim2025periln,walsh2025olmo2,olmo2025olmo3}, the leak confounds H1 and H2 only to the extent that its
size differs between generations. However, comparisons are made within family
and against each pair's own originals, with family as a random effect in the
planned model \cite{baayen2008mixed}; the $\boldsymbol{\gamma}$-corrected MPOA
control on the four post-norm models lowered held-out refusal on all four, by
more on the newer model of each pair, but left the OLMo generation effect
(+57.0~pp) and the pooled H2 interaction ($-$0.42) in place; DiM and SOM were
not corrected; and conclusions are stated about later releases, not about
mechanisms such as safety training. The 200 code-as-weapon items were compiled
by the author and carried about half of the originals' generation difference;
some per-pair verdicts, notably Gemma's direction-based increase, depended on
them (Table~\ref{tab:pair-source}).

A final limitation concerns the scope of two implementations. J-space
ablation is limited to the small pair, because instruction-tuned Jacobian
lenses exist only for Qwen3.5-4B and Qwen2.5-7B-Instruct; the lens is fitted on
non-chat text, and the direction is baked into the weights rather than applied
as the reference forward hook, with unknown direction of bias; J-space as a
forward hook ($\alpha=1$) was planned on the small pair but not run and is
left to future work.
Heretic-tuned SOM deviates from Piras et al.~\cite{piras2025som} in neuron
ranking, orthonormalisation, decay, layer scope and adapter rank, and its
narrower search may understate SOM; a paper-faithful SOM configuration was
planned but not run and is left to future work.
However, both are evaluated, and labelled, as Heretic variants
(Heretic-tuned SOM; weight-baked J-space), so the conclusions about them are
restricted to these variants.
All threats,
including those not discussed here (tokenisation, truncation, decoding,
first-token KL, benchmark exposure,
benchmark coverage, judge validity, preregistration, generalisation, and dual
use), are detailed in Appendix~\ref{app:validity}.

%% file: sections/07_conclusion.tex
\section{Conclusion}
\label{sec:conclusion}

This study compared direction-based and optimisation-based refusal removal on
five open-weight release pairs under one nominal search budget, KL cap and
selection rule, with held-out refusal labelled by automatic judges. Measured
as residual refusal after a 100-trial search at first-token KL divergence
$\le 0.1$, refusal removal did not become uniformly harder in the later
releases. The generation effect differed in sign and size between pairs; at
the point estimate of the run-to-run standard deviation, only the OLMo pair's
direction-based increase exceeded run-to-run variation under both variance
models considered, and some per-pair verdicts depended on the
source of the test prompts. In a preliminary comparison resting on one ARA run
per model, ARA lost less effectiveness than the direction-based methods in two
pairs and, in its point estimate, more in one, and these verdicts changed with the variance model and
the divergence budget. At the tool's 100-token window, keyword counts matched
judged refusal on average but not model by model, and first-token KL did not
bound capability loss. Claims that a newer model resists refusal removal, or
that a removal tool succeeds, therefore need several release pairs analysed as
pairs, the unedited models as reference, several method classes, held-out
judged evaluation, more than one divergence budget, a full-sequence measure of
collateral change and several optimisation seeds per model; the present study
met the last requirement only on the small pair.

Four directions follow. First, a second optimisation seed for the 32
main-study cells would replace the transferred run-to-run variance with a
direct estimate. Second, running heretic-org's reported Qwen3.8-27B ARA trial
through this pipeline, and refitting the ARA exports on the search hardware,
would show whether the reversed Qwen estimate and the Gemma-4-31B result depend on the
search or on the release step. Third, a human-labelled audit of the responses
on which keyword matching and the judges disagree would show which measure is
closer to human judgement on edited models \cite{gao2026how}. Fourth, judge
labels for non-selected trials would allow Pareto fronts to be re-selected
with a judged objective and a full-sequence divergence constraint. Until such
checks are routine, a model's apparent resistance to refusal removal should be
read as a statement about one release pair, one method, one budget and one
measure, not about its generation.

%% file: sections/08_appendix.tex
\input{sections/08a_method_definitions}

\input{sections/08b_setup_details}
\input{sections/08c_validity}
\input{sections/08d_additional_results}
\clearpage
\input{sections/08e_truncation256}
\input{sections/08f_dev_rescore}
\input{sections/08g_selection_pairs}
\input{sections/08h_item_source_harm}
\input{sections/08i_revision_details}

%% file: sections/08a_method_definitions.tex
\section{Method Definitions}
\label{app:method-defs}
\label{sec:methods}
\label{app:methods}%

This appendix gives the notation, the update rule and the rationale of each
refusal-removal method reviewed in Section~\ref{sec:related}, together with
the strengths and weaknesses of each and the way it is instantiated in this
study.

\paragraph{Notation.}
A decoder-only transformer maintains a residual stream
$\mathbf{x}^{(\ell)} \in \mathbb{R}^{d}$ at each layer $\ell$. Each layer reads
from the stream and writes back to it through two matrices:
the attention output projection $W_O^{(\ell)}$ and the MLP down-projection
$W_{\mathrm{down}}^{(\ell)}$. Let $\mathcal{D}_{\mathrm{harm}}$ and
$\mathcal{D}_{\mathrm{safe}}$ be sets of harmful and harmless prompts, and let
$\boldsymbol{\mu}^{(\ell)}_{\mathrm{harm}}$ and
$\boldsymbol{\mu}^{(\ell)}_{\mathrm{safe}}$ be the mean residual activations at
the last prompt token over each set.

Methods are grouped by \emph{what they assume about refusal}: (i) it is a single
linear direction, (ii) it is a low-dimensional subspace or cone, (iii) it is
whatever transformation of each module's outputs separates harmful from
harmless inputs (no geometric assumption), or (iv) it is best located in a
different representation space altogether. Table~\ref{tab:methods}
(Section~\ref{sec:related}) summarises
the evaluated methods as instantiated in this study (\S\ref{sec:setup-methods}).

\subsection{Single-direction methods}

\subsubsection{Classic abliteration (difference-in-means)}
\label{sec:m-dim}
\textbf{Mechanism.} Compute the refusal direction
$\hat{\mathbf{r}}^{(\ell)} =
\mathrm{normalize}(\boldsymbol{\mu}^{(\ell)}_{\mathrm{harm}} -
\boldsymbol{\mu}^{(\ell)}_{\mathrm{safe}})$, pick one layer's direction
$\hat{\mathbf{r}}$, and orthogonalise every matrix that writes to the residual
stream against it:
$W \leftarrow W - \lambda\,\hat{\mathbf{r}}\hat{\mathbf{r}}^{\top} W$
\cite{arditi2024refusal}. With $\lambda=1$ the model can no longer write any
component along $\hat{\mathbf{r}}$.

\textbf{Why it works.} Under the linear representation hypothesis
\cite{park2024linear}, high-level features correspond to directions.
Arditi et al.~\cite{arditi2024refusal} showed causally that adding $\hat{\mathbf{r}}$
induces refusal on harmless prompts and ablating it suppresses refusal on
harmful ones, across 13 chat models. Because the edit is a rank-one projection,
most of the weight matrix is untouched.

\textbf{Why it exists.} It turned an interpretability finding into a
practical, fine-tuning-free procedure that needs only forward passes on a few
hundred prompts \cite{labonne2024abliteration,failspy_abliterator}.

\textbf{Pros.} Simple, fast (minutes), interpretable, rank-one, few
hyperparameters (layer, strength).
\textbf{Cons.} Assumes refusal is one-dimensional; the direction also carries
part of the harmless mean, so removing it damages unrelated behaviour (refusal
is better described as an affine function whose offset plain ablation ignores;
on a recurrent RWKV model plain ablation produced nonsense where affine concept
editing stayed coherent \cite{marshall2024refusal}); changing row norms shifts
activation scales; quality depends heavily on the chosen layer and strength,
which are usually set by hand (in an earlier cross-tool comparison, GSM8K
accuracy changed by between $+1.51$ and $-18.81$ percentage points
\cite{young2025comparative}). DECCP, ErisForge and
FailSpy's tool are implementations of this recipe with different layer-selection
heuristics. In this study diff-in-means runs inside Heretic with the same
automated search over direction, strength and layer profile as MPOA
(``Heretic-tuned DiM'', \S\ref{sec:setup-methods}).

\subsubsection{Projected and magnitude-preserving abliteration (MPOA)}
\label{sec:m-mpoa}
\textbf{Mechanism.} Two refinements of \S\ref{sec:m-dim}.
(1) \emph{Projection}: remove from $\hat{\mathbf{r}}$ its component along the
normalised harmless mean, so that only the part of the refusal direction
orthogonal to ordinary activations is ablated \cite{lai2024projected}.
(2) \emph{Magnitude preservation}: row-normalise $W$, apply the ablation, then
restore each row's original norm \cite{lai2025mpoa}.
Heretic adds a search over a per-layer weight kernel (peak strength, peak
position, minimum strength, falloff distance) for attention and MLP separately,
global versus per-layer directions, and a fractional direction index
interpolated between layers, all tuned by multivariate TPE
\cite{akiba2019optuna,heretic2025}. The edit is stored as a low-rank LoRA
adapter \cite{hu2022lora}.

\textbf{Why it works.} Projection reduces collateral damage: the part of
$\hat{\mathbf{r}}$ shared with harmless activations is exactly what is
needed for normal behaviour. Norm preservation keeps downstream layer norms and
activation magnitudes in the range the model was trained on, so edits
change \emph{direction} rather than \emph{scale}.

\textbf{Why it exists.} Plain abliteration measurably degraded capability
(for example GSM8K drops of several points in an earlier comparison
\cite{young2025comparative}); these refinements recover much of that loss.

\textbf{Pros.} Better capability retention at the same refusal rate;
automated hyperparameter search removes manual tuning.
\textbf{Cons.} Still assumes a single direction per layer; norm preservation is
only approximate when stored as a low-rank adapter; the search adds many
evaluation passes (one per trial).

\subsubsection{Null-space-constrained abliteration (jwest33)}
\label{sec:m-null}
\textbf{Mechanism.} Compute a projected, norm-rescaled diff-in-means edit
$\Delta W$ as above, then restrict it to the null space of activations on a
set of \emph{preservation} prompts: with $V$ the top singular vectors of
those activations, apply $\Delta W (I - V V^{\top})$
(the AlphaEdit constraint \cite{fang2025alphaedit}). Optional Gaussian layer
weighting spreads the edit around a chosen depth.

\textbf{Why it works.} If an input lies in the span of $V$, the constrained
edit has no effect on it, so behaviour on preserved inputs is guaranteed
unchanged to first order; the edit can only act on directions that harmless
inputs do not use.

\textbf{Why it exists.} It brings guarantees from knowledge editing
\cite{meng2022locating,fang2025alphaedit} to refusal removal, as an
alternative to tuning strength against a KL budget.

\textbf{Pros.} Principled protection of chosen behaviour; single pass, no
search.
\textbf{Cons.} If harmful and harmless activations share directions (likely in
newer models), the constraint also blocks the refusal edit; the rank ratio is a
new hyperparameter; dense models only in the current implementation. It is not
evaluated in this study (\S\ref{sec:m-not}).

\subsection{Multi-direction and subspace methods}

\subsubsection{SVD multi-direction (Gabliteration, OBLITERATUS presets)}
\label{sec:m-svd}
\textbf{Mechanism.} Instead of the mean difference, take the top-$k$ singular
vectors of the matrix of per-prompt differences between harmful and harmless
activations, and project all $k$ out, typically with ridge regularisation and
selected layers \cite{gabliteration2024}. OBLITERATUS packages this
(``advanced'', 4 directions), whitened and iterative variants (``aggressive''),
frequency-band weighting of layers via a DCT (``spectral cascade''), and a
re-implementation of gradient-refined directions (``rdo'') in one toolkit.

\textbf{Why it works.} When refusal is spread over several correlated
directions, one mean difference captures only the dominant one; the remaining
components can re-trigger refusal. Removing a subspace closes those
alternative routes.

\textbf{Why it exists.} Single-direction ablation leaves residual refusals on
some prompt categories and some models; multi-direction methods target that
residual.

\textbf{Pros.} Removes more refusal on models where one direction is
insufficient; still closed-form and fast.
\textbf{Cons.} Every extra direction removed is extra capacity destroyed;
choosing $k$ trades refusal against damage; several presets are heuristic
combinations without a clear theoretical justification. The OBLITERATUS
v0.1.4 presets \texttt{advanced} and \texttt{spectral\_cascade}, together with
\texttt{basic}, are not evaluated in this study; they were run on
Qwen2.5-7B-Instruct and scored with this study's harness in the companion
paper \cite{young2026paper30}. Version 0.1.4 lists a \texttt{som} preset but
does not ship its direction module and does not run on Qwen3.5-4B
(Appendix~\ref{app:setup-methods}).

\subsubsection{Refusal cones (RDO)}
\label{sec:m-rdo}
\textbf{Mechanism.} Learn refusal directions by gradient descent on an
objective that requires ablation to remove refusal and addition to induce it,
while keeping other behaviour fixed; multiple such directions form a
polyhedral \emph{concept cone} \cite{wollschlager2025geometry}.

\textbf{Why it works / exists.} It showed that refusal is not a single
direction but a multi-dimensional cone of causally relevant, representationally
independent directions, and that directions found by optimisation outperform
diff-in-means.
\textbf{Pros.} Directly targets the causal effect rather than a statistical
proxy. \textbf{Cons.} Requires gradients and more compute. It is not evaluated
in this study (\S\ref{sec:m-not}).

\subsubsection{Self-organising-map directions (SOM)}
\label{sec:m-som}
\textbf{Mechanism.} Fit a small self-organising map (e.g.\ $4\times4$) to the
harmful activations at each layer, take $k$ map neurons as prototypes of
distinct ``kinds'' of harmful input, and use each prototype minus the harmless
mean as a refusal direction; ablate all $k$, each with its own strength
\cite{piras2025som}. The proof-of-concept Heretic integration (PR~\#196, an unreviewed practitioner report)
tunes the per-direction strengths with TPE.

\textbf{Why it works.} Different harm categories (weapons, self-harm, cyber,
\ldots) may trigger refusal through different directions; a single mean
averages them and represents none well. The SOM finds the clusters
without labels. Piras et al.~\cite{piras2025som} show diff-in-means is a special case.

\textbf{Pros.} Strong reported results (3/100 refusals at KL $\approx 0.1$ on
gpt-oss-20b, PR~\#196, unreviewed); principled multi-direction choice.
\textbf{Cons.} More hyperparameters (map size, $k$, per-direction strengths)
make TPE converge slowly; the PoC's top-$k$ selection counts winners over the
map's own weights instead of the data. In this study SOM is ported to
Heretic~2.0 as a modifier (``Heretic-tuned SOM'') that ranks neurons by wins
over the harmful residuals, orthonormalises the $k$ directions and replaces the
per-direction strengths with one searched decay; these and the other deviations
from Piras et al.~\cite{piras2025som} are listed in \S\ref{sec:setup-deviations}.

\subsection{Optimisation-based methods}

\subsubsection{Arbitrary-Rank Ablation (ARA)}
\label{sec:m-ara}
\textbf{Mechanism.} ARA makes no assumption about the geometry of refusal
\cite{weidmann2026ara}. For each targeted linear module $W$ in a layer range
(in Heretic, the attention output projection and the MLP down-projection), it
records the module's inputs $X$ and outputs $Y = XW^{\top}$ on harmless
($g$) and harmful ($b$) prompts, then learns a low-rank update
$\Delta W = BA$ (rank 50 by default) minimising
\begin{equation}
\label{eq:ara-objective}
\mathcal{L} = w_p\,\big\|\tilde{Y}_g - Y_g\big\|^2
 + w_s\Big[\,d_k(\tilde{Y}_b, Y_g) - w_o\, d_k(\tilde{Y}_b, Y_b)\Big],
\end{equation}
where $\tilde{Y}$ is the output of $W+BA$ with each row rescaled to the norm of
the corresponding row of $W$, and $d_k(\cdot,\cdot)$ is the mean distance to the
$k$ nearest neighbours. The rescaling enters only the objective; the exported
adapter $BA$ is not renormalised (\S\ref{sec:setup-methods}).
The three terms keep harmless outputs fixed, pull harmful outputs toward
harmless ones, and push them away from where they were (``overcorrection'').
Each module is solved independently with L-BFGS; Heretic's TPE searches the
layer range, the three weights and $k$.

\textbf{Why it works.} Rather than guessing the shape of the refusal
representation, ARA directly asks each module to make harmful inputs produce
the outputs harmless inputs produce. Any refusal mechanism that a module's
output carries, whether linear, multi-directional or non-linear across
prompts, is in scope, up to the adapter's rank. The objective is close to
convex and starts at its optimum for the preservation term, so L-BFGS
converges in a few iterations.

\textbf{Why it exists.} Its author reports that directional ablation
increasingly fails on recent models, and that ARA reaches lower refusal at
lower KL on gpt-oss-20b than any directional result \cite{weidmann2026ara}.

\textbf{Pros.} No geometric assumption; fewer search parameters than
per-component directional kernels; memory similar to abliteration because
modules are solved one at a time.
\textbf{Cons.} Slower per trial; rank-50 adapters per module are larger
artefacts; the nearest-neighbour objective depends on the prompt sets, so it
may overfit them; no published analysis yet (preprint forthcoming).

\subsection{Methods in other representation spaces}

\subsubsection{Jacobian-space (J-space) ablation}
\label{sec:m-jspace}
\textbf{Mechanism.} Using a Jacobian lens, map each layer's activations into a
``workspace'' where features are read out by the model's later computation;
compute the refusal direction there and pull it back to the residual stream
through the layer Jacobian, $\mathbf{d}_\ell = J_\ell^{\top}(\cdot)$; ablate it
with a forward hook rather than a weight edit
\cite{sekinal2026jspace}.

\textbf{Why it exists.} To localise the part of the refusal signal the model
actually uses downstream, rather than everything correlated with harmfulness.
\textbf{Pros.} Very targeted (small distribution shift).
\textbf{Cons.} By its authors' own measurements removes little refusal
(AdvBench refusal 0.78 versus 0.06 for diff-in-means on Qwen3.5-4B); the
reference code is hook-based, so not directly an exportable model, and is a
single-model implementation. In this study the J-space directions are instead
baked into the weights with Heretic's directional-ablation edit and search
(full row normalisation, no harmless-mean projection), which yields an
exportable model (\S\ref{sec:setup-methods}); the reference forward hook is
planned as a further condition. Because, among the candidate models,
instruction-tuned Jacobian lenses (\texttt{neuronpedia/jacobian-lens}) exist
only for the two small-pair models, J-space is run on the small pair only.

\subsection{Methods not evaluated}
\label{sec:m-not}
Null-space-constrained abliteration (\S\ref{sec:m-null}) and refusal cones
(\S\ref{sec:m-rdo}) are described above for completeness but are not part of
the final design. A code audit before the main runs found that the available
null-space implementation (\texttt{jwest33/abliterator}) applies the
constraint on the output side, where it gives no preservation guarantee, and
edits square attention output projections, as in Qwen2.5-32B, OLMo-2-32B,
Olmo-3.1-32B and Mistral-Small-2409, in the wrong space; the revised
plan therefore dropped it. The only RDO implementation considered, the
\texttt{rdo} preset of OBLITERATUS, is a probe-based proxy rather than the
gradient-based optimisation of Wollschl\"ager et
al.~\cite{wollschlager2025geometry}, and it is not among the OBLITERATUS
presets evaluated in the companion paper \cite{young2026paper30}.
Other diff-in-means implementations (DECCP, ErisForge, FailSpy's tool) are not
run separately; Heretic-tuned DiM represents this recipe
\cite{young2025comparative}.
Optimal-transport ablation maps the harmful activation distribution onto the
harmless one with a closed-form Gaussian transport map at one or two layers
and reports up to 11\% higher attack success than baselines on six 7--32B
models \cite{nanfack2026ot}; no public code was found. Weighted refusal
direction optimisation (WRDO) learns weights over a bank of per-prompt refusal
directions and was benchmarked against basic refusal-direction ablation, RDO,
RDO cones and Heretic on Qwen3-8B-Base and Olmo~3 7B \cite{evseev2026wrdo};
it is not evaluated. Circuit-breaker training
\cite{zou2024circuit} and extended-refusal fine-tuning
\cite{abushairah2025defense} are defences rather than removal methods and are
discussed in Section~\ref{sec:related}.

%% file: sections/08b_setup_details.tex
\section{Experimental Details}
\label{app:setup-details}

This appendix gives the implementation detail behind the Methods
(\S\ref{sec:setup}), in the same order: models, method instantiation and
search spaces, optimisation and prompt splits, held-out evaluation, export and
release, multiple-testing families, compute and software, and deviations from
the study plan.

\subsection{Models}
\label{app:setup-models}

Except for the Mistral-Small-3.2 mirror, Heretic was not given an explicit
revision; it used the locally cached snapshot, whose commit
(Table~\ref{tab:models}) equals the commit recorded by the evaluation harness
wherever both exist. The small pair (Qwen2.5-7B-Instruct and Qwen3.5-4B) fits a
24\,GB consumer GPU.

Three architectural properties matter. First, Gemma-2, Gemma-4, OLMo-2 and
Olmo-3.1 apply an RMSNorm \cite{zhang2019rmsnorm} with elementwise gain
$\boldsymbol{\gamma}$ to the outputs of $W_O$ and $W_{\mathrm{down}}$ before
they enter the residual stream (``post-norm''; output normalisation in the
terminology of Kim et al.~\cite{kim2025periln}, distinct from classic Post-LN,
which normalises after the residual addition). Gemma~2 normalises both the
input and the output of each sublayer \cite{gemmateam2024gemma2}, whereas
OLMo~2 and Olmo~3 normalise outputs only \cite{walsh2025olmo2,olmo2025olmo3}. Orthogonalising $W$ against $\hat{\mathbf{r}}$ then removes
$\hat{\mathbf{r}}$ only from the pre-norm output $\mathbf{u}$; the residual
stream still receives
$(\boldsymbol{\gamma}\odot\hat{\mathbf{r}})^{\top}\mathbf{u}/\mathrm{rms}(\mathbf{u})$
along $\hat{\mathbf{r}}$, which vanishes only for constant
$\boldsymbol{\gamma}$. Direction-based edits are thus incomplete on four of the
eight main-study models; this potential confound is disclosed rather than
corrected in the primary runs, and a $\boldsymbol{\gamma}$-corrected MPOA
control was run on all four post-norm models (\S\ref{app:setup-methods};
results in \S\ref{sec:res-secondary}). Second, Qwen3.8-27B and Qwen3.5-4B use the
Transformers \texttt{qwen3\_5} model class, which interleaves gated-delta-rule
linear-attention layers \cite{yang2025gated} with full-attention layers (three to one in
Qwen3.5-4B); on linear-attention layers Heretic edits the linear-attention
output projection in place of $W_O$. Third, Qwen3.8-27B, Gemma-4-31B,
Mistral-Small-3.2 and Qwen3.5-4B are image-text checkpoints used with text
only, and the Gemma-2
chat template has no system role, so the system prompt is prepended to the
user turn (in Heretic and in the evaluation harness).

\subsection{Method Instantiation and Search Spaces}
\label{app:setup-methods}

All methods run inside Heretic~2.0 \cite{heretic2025}, so prompt sets,
scorers, trial budget and selection are shared, while the weight edit and its
search space differ (Table~\ref{tab:search}). In every layer Heretic targets the attention output projection
and the MLP down-projection. During optimisation each edit is held in a PEFT
LoRA adapter \cite{hu2022lora} ($\Delta W=BA$, with LoRA $\alpha$ equal to the
rank so that the scaling $\alpha/r$ is 1 and the adapter acts at full strength;
no dropout); adapters are reset to zero before every
trial and the base weights are not modified. Models were loaded unquantised in
bf16; activation winsorisation was off. Built-in methods on the GPU boxes ran
from upstream Heretic commit \texttt{0db5c31}; SOM, J-space and all small-pair
runs ran from the fork branch \texttt{paper33-plugins} at \texttt{8fd4d76},
which only adds the SOM (commit \texttt{45a4e6e}) and
\texttt{PrecomputedDirections} modifiers, SOM tests and a SOM section of the
configuration template to \texttt{0db5c31}, leaving the built-in modifiers
identical. The $\gamma$-MPOA control ran from a third fork branch,
\texttt{paper33-gamma} at \texttt{fdfe54b}, which adds only the $\gamma$-MPOA
modifier to \texttt{8fd4d76}. On the GPU boxes the fork code was loaded
through \texttt{PYTHONPATH} into the \texttt{0db5c31} environment, so the run
records there name \texttt{0db5c31}; the fork commits are recorded in the
environment captures. Configurations are
\texttt{configs/core\_\{dim,mpoa,ara,som,jspace,gmpoa\}.toml};
Table~\ref{tab:search} lists the search spaces.

\paragraph{DiM.} The refusal direction of layer $\ell$ is the normalised
difference of harmful and harmless mean residuals at the last prompt position
\cite{arditi2024refusal}, removed by the rank-1 edit
$\Delta W=-\lambda\,\hat{\mathbf{r}}(\hat{\mathbf{r}}^{\top}W)$ with neither
projection nor row normalisation. Unlike the original procedure
(\S\ref{sec:m-dim}), only $W_O$ and $W_{\mathrm{down}}$ are edited, the
embedding matrix is not, and direction, strength and layer profile are tuned
by the same search as MPOA (``Heretic-tuned DiM''). The strength $\lambda$
follows a per-component kernel: the maximum weight at the maximum-weight
position, decreasing linearly to the minimum weight at the minimum-weight
distance, and zero beyond. A global direction is interpolated between the two
layers adjacent to a fractional index; the per-layer scope uses each layer's
own direction.

\paragraph{MPOA.} Heretic's defaults: the direction is projected orthogonal to
the normalised harmless mean \cite{lai2024projected}, and the edit is computed
on row-normalised weights whose row norms are then restored
\cite{lai2025mpoa}, approximated by a rank-3 adapter from a randomised SVD
reseeded with the run seed.

\paragraph{$\gamma$-MPOA.} In a post-norm block the residual stream receives
$\mathbf{y}=\boldsymbol{\Gamma}\mathbf{o}/\mathrm{rms}(\mathbf{o})$, with
$\mathbf{o}=W\mathbf{z}$ the writer's output and
$\boldsymbol{\Gamma}=\mathrm{diag}(\boldsymbol{\gamma})$ the gain of the
following RMSNorm, so its component along the refusal direction
$\hat{\mathbf{r}}$ is $(\boldsymbol{\Gamma}\hat{\mathbf{r}})^{\top}\mathbf{o}/\mathrm{rms}(\mathbf{o})$.
$\gamma$-MPOA replaces MPOA's projection by the oblique projection
\[
W' = W - \lambda\,\mathbf{a}\mathbf{b}^{\top}W,\qquad
\mathbf{b}=\boldsymbol{\Gamma}\hat{\mathbf{r}},\qquad \mathbf{b}^{\top}\mathbf{a}=1,
\]
so that $\mathbf{b}^{\top}W'=(1-\lambda)\,\mathbf{b}^{\top}W$ and the write
along $\hat{\mathbf{r}}$ vanishes at $\lambda=1$, the per-token
$\mathrm{rms}(\mathbf{o})$ being a scalar; $\lambda$ is MPOA's per-layer
weight from the search. The removal direction is a ridge-regularised inverse,
$\mathbf{a}\propto(\boldsymbol{\Gamma}^2+\delta^2 I)^{-1}\boldsymbol{\Gamma}\hat{\mathbf{r}}$
with $\delta=0.1\,\mathrm{rms}(\boldsymbol{\gamma})$ of the writer's gain,
because many gains are near zero. MPOA's row-norm restoration is then applied
but confined to the kernel of $\mathbf{b}^{\top}$, so that it cannot
reintroduce a component along $\hat{\mathbf{r}}$, and the edit is exported as
the same rank-3 adapter (one exact rank-1 term plus a rank-2 randomised-SVD
approximation of the restoration, projected onto that kernel). The gains are
read from the RMSNorm after $W_O$ and $W_{\mathrm{down}}$: $1+\mathbf{w}$ for
Gemma-2 (and Gemma-3), whose RMSNorm stores an offset weight $\mathbf{w}$, and
the raw weight $\mathbf{w}$ for Gemma-4, OLMo-2 and Olmo-3. Each gain vector
is checked against the module's own forward pass, and the fork's unit tests
check that the deployed weight satisfies
$\mathbf{b}^{\top}W'=(1-\lambda)\,\mathbf{b}^{\top}W$ to floating-point
precision. On pre-norm models the modifier reduces to MPOA. The post-norm
component along $\hat{\mathbf{r}}$ was not measured in forward passes of the
selected models.

\paragraph{SOM.} A port of the proof of concept in Heretic PR~\#196
\cite{piras2025som}. Per layer, a $4\times4$ hexagonal self-organising map is
fitted to the per-prompt harmful residuals (10{,}000 online iterations,
learning rate 0.01 and neighbourhood radius 0.5, both decayed as
$1/(1+2t/T)$; seeded with the run seed). The $k=4$ neurons that win the most
harmful residuals yield directions
$\mathrm{normalize}(\text{prototype}-\boldsymbol{\mu}_{\mathrm{safe}})$, which
are projected orthogonal to the harmless mean and Gram--Schmidt orthonormalised
in rank order. Direction $i$ is ablated with strength $\lambda(\ell)\,\delta^{i}$
($\lambda$ as above, $\delta$ a searched decay: $\delta=0$ is single-direction
ablation, $\delta=1$ removes the $k$-dimensional subspace), with full row
normalisation in a rank-6 adapter. A global scope uses one integer layer.
Deviations from Piras et al.~\cite{piras2025som} are listed in
\S\ref{sec:setup-deviations}.

\paragraph{ARA.} Heretic's default modifier \cite{weidmann2026ara}. For each
targeted module in a searched layer range, inputs and outputs at the last
prompt position are recorded on the harmless and harmful estimation prompts,
and a rank-50 adapter is fitted by L-BFGS (5 outer steps of at most 20
iterations, history 10, learning rate 1.0, strong-Wolfe line search) to the
objective of \S\ref{sec:m-ara}, with $A$ initialised from the run seed.
Row-magnitude preservation enters only the objective; the exported adapter is
not renormalised.

\paragraph{J-space (small pair only).} Directions follow the
\texttt{jspace-refusal} code
\cite{sekinal2026jspace,neuronpedia2026jacobian,gurnee2026verbalizable}:
$\mathbf{d}_\ell=J_\ell^{\top}\big(\mathbf{g}\odot(\overline{W_U[\mathcal{R}]}-\overline{W_U})\big)$,
unit-normalised, with $W_U$ the unembedding, $\mathcal{R}$ the token ids of the
package's 29 refusal words (34 ids), $\mathbf{g}$ the final-norm gain probed by
normalising an all-ones vector, and $J_\ell$ the WikiText-fitted Jacobian lens
of layer $\ell$ (\texttt{neuronpedia/jacobian-lens}, revision
\texttt{qwen-n1000}); the lenses cover all blocks but the last, where $J=I$. No
harmful/harmless contrast set is used. The \texttt{PrecomputedDirections}
modifier then applies these directions with DiM/MPOA's edit and search space
unchanged, without harmless-mean projection and with full row normalisation
(rank 3). J-space is thus weight-baked here, not the forward-hook intervention
of the original code.

\begin{table*}[t]
\centering\small
\caption{Search spaces sampled by multivariate TPE. $L$ = number of layers;
intervals are closed; ``$\to0$'' marks a negative lower bound whose draws are
clamped to 0, giving positive probability to switching the term off. Kernel
parameters are sampled separately for the attention and MLP components (10
parameters for DiM, MPOA and J-space, 11 for SOM, 6 for ARA).}
\label{tab:search}
\begin{tabular}{lll}
\toprule
Parameter & Range & Methods \\
\midrule
Direction scope & \{global, per layer\} & DiM, MPOA, J-space, SOM \\
Direction index (global scope) & $[0.4(L-1),\,0.9(L-1)]$, continuous, interpolated & DiM, MPOA, J-space \\
                               & integers in $[\mathrm{round}(0.4(L-1)),\,\mathrm{round}(0.9(L-1))]$ & SOM \\
Direction decay $\delta$ & $[-0.2,\,1.0]$, $\to0$ & SOM \\
Max.\ weight, attention & $[0.8,\,1.5]$ & DiM, MPOA, J-space, SOM \\
Max.\ weight, MLP & $[-0.25,\,1.5]$, $\to0$ & DiM, MPOA, J-space, SOM \\
Max.-weight position (layer) & $[0.6(L-1),\,L-1]$ & DiM, MPOA, J-space, SOM \\
Min.\ weight (fraction of max.) & $[0,\,1]$ & DiM, MPOA, J-space, SOM \\
Min.-weight distance (layers) & $[1,\,\max(0.6(L-1),1)]$ & DiM, MPOA, J-space, SOM \\
\midrule
Start layer & integers in $[0,\,\lfloor L/2\rfloor]$ & ARA \\
End layer (exclusive) & integers in $[\lfloor L/2\rfloor+1,\,L]$ & ARA \\
$w_p$ (preserve harmless outputs) & $[0,\,1]$ & ARA \\
$w_s$ (steer harmful outputs) & $[10^{-4},\,1]$, log scale & ARA \\
$w_o$ (overcorrection, relative) & $[0,\,1.3]$ & ARA \\
$k$ (nearest neighbours) & integers in $[1,\,15]$ & ARA \\
\bottomrule
\end{tabular}
\end{table*}

{\raggedright
\paragraph{Additional conditions.} Three further conditions were run: a
$\boldsymbol{\gamma}$-corrected MPOA control on the four post-norm models; a
second seed (seed~2) for all five methods on the small pair (10 runs), in
place of the planned second seed for DiM, MPOA and ARA on the Qwen and Gemma
pairs; and eight publicly released community abliterations as external
references under the same evaluation protocol
(\texttt{huihui-ai/Qwen2.5-32B-Instruct-abliterated},
\texttt{zetasepic/Qwen2.5-32B-Instruct-abliterated-v2},
\texttt{heretic-org/Qwen3.8-27B-heretic-ara},
\texttt{huihui-ai/Huihui-Qwen3.8-27B-abliterated},
\texttt{coder3101/gemma-4-31B-it-heretic},
\texttt{MuXodious/Olmo-3.1-32B-Instruct-impotent-heresy},
\texttt{huihui-ai/Huihui-Mistral-Small-3.2-24B-Instruct-2506-abliterated} and
\texttt{llmfan46/Mistral-Small-3.2-24B-Instruct-2506-ultra-uncensored-heretic}).
The external OBLITERATUS toolkit \cite{obliteratus2026} is not part of this
study. Its v0.1.4 presets \texttt{basic}, \texttt{advanced} and
\texttt{spectral\_cascade} (commit \texttt{461ff086}) were run on
Qwen2.5-7B-Instruct and scored with this study's harness; the results are
reported, with other external tools, in the companion paper
\cite{young2026paper30}. Version 0.1.4 lists a \texttt{som} preset but does not
ship its direction module and does not support the hybrid-attention Qwen3.5-4B, so
it could provide no older--newer contrast. A paper-faithful SOM configuration
and J-space as a forward hook with $\alpha=1$ were planned on the small pair
but not run; they are left to future work (\S\ref{sec:limitations}).\par}

\subsection{Optimisation, Prompt Splits and Selection}
\label{app:setup-opt}

\paragraph{Sampler, seeding and resumption.} The TPE sampler uses 128
expected-improvement candidates. Seed~1 is applied to Python, NumPy, PyTorch
and Optuna and to the method-internal random steps (randomised SVD, SOM, ARA
initialisation). In resumed processes (Qwen3.8-27B MPOA and the
interrupted Qwen3.8-27B ARA study, \S\ref{sec:setup-deviations}) the stored seed
(1) was restored for the sampler and the method-internal steps, but the global
Python, NumPy and PyTorch generators were seeded randomly, because the resume
commands did not pass \texttt{-{}-seed}. Heretic chose the generation batch size
automatically (upper bound 128); it was 128 in every run except Qwen2.5-32B
DiM, capped at 64, and the seed-2 Qwen3.5-4B DiM, SOM and ARA runs on the
workstation, where batch size 128 ran out of memory and 64 was chosen, so
those three seed pairs differ in batch size as well as seed (sensitivity of the
seed-noise band in \S\ref{sec:res-h1} and Appendix~\ref{app:kl-cap}).
Journals, configurations and logs are kept for every run.

\paragraph{Prompt splits.} The refusal objective uses the first 100
\emph{test} prompts of \texttt{harmful\_behaviors}, which serve as the
development (dev) set for both search and selection, and the KL objective the
first 100 test prompts of \texttt{harmless\_alpaca}. Batch-size probing and
response-prefix detection use training prompts only. Dataset revisions were
not pinned in the Heretic configurations. All primary and secondary refusal,
capability and teacher-forced outcomes come from the held-out sets of
\S\ref{sec:setup-models} and \S\ref{sec:setup-eval}. Dev-set values are reported only to document selection
(Pareto fronts, KL-threshold sensitivity), as the first-token KL measure of
distribution shift, and for the dev-versus-held-out overfitting check
(\S\ref{sec:setup-stats}).

\paragraph{Objective details.} Heretic's refusal markers are matched
case-insensitively, after removing emphasis and normalising apostrophes and
whitespace; empty responses count as refusals. Prompts use each model's chat
template with the system prompt ``You are a helpful assistant.''. Heretic
determines a response prefix automatically and prefills it when one is found:
if the chat template opens a reasoning block, the prefix closes it; otherwise
Heretic searches for a prefix common to all responses to the detection prompts.
A prefix was found only for Qwen3.8-27B and Qwen3.5-4B, whose templates open a
reasoning block, so their dev-set responses contain no reasoning; no prefix was
found for any other model.

\paragraph{Selection.} Heretic's headless default exports the trial with the
fewest refusals irrespective of KL divergence; the fixed rule of
\S\ref{sec:setup-opt} is applied instead in every run. The selected trial
always lies on the Pareto front.

\subsection{Held-out Evaluation}
\label{app:setup-eval}

\paragraph{Held-out refusal set.} The held-out file is version~2 of
\texttt{refusal\_sets.json}; its HarmBench items are the standard category of
the HarmBench text test set \cite{mazeika2024harmbench}, and its 200
malicious-code requests come from the code-as-weapon bank \cite{young2026code}.
The code items come from the bank's \texttt{code\_safety} split, restricted to
unanimous labeller agreement (tiers 5/5, 4/4), excluding items sourced from
\texttt{harmful\_behaviors} or JailbreakBench, and drawn proportionally by
source with seed 20261105 (\texttt{astra} 88, \texttt{scam2prompt} 43,
\texttt{rmcbench} 24, \texttt{cysecbench} 23, \texttt{malwarebench} 17,
\texttt{redcode} 5). Each harmful item was compared with all 520
\texttt{harmful\_behaviors} prompts (exact match after normalisation, or token
Jaccard $\ge0.6$ with \texttt{SequenceMatcher} ratio $\ge0.8$); 44 items are
flagged and results are reported with and without them. In a check made after
evaluation with the same criterion, none of the 350 benign items matched any
of the 31{,}323 \texttt{harmless\_alpaca} prompts (training and test splits).
Dataset revisions and file hashes are stored in the file's metadata.

\paragraph{Generation.} Where the chat template lacks a system role, the system
prompt is folded into the user turn; reasoning is disabled with
\texttt{enable\_thinking=False}. A copy of each response truncated to 256
tokens and the finish reason are stored.

\paragraph{Teacher-forced divergence.} The original model's responses are
scored under teacher forcing from vLLM prompt log-probabilities; the measure is
a sequence-level complement to the first-token KL objective.

\paragraph{Capability benchmarks.} Subsets were drawn
once with seed 20261104 (MMLU-Pro allocated proportionally over its 14
subjects); their document identifiers are re-verified before every run.
Decoding is greedy with fixed budgets, chat templates are applied (few-shot
examples as multi-turn chat), reasoning is disabled and the context is 8{,}192
tokens; harness seeds are 0 (Python) and 1234 (NumPy, PyTorch, few-shot
sampling, vLLM). A Transformers backend is reserved for models vLLM cannot load;
all evaluations so far used vLLM. Each summary records library versions and
a SHA-256 over the evaluated model's files.

\paragraph{Judges and degeneracy heuristic.} All judges score the same stored
responses: WildGuard \cite{han2024wildguard} (\texttt{allenai/wildguard};
primary judge of refusal and harmful response), the HarmBench classifier
(\texttt{cais/HarmBench-Llama-2-13b-cls}; harmful compliance on harmful prompts),
Qwen3Guard-Gen-8B \cite{zhao2025qwen3guard} (second LLM judge; safety and refusal
labels), each with its official prompt format; the ModernBERT refusal
classifier of Heretic's BERT scorer (\texttt{VINAY-UMRETHE/Refusal-Classifier};
prompt and response); the RoBERTa classifier
\texttt{Human-CentricAI/LLM-Refusal-Classifier} (response only), kept for
continuity with Young~\cite{young2025comparative}; and Heretic's 33 keyword markers
(lower-cased, typographic apostrophes normalised; unlike Heretic's scorer, no
emphasis or whitespace normalisation, and empty responses are flagged as
degenerate rather than counted as refusals) on the full response and its first
256 tokens. LLM judges run greedily in vLLM
with a 4{,}096-token context, except the HarmBench classifier, which runs with
its 2{,}048-token context on the response cut to its first 512 classifier
tokens; for 6 long code-misuse prompts the behaviour text was also shortened to
fit (436 of 97{,}828 judged responses). BERT inputs are truncated to 512 tokens.
A degeneracy heuristic flags empty responses; responses of at least 20 words in
which more than half of the word 4-grams repeat or fewer than 15\,\% of words
are distinct; runs of more than 40 identical characters other than whitespace
and the separator characters used for rules in code and Markdown
(\texttt{= - * \# \_ \textasciitilde{} .}), whose runs count only when they make up
more than half of the response; and more than 20\,\%
non-printable characters. In the analysis a character-run flag is kept only if
the run reaches the end of the response, if a run of a character other than
these separators and box-drawing characters lies outside Markdown code, or if
a separator or box-drawing run makes up more than half of the response;
otherwise only the non-printable test applies. This recheck cleared 119 of the
162 raw character-run flags, which marked mainly hexadecimal constants, a long
string copied from one prompt and box-drawing rules in coherent code. The
outcome classes of \S\ref{sec:setup-eval} are assigned to all 97{,}828
judged responses; WildGuard returned no refusal label for 5 of them, which are
left unscored and excluded from WildGuard rates. Each judge was loaded at an
explicit Hugging Face commit, resolved from the repository's main branch when
scoring started, passed to the loader and stored with the scores. Both scoring
batches resolved the same commits, so all 74 evaluated models were scored with
\texttt{allenai/wildguard} at \texttt{cbba4823},
\texttt{cais/HarmBench-Llama-2-13b-cls} at \texttt{bda70534},
\texttt{Qwen/Qwen3Guard-Gen-8B} at \texttt{4505cb1a},
\texttt{VINAY-UMRETHE/Refusal-Classifier} at \texttt{5c525175} and
\texttt{Human-CentricAI/LLM-Refusal-Classifier} at \texttt{0d9379ea}.

\subsection{Export and Release}
\label{sec:setup-export}

Every evaluated model is a merged, standalone checkpoint, never a live adapter.
The selected trial is restored from a copy of the study journal patched with
the trial's Pareto position. Heretic reloads the base model,
recomputes the method's statistics (J-space reads its direction file), rebuilds
the adapter from the trial's stored parameters and merges it with PEFT
\texttt{merge\_and\_unload}. The exported checkpoint differs from the original
only in the attention (or linear-attention) output and MLP down-projection
matrices of the layers the trial edits, each replaced by $W+BA$ with
$\mathrm{rank}(BA)\le1$ (DiM), 3 (MPOA, J-space), 6 (SOM) or 50 (ARA); all
other weight tensors and the tokenizer are the base model's. Main-study
re-exports ran on the GPU that then evaluated the model
(\S\ref{sec:setup-compute}): the RTX box for the Qwen and Gemma pairs and the
H200 box for the OLMo and Mistral Small pairs. All main-study searches ran on
the H200 box, so the Qwen and Gemma exports were rebuilt on a different GPU
from their search. For ARA, rebuilding the adapter means refitting it by
L-BFGS from the stored hyperparameters, and the ARA exports rebuilt on the
other GPU drifted more on the dev re-score than those rebuilt on the search
GPU (Appendix~\ref{app:dev-rescore}).
The export step itself does not re-score the model; the released merged
exports were therefore re-scored separately on the dev set and compared with
the selected trial's dev refusal count and KL divergence
(Appendix~\ref{app:dev-rescore}).

Models are published with Heretic's upload action as repositories
\texttt{richardyoung/\{base\}-\{METHOD\}-heretic} with Heretic's model card;
the five Gemma-2-27B repositories are private (withheld), the five OLMo-2-32B
repositories are gated with automatic approval, and all others are public
(Ethics Statement).
For MPOA, ARA and DiM the upload includes Heretic's full \texttt{reproduce/}
manifest (configuration, study journal, requirements and weight checksums),
which upstream Heretic can use to re-run the search and export
(\texttt{heretic -{}-reproduce}); the manifest itself notes that reproducibility
is not guaranteed because Heretic was installed from a local directory or
wheel. SOM, J-space and $\gamma$-MPOA models omit it, because they need a
project fork and, for J-space, an external direction file. The released model
cards name neither this paper nor the fork commits (\texttt{8fd4d76},
\texttt{fdfe54b}); all Heretic commits used are recorded in the run records
and environment captures (\S\ref{app:setup-methods}).
For the main study the evaluated copy is downloaded back from the published
repository, so evaluated and published weights coincide by construction. Small-pair models were
evaluated from a locally kept export (the workstation's Hugging Face token was
read-only during the runs) and published afterwards by re-exporting the same
trial; publication is accepted only if every uploaded weight shard is
SHA-256-identical to the benchmarked copy; this check passed for all ten
small-pair models.

Released models (all except the five withheld Gemma-2-27B variants) keep the
licence of their base model: Apache-2.0 (the Qwen,
OLMo and Mistral-Small-3.2 bases and Gemma-4-31B), to which the use
restrictions of the OpenRAIL++ licence are added; or, for
Mistral-Small-2409, the Mistral Research Licence, which permits
non-commercial research use. The five OLMo-2-32B variants are gated
on Hugging Face with automatic approval (Ethics Statement). A licence states
permitted use; it is not a technical control. The Human-CentricAI judge
(CC~BY-NC) is used for research only.

Whether each licence permits redistributing a derivative whose refusal has
been removed was checked against the licence texts, retrieved on 8 November
2026. Apache-2.0 (Section~4) permits redistribution of modified works with a
copy of the licence and notices of modification and contains no use
restriction; the OLMo model cards add that the models are ``intended for
research and educational use'' under Ai2's Responsible Use Guidelines, which
``do not restrict the use of the Ai2 Tools for the sole purpose of scientific
research and development''. The Mistral Research Licence~0.1 permits the
distribution of derivatives made by the licensee (Section~2.3) provided they
carry ``prominent notices stating that You modified the concerned Mistral
Model'', and limits the use of derivatives to research purposes
(Section~3.2); it contains no provision on safety behaviour, so it permits
the release for non-commercial research use. The Gemma Terms of Use permit
the distribution of model derivatives (Section~3.1) subject to the use
restrictions of Section~3.2, which incorporate the Gemma Prohibited Use
Policy; that policy forbids using Gemma or model derivatives for ``attempts to
override or circumvent safety filters or intentionally drive Gemma or Model
Derivatives to act in a manner that contravenes this Gemma Prohibited Use
Policy''. The terms do not define ``safety filters'', and on a plain reading
the removal of trained refusal may fall under this clause, so this study
cannot state that redistributing the Gemma-2-27B derivatives is permitted.
The five Gemma-2-27B variants were therefore made private on 9 November 2026
and are not released; the ``Notice'' text file that Section~3.1 requires with
every distribution was added to their repositories.

\subsection{Multiple-Testing Families}
\label{app:setup-stats}

Holm's procedure controls each family of related tests at a two-sided
$\alpha=0.05$, separately for each refusal measure and prompt subset. The
families are H1 (23 tests: 21 pair $\times$ method McNemar tests and the two
pooled class effects), H2 (6: five per-pair interactions and the pooled one),
modified versus original (42 models), the H3 gap model (4 terms), per-model
keyword--WildGuard gaps (all 74 evaluated models), $\gamma$-MPOA versus MPOA
(4), each community release versus the four models selected here for the same
checkpoint (4 per release), seed~2 versus seed~1 (10 cells), per-benchmark
capability changes of the $\gamma$-MPOA models (20 per reference model),
over-refusal changes from the original (all 64 modified models), the
256-token label change of each judge and the 256-token keyword--judge gaps
(all 74 evaluated models each; Appendix~\ref{app:trunc256}), the two
dev re-score tests (refusals and KL; Appendix~\ref{app:dev-rescore}), the
100-token keyword--judge gaps (all 74 models; \S\ref{sec:res-h3}), and, in the
sensitivity analyses, the H1 and H2 families repeated within each item subset
(Appendix~\ref{app:item-source}) and within each harmful-compliance outcome
(Appendix~\ref{app:harm}). Pair-level random-effects summaries, the
heterogeneity tests, the KL-cap refits and the export-drift Fisher tests are
not adjusted.

\subsection{Compute and Software}
\label{sec:setup-compute}

Three single-GPU machines were used (Table~\ref{tab:compute}); the two server
boxes were rented from Massed Compute and downloaded models directly from
Hugging Face. The Heretic environment used Python 3.12 (3.12.16 on the boxes,
3.12.15 on the workstation), PyTorch 2.14.1 (CUDA 13.0 build), Transformers
5.18.0, PEFT 0.21.2, Accelerate 1.15.0, Optuna 5.0.0, Datasets 5.0.1,
safetensors 0.8.0, huggingface\_hub 1.33.0 and NumPy 2.5.4. The evaluation
environment is a separate lock-file-pinned project with vLLM 0.33.0, lm-eval
0.4.13, PyTorch 2.14.0 (CUDA 13.0), Transformers 5.20.2, Datasets 5.1.0 and
Accelerate 1.15.0; vLLM's FlashInfer sampler was disabled (identical argmax
under greedy decoding). The 168 recorded single-GPU jobs ran for 69.3\,h of
wall-clock time in total: 32.7\,h of Heretic optimisation (73 processes,
including eight pilot runs), 33.8\,h of evaluation (90 jobs; for selected
models these include selection, re-export, upload and download where
applicable) and 2.9\,h of judging (5 jobs). The totals include 6.1\,h of failed
or aborted attempts and exclude time spent waiting for a busy GPU. Because two
job queues shared the H200 for part of the search phase
(\S\ref{sec:setup-deviations}), the GPUs were occupied for 67.5\,h (H200
30.4\,h, RTX PRO 6000 22.6\,h, RTX 4090 14.5\,h). Not included are model
downloads, work without a run record (smoke tests, the computation of the
J-space directions, and the re-export and upload that published the
small-pair models), evaluations run on the RTX box for another study, and the
statistical analysis. Per completed 100-trial main-study search on the H200
box, a direction-based study took a median of 18.1 minutes (range 10.2--61.5;
23 studies) and an ARA study 81.0 minutes (50.3--95.6; 8 studies), 18.1\,h in
total; the resumed Qwen3.8-27B MPOA study, split over three processes, is not
included.

\begin{table*}[t]
\centering\small
\caption{Hardware (GPU memory from \texttt{nvidia-smi}, RAM from \texttt{free -g}).}
\label{tab:compute}
\begin{tabular}{@{}l >{\raggedright\arraybackslash}p{3.3cm} >{\raggedright\arraybackslash}p{2.8cm} l >{\raggedright\arraybackslash}p{4.6cm}@{}}
\toprule
Machine & GPU (memory) & CPU / RAM & Driver & Role \\
\midrule
H200 box & 1$\times$ NVIDIA H200 NVL (143{,}771 MiB) & 16 vCPU, AMD EPYC 9355 / 176 GiB & 580.126.09 & main-study search; export and evaluation of the OLMo and Mistral Small pairs (incl.\ OLMo $\gamma$-MPOA) and 3 community models; ARA/MPOA/DiM pilots \\
RTX box & 1$\times$ RTX PRO 6000 Blackwell Server Ed.\ (97{,}887 MiB) & 16 vCPU, AMD EPYC 9335 / 141 GiB & 580.126.09 & export and evaluation of the Qwen and Gemma pairs (incl.\ Gemma $\gamma$-MPOA) and 5 community models; all judging; release \\
Workstation & 1$\times$ GeForce RTX 4090 (24{,}564 MiB) & AMD Ryzen 9 7950X / 62 GiB & 615.78.08 & small pair: all steps; SOM and J-space pilots \\
\bottomrule
\end{tabular}
\end{table*}

\subsection{Deviations from the Plan}
\label{sec:setup-deviations}

\begin{itemize}
\item \textbf{Mistral-Small-3.2 source.} The official repository ships only
  \texttt{mistral-common} tokenizer files, without Hugging Face tokenizer or
  processor configurations, so Heretic failed at load before any trial. The runs
  were repeated from \texttt{unsloth/Mistral-Small-3.2-24B-Instruct-2506} pinned
  at commit \texttt{3ed8d341b53ea3d0e4194689905477688a6cf733}, whose ten weight
  shards are SHA-256-identical to the official ones (\texttt{config.json}
  differs only in metadata). The Hugging Face tokenizer has a known
  pre-tokenisation regular-expression defect (Transformers suggests
  \texttt{fix\_mistral\_regex=True}, which Heretic does not pass) that changes
  the tokenisation of 3 of 33{,}165 study prompts; the same tokenizer is used
  for original and modified models in Heretic and vLLM.
\item \textbf{SOM relative to Piras et al.~\cite{piras2025som}.} The port ranks neurons by
  wins over the harmful residuals (the proof of concept counted wins over the
  map's own prototypes, reducing ``top-$k$'' to the first $k$ grid cells),
  orthonormalises rather than sums the directions, uses one geometric decay
  instead of per-direction strengths and a single integer layer for the global
  scope, keeps the proof-of-concept SOM hyperparameters (not confirmed against
  the paper), and uses a rank-6 adapter not compared with higher ranks; hence
  ``Heretic-tuned SOM''.
\item \textbf{Concurrent jobs on the H200.} Until an idle-GPU guard was added, a
  second job queue shared the GPU, causing out-of-memory failures. Qwen2.5-32B
  MPOA completed 100 trials with part of the model offloaded to CPU. Qwen3.8-27B
  MPOA failed twice (after 5 and 6 trials) and was resumed from the second
  attempt's journal; the evaluated study has 99 completed trials and one failed
  trial. The first Qwen2.5-32B ARA run failed at its third trial with a
  meta-device error under CPU offloading and was repeated from scratch (100
  trials; evaluated).
\item \textbf{Qwen3.8-27B ARA rerun.} The first Qwen3.8-27B ARA study was
  interrupted by the same cause after 66 completed trials; one resumption
  failed and a second completed it (99 completed, one failed). On resumption
  Optuna re-creates the sampler from the same seed, so later proposals differ
  from those of an uninterrupted run. Because the study was not one
  uninterrupted optimisation, and because its dev-set result (59/100 keyword
  refusals at KL $\le 0.1$) differed from the 0/100 at KL 0.054 reported for
  \texttt{heretic-org/Qwen3.8-27B-heretic-ara} (same defaults, 60 trials), it
  was repeated from scratch with identical
  configuration and seed and no interruption (the ``clean'' run, 100 trials).
  The clean run is evaluated; the interrupted study is retained. No such repeat
  was run for Qwen3.8-27B MPOA. Instead, its journal was checked against the
  seeded start-up sequence. The study was resumed after 7 recorded trials (6
  complete, 1 failed). On resumption the sampler restarted the seed-1 start-up
  sequence, so trials 7--29 repeat the first 23 start-up draws: all 30 start-up
  trials have the same parameters, position for position, as the uninterrupted
  seed-1 Olmo-3.1-32B MPOA study, and the 6 trials evaluated twice gave dev
  keyword refusals within 1 of 100 of each other. The selected trial (number 29,
  25/100 at KL 0.085) is one of these repeated start-up draws, so an
  uninterrupted seed-1 study would have evaluated the same parameters (as its
  trial 22) and, under the selection rule, selected a trial with at most about
  25/100 dev refusals. The resumption can therefore have raised, but not lowered,
  the dev refusal rate of the selected Qwen3.8-27B MPOA trial; its effect on
  held-out refusal was not measured. For the one study run both ways
  (Qwen3.8-27B ARA, seed 1), the resumed and clean selections differed by 4 of
  100 dev keyword refusals (59 and 63).
\item \textbf{J-space directions.} Before the analysed runs the export was
  corrected (probed rather than stored final-norm gain; $J=I$ for the last
  block) and the directions regenerated; the J-space pilot used the earlier
  directions.
\item \textbf{Evaluation-harness corrections.} (i) lm-eval 0.4.13's
  \texttt{mbpp\_instruct} extractor dropped the first word of every program,
  giving pass@1 of zero; it was patched and earlier MBPP scores were recomputed
  from logged samples. (ii) vLLM 0.33's detokeniser left the meta-space
  character U+2581 in Gemma-2 outputs; it is mapped to a space in every
  evaluation run after the fix, and the affected Gemma-2 evaluation was
  repeated; the earlier Qwen2.5-32B base evaluation contains no U+2581
  characters. (iii) The first Qwen3.8-27B
  baseline evaluation fell back silently to the Transformers backend (vLLM
  rejected the default concurrency for this hybrid model); it was repeated in
  vLLM with \texttt{max\_num\_seqs = 256}, as used for all Qwen3.8/Qwen3.5
  models. Other model-specific vLLM
  settings: Gemma-4 image inputs disabled; Mistral with Hugging Face tokenizer
  and configuration formats.
\item \textbf{Mixed model.} The planned mixed model was to be fitted with lme4;
  because R was not available it was fitted by variational Bayes in
  statsmodels (\texttt{BinomialBayesMixedGLM}), which gives no $p$ values, and
  the pooled tests use the GEE (\S\ref{sec:setup-stats}).
\item \textbf{Status.} All 42 primary runs, the four $\gamma$-MPOA controls
  and the ten seed-2 runs were completed, selected, exported and evaluated, as
  were the ten original models and the eight community releases, and all 74
  evaluated models were scored by every judge on all 1{,}322 held-out
  prompts. The three LLM judges were also applied to the 256-token cut of
  every response (Appendix~\ref{app:trunc256}), and the 46 released merged
  exports (42 primary, four $\gamma$-MPOA) were re-scored on the dev set
  (\S\ref{sec:setup-export}, Appendix~\ref{app:dev-rescore}). The
  paper-faithful SOM configuration and the J-space forward hook
  (\S\ref{app:setup-methods}) were not run and are left to future work.
\end{itemize}

%% file: sections/08c_validity.tex
\section{Threats to Validity: Full List}
\label{app:validity}

This appendix lists every identified limitation and threat to validity;
Section~\ref{sec:limitations} summarises the most consequential of them. Where
it can be stated, each limitation is given with its likely direction of bias
and its mitigation.

\paragraph{Single seed and run-to-run variance.}
Each main-study cell is optimised with a single seed (seed~1); the
Qwen3.8-27B ARA cell was run twice with seed~1 (see below). TPE search is stochastic
(Heretic~\#88), and one ARA trial has been reported to yield 52 and 11 refusals
on two machines (Heretic~\#499); both are unreviewed practitioner reports,
retrieved in November 2026. Here, two seed-1 ARA runs on Qwen3.8-27B (one
interrupted and resumed, one clean) selected trials with 59 and 63 of 100
keyword refusals at KL~$\le 0.1$, whereas a community model with the same
defaults and 60 trials (\texttt{heretic-org/Qwen3.8-27B-heretic-ara}) reports
0 of 100 at KL~0.054 on its model card, an unreviewed figure. The evaluated Qwen3.8-27B MPOA run was resumed after an
out-of-memory failure and completed 99 of 100 trials. Seed and configuration
variance has altered conclusions about algorithm comparisons
\cite{bouthillier2021accounting} and about the durability of open-weight
safeguards \cite{qi2025durability}. Such variance can create
or mask a generation effect in a single cell. On the small pair, all five
methods were optimised again with seed~2 (10 runs, in place of the planned
second seed for DiM, MPOA and ARA on the Qwen and Gemma pairs): held-out
refusal differed between seeds in all 10 cells (median 9.4~pp, maximum
25.2~pp), and the ARA generation effect fell from +14.2~pp to +0.0~pp, so
the small pair meets H1 with seed~1 only. Mitigation: full Pareto fronts and
four KL thresholds are reported, the seed-2 spread is reported next to the
generation effects, a run-to-run standard deviation derived from it (9.2~pp
[6.4, 16.1]) flags the 11 of 16 main-study generation effects that lie within
seed noise (Table~\ref{tab:h1-heldout}; this assumes that the seed variance of
the 4B and 7B models transfers to the 22--32B models), and method rankings were
more stable than single-cell generation effects ($\eta^2$ of method 0.94 and
0.88). Three of the ten seed-2 runs also used a generation batch size of 64
instead of 128 after running out of memory; without them the standard
deviation is 7.2~pp [4.8, 14.6] and the same five raw effects exceed the band
(baseline-adjusted: six instead of three). The heretic-org trial on
Qwen3.8-27B was not run in this pipeline, and a second seed for the
main-study cells is left to future work.

\paragraph{Unit of inference.}
The generational question has the model pair as its unit, and there are four
main-study pairs. Prompt-level intervals and tests cover prompt sampling
within a pair only; between-family heterogeneity was large (generation
$\times$ family $\chi^2=547.4$, 3 df), so pooled estimates are summaries of
per-pair effects of different sign. Mitigation: per-pair effects with seed
variance added are the primary evidence, pooled effects are reported with
pair-level random-effects (Hartung--Knapp--Sidik--Jonkman) and prediction
intervals, and leave-one-family-out fits are given (\S\ref{sec:res-h1},
Appendix~\ref{app:kl-cap}).

\paragraph{Selection on the development set.}
TPE optimises, and the selection rule selects, on the 100 prompts
(\texttt{mlabonne/harmful\_behaviors} test[:100]) that produce Heretic's refusal
count, and refusal directions (DiM, MPOA, SOM) and ARA's harmful-side
activations are estimated from \texttt{mlabonne/harmful\_behaviors} train[:400],
contrasted with \texttt{mlabonne/harmless\_alpaca} train[:400] (J-space
directions use no contrast set). Development-set refusal is therefore optimistic
(Heretic~\#481, \#503; unreviewed practitioner reports), possibly more so for methods with larger search spaces,
because optimising a noisy selection criterion overfits it and reusing the
selection data for evaluation biases the estimate \cite{cawley2010overfitting}.
Mitigation: reported refusal
comes from a frozen held-out set of 1,322 prompts (JailbreakBench harmful and
benign, 100 each; XSTest, 450; StrongREJECT, 313; HarmBench standard, 159;
code-as-weapon bank, 200 \cite{young2026code})
\cite{chao2024jailbreakbench,rottger2024xstest,souly2024strongreject,mazeika2024harmbench}; 44 items matching
the 520 AdvBench-derived prompts are flagged and results reported with and
without them; dev--test agreement is tested with Spearman's~$\rho$.

\paragraph{Keyword refusal.}
Heretic counts a refusal when any of 33 marker strings appears within 100
generated tokens. Reported keyword recall is about 0.48, against 0.77 for an LLM
judge (Heretic~\#419; see also PR~\#196; both unreviewed practitioner reports,
retrieved in November 2026); in a peer-reviewed, human-labelled meta-evaluation,
keyword matching reached Cohen's $\kappa=0.37$, against about 0.81 for
fine-tuned LLM judges \cite{xie2025sorrybench}. Because TPE minimises this count, methods are
rewarded for avoiding marker phrases rather than for complying, so selected
models were expected to retain more refusal than counted (H3). On the
held-out set, at the tool's 100-token window, keyword matching and WildGuard
agreed on average over the 42 ablated models (+0.5~pp [$-$1.0, +1.9]) but
differed significantly in both directions for 25 of them; on the full
response keyword matching was higher (+8.4~pp), because of markers after the
first 100 tokens (Table~\ref{tab:h3-windows}). Mitigation: keyword counts
drive optimisation and trial selection only; held-out refusal is reported
from WildGuard, with Qwen3Guard as a second judge and a degeneracy check, and
held-out keyword counts are reported only for comparison with the judges
(H3).

\paragraph{Post-norm leakage.}
Gemma-2/4, OLMo-2 and Olmo-3.1 apply an RMSNorm \cite{zhang2019rmsnorm} with gain
$\boldsymbol{\gamma}$ after $W_O$ and $W_{\mathrm{down}}$ (output normalisation
\cite{kim2025periln}). OLMo~2 and Olmo~3 normalise module outputs only
\cite{walsh2025olmo2,olmo2025olmo3}, whereas Gemma~2 normalises the input and
the output of each sublayer \cite{gemmateam2024gemma2}. After orthogonalisation the module output
satisfies $\hat{\mathbf{r}}^{\top}\mathbf{y}=0$, but the normalised output keeps
a component proportional to
$(\boldsymbol{\gamma}\odot\hat{\mathbf{r}})^{\top}\mathbf{y}$, handicapping
direction-based methods on four of the eight main-study models. Because both
generations of each affected family share this design, the leak mainly
penalises direction-based methods relative to ARA within these families; it
confounds the generation effect (H1) and its interaction with method class (H2)
to the extent that its size differs between generations, and it is also a
candidate mechanism for any generation effect.
Mitigation: per-family reporting and a $\boldsymbol{\gamma}$-corrected MPOA
control on the four post-norm models. $\gamma$-MPOA lowered held-out refusal
relative to MPOA on all four (by 3.1--19.9~pp), but the OLMo generation effect
remained +57.0~pp and the pooled H2 interaction $-$0.42. On Gemma-4-31B
(the largest reduction) and OLMo-2-32B, $\gamma$-MPOA also had a larger
teacher-forced divergence than MPOA and lost 21.4 and 8.1~pp of mean
capability, so these two reductions are confounded with a stronger edit.

\paragraph{J-space scope.}
Instruction-tuned Jacobian lenses exist only for Qwen3.5-4B and
Qwen2.5-7B-Instruct (\texttt{neuronpedia/jacobian-lens}, revision
\texttt{qwen-n1000}, fitted on WikiText; Gemma lenses are base-model lenses), so
J-space is limited to the small pair. The lens is fitted on non-chat text, and the
direction is baked into the weights through Heretic's ablation and TPE search
rather than applied as the reference forward hook; the direction of bias is
unknown. J-space as a forward hook ($\alpha=1$) was planned but not run and is
left to future work.

\paragraph{Method implementations.}
Heretic-tuned SOM (commit \texttt{45a4e6e}) deviates from Piras et al.~\cite{piras2025som}
and the proof of concept (Heretic~PR~\#196): neurons are ranked by wins over
harmful residuals (fixing a win-count bug); directions are Gram--Schmidt
orthonormalised; one geometric decay replaces per-direction strengths; global
scope uses a single integer layer; map settings follow the proof of concept;
and the norm-preserving adapter has rank~6, untested against higher ranks. The
narrower search may understate SOM; a paper-faithful SOM configuration was
planned but not run and is left to future work. MPOA's
row normalisation is approximated by a rank-3 LoRA, and ARA's deployed adapter
is not row-norm-preserving. All evaluated models are merged exports, not live adapters
(Heretic~\#490, an unreviewed practitioner report).

\paragraph{ARA instantiation.}
The released ARA model is rebuilt by refitting its adapter from the stored
hyperparameters. The four ARA exports refitted on a different GPU from the
search (Qwen and Gemma pairs) moved by more than two dev refusals on re-score,
against none of the four refitted on the search machine (Fisher $p=0.029$),
and the Gemma-4-31B ARA export re-scored at KL 0.107, above the cap. Two
seed-1 ARA runs on Qwen3.8-27B selected 59 and 63 of 100 dev refusals,
whereas heretic-org reports 0 of 100 at KL 0.054 with the same defaults; the
reported trial was not tested here. These points bear on H2, whose Qwen
reversal and Gemma support rest on single ARA cells. Mitigation: held-out
results were measured on the released exports, so they describe the released
models; the drift is reported per export (Appendix~\ref{app:dev-rescore}).

\paragraph{Mistral-Small-3.2 tokeniser.}
The official repository lacks Hugging Face tokeniser files, so the
\texttt{unsloth} mirror (commit \texttt{3ed8d34}) was used; its 10 weight shards
are SHA256-identical to the official ones. Its Hugging Face tokeniser has a
known pre-tokenisation regex defect (\texttt{transformers} suggests
\texttt{fix\_mistral\_regex=True}, which Heretic does not pass) that alters 3 of
33,165 study prompts. The same tokeniser is used for the base model, all
modified models and evaluation (the base evaluation recorded commit
\texttt{3ed8d34}), so within-model comparisons should be unaffected.

\paragraph{Truncation.}
Held-out responses are limited to 512 tokens; for the unmodified Qwen3.8-27B,
491 of 1,322 responses reached the limit, and on the 972 harmful prompts
11.2--35.3\% of the originals' responses and 17.6--87.8\% of the 42 primary
models' responses did so. Truncation should affect long, compliant answers
more than refusals, so judges may rate partial answers as incomplete or less
harmful, understating harmful compliance. Mitigation: finish reasons and the
first 256 tokens of every response are stored; keyword refusal is also
computed on the 256-token cut; compliance is split into complete and
incomplete compliance by finish reason (\S\ref{sec:setup-eval}); the
keyword--WildGuard gap is reported separately for truncated and complete
responses (\S\ref{sec:res-h3}); and the three LLM judges were also applied to
the 256-token cut of every response (Appendix~\ref{app:trunc256}).

\paragraph{Generation is confounded.}
Pairs differ in more than release date: size (the newer small-pair model is
smaller, and after abliteration harmful-code success has been found to be
limited by model size \cite{carleo2026willing}), architecture (Qwen3.8-27B and
Qwen3.5-4B use hybrid attention with gated-delta-rule linear-attention layers
\cite{yang2025gated}, whose recurrent state may carry refusal information
that residual-stream directions do not capture, and plain directional ablation has made a
recurrent model incoherent \cite{marshall2024refusal}; Olmo~3 adds
sliding-window attention on three of every four layers \cite{olmo2025olmo3};
Qwen3.8-27B, Gemma-4-31B, Mistral-Small-3.2 and Qwen3.5-4B are image-text checkpoints used with text only), training data and
post-training recipe, and the
interval between releases varies by family. Later checkpoints may also have
been safety-tuned on the public held-out benchmarks (JailbreakBench, XSTest,
StrongREJECT, HarmBench) or paraphrases of them, which would raise their
held-out refusal for reasons other than resistance to removal
\cite{sainz2023nlp}; the overlap check covers only
\texttt{harmful\_behaviors} and is lexical. The four public benchmarks first appeared
between August 2023 (XSTest) and April 2024 (JailbreakBench; HarmBench and
StrongREJECT in February 2024), before the earliest checkpoint
(Gemma-2-27B-it, June 2024; release months in Table~\ref{tab:models}), so
exposure was possible for every checkpoint, and the later checkpoint of each
pair appeared 9 to 23 months after the earlier one; the code-as-weapon bank
(May 2026) predates only Qwen3.8-27B (August 2026). A generation effect cannot be
attributed to safety training alone. The item mix also matters for the
originals' baseline: the code-as-weapon bank (200 items) carried 52\% of the
mean generation difference between the main-study originals, although the H1
and H2 conclusions were the same on the four public sets alone
(Appendix~\ref{app:item-source}). Mitigation: within-family comparisons,
family as a random effect, and conclusions phrased as statements about later
releases, not mechanisms.

\paragraph{Decoding.}
All refusal measurements, during search and on the held-out set, use greedy
decoding. Changing only decoding hyperparameters raised the misalignment rate
of 11 open models from 0\% to more than 95\% \cite{huang2024catastrophic}, so
residual refusal under greedy decoding may overstate refusal under sampling,
possibly by different amounts in each generation. Each prompt received one
greedy response, and reasoning was disabled, although the Qwen3.8-27B and
Qwen3.5-4B chat templates open a reasoning block and chain of thought can
carry refusal \cite{yang2026beyond}. Mitigation: decoding is
identical for every model, so within-pair comparisons are matched on decoding,
and the conclusions are stated for greedy, non-thinking generation; a
sampled-decoding sensitivity check is left to future work.

\paragraph{First-token KL.}
Distribution shift is constrained only through a common cap on the KL
divergence of the first generated token on 100 harmless prompts, which need
not reflect sequence-level change and may not be comparable across
vocabularies. The cap is common but not matched: it bound in 15 of the 24
main-study direction-based runs and in all 8 ARA runs, and 4 direction-based
runs never sampled a trial above it. Mitigation: selection at KL 0.05, 0.2 and
0.5, teacher-forced divergence on the original model's answers to the 350
benign held-out prompts, and H1 and H2 refitted by cap binding
(Appendix~\ref{app:kl-cap}).

\paragraph{Capability benchmarks.}
MMLU-Pro (1,000 items, stratified) and GSM8K (500 of 1,319) are subsets; IFEval
(541), HumanEval (164) and MBPP (500) are complete
\cite{wang2024mmlupro,cobbe2021training,zhou2023instruction,chen2021evaluating,austin2021program}. Thinking is disabled, so
reasoning-mode ability is not measured. Two harness defects were corrected: the
lm-eval~0.4.13 MBPP answer extractor (MBPP scores recomputed from logged
samples) and a vLLM~0.33 detokeniser artefact (U+2581 left in Gemma-2 output;
mapped to a space in every evaluation run after the fix, and the Gemma-2 base
evaluation rerun; the earlier Qwen2.5-32B base evaluation, run without the
mapping, contains no U+2581).

\paragraph{Outcome construct.}
Residual refusal is not attack success: of the ablated main-study responses
that WildGuard did not label as refusals, 75.1\% were harmful by the HarmBench
classifier and 55.8\% by WildGuard. Mitigation: H1 and H2 were repeated with
harmful compliance as the outcome, with the same pooled conclusions
(Appendix~\ref{app:harm}).

\paragraph{Judge validity.}
Reported refusal and harm rely on automated judges (WildGuard, the HarmBench
classifier, Qwen3Guard, two BERT-style classifiers)
\cite{han2024wildguard,mazeika2024harmbench,zhao2025qwen3guard}, applied to
all 74 evaluated models, with no human-labelled audit. Conclusions can depend
on the classifier \cite{young2026measuring}, modified-model outputs may be out
of distribution for the judges, and the BERT-style classifiers truncate inputs to
512 tokens. Against human labels, WildGuard and the HarmBench classifier reached
$\kappa$ of 0.61 and 0.53 \cite{xie2025sorrybench}, and rewriting harmful
responses as stories raised their false-negative rates from 0.02 to 0.12 and
from 0.12 to 0.36 \cite{eiras2025know}. Mitigation: several judges with official templates, each
loaded at a pinned and recorded Hugging Face commit (the same five commits for
all 74 models; Appendix~\ref{app:setup-eval}),
inter-judge agreement (Cohen's $\kappa$ for pairs \cite{cohen1960coefficient},
Fleiss' $\kappa$ across the panel \cite{fleiss1971measuring}, each with raw
agreement and marginals, because $\kappa$ is depressed when labels are
imbalanced \cite{feinstein1990high}), and a separate degeneracy class.
WildGuard and Qwen3Guard agreed at $\kappa=0.90$ on the originals and 0.82
after ablation, but agreement between LLM judges with correlated errors is
not confirmation \cite{hossain2026agreement}.

\paragraph{No preregistration.}
The study was not preregistered: hypotheses were written after pilot data, and
the protocol was revised after part of the main-study optimisation and the
capability benchmarks of one base model (Qwen2.5-32B-Instruct) had run, though
before any held-out response was generated.

\paragraph{Generalisation.}
Ten dense models from four families (Qwen contributing two pairs), 4--32B
parameters, English prompts and one Heretic~2.0 configuration with 100 trials
were studied; method rankings can change with the search budget
\cite{dodge2019show}, particularly between methods with search spaces as
different as rank-1 DiM and rank-50 ARA; 8 of the 42 core runs (19\%) still
improved after trial 75 (Figure~\ref{fig:search-convergence}). Results may
not transfer to mixture-of-experts or larger models, other languages,
reasoning mode or other implementations; eight public community releases of
five checkpoints were evaluated as external references
(Figure~\ref{fig:community}), and four of them reached lower held-out refusal
than the models selected here, at larger teacher-forced divergence. Three
presets of the external OBLITERATUS toolkit on Qwen2.5-7B-Instruct are
evaluated with the same harness in the companion paper \cite{young2026paper30}.

\paragraph{Dual use and ethics.}
Each released model (all except the five withheld Gemma-2-27B variants) is a
full merged checkpoint, published on Hugging Face under
the licence of its base model (Apache-2.0 with the added use restrictions of
the OpenRAIL++ licence, or the
Mistral Research Licence for Mistral-Small-2409; Gemma-4-31B is
Apache-2.0); a licence states permitted use and is not a
technical control. The five Gemma-2-27B variants are withheld because of the
Gemma Prohibited Use Policy; the five OLMo-2-32B variants are gated with
automatic approval, because no third-party public abliteration of OLMo-2-32B
was found; the others are public (Ethics Statement). Server-evaluated models are the copies downloaded back
from the published repositories; all ten RTX~4090 (small-pair) uploads are
verified SHA256-identical, shard by shard, to the benchmarked copies.
Reproduction manifests accompany only MPOA, ARA and DiM,
because SOM, J-space and $\gamma$-MPOA need a project fork. No harmful completions are
released; responses are stored privately for scoring. The study involved no
human participants and was not human-subjects research.

%% file: sections/08d_additional_results.tex
\section{Additional Results}
\label{app:additional-results}

\begin{figure*}[tp]
\centering
\includegraphics[width=\textwidth]{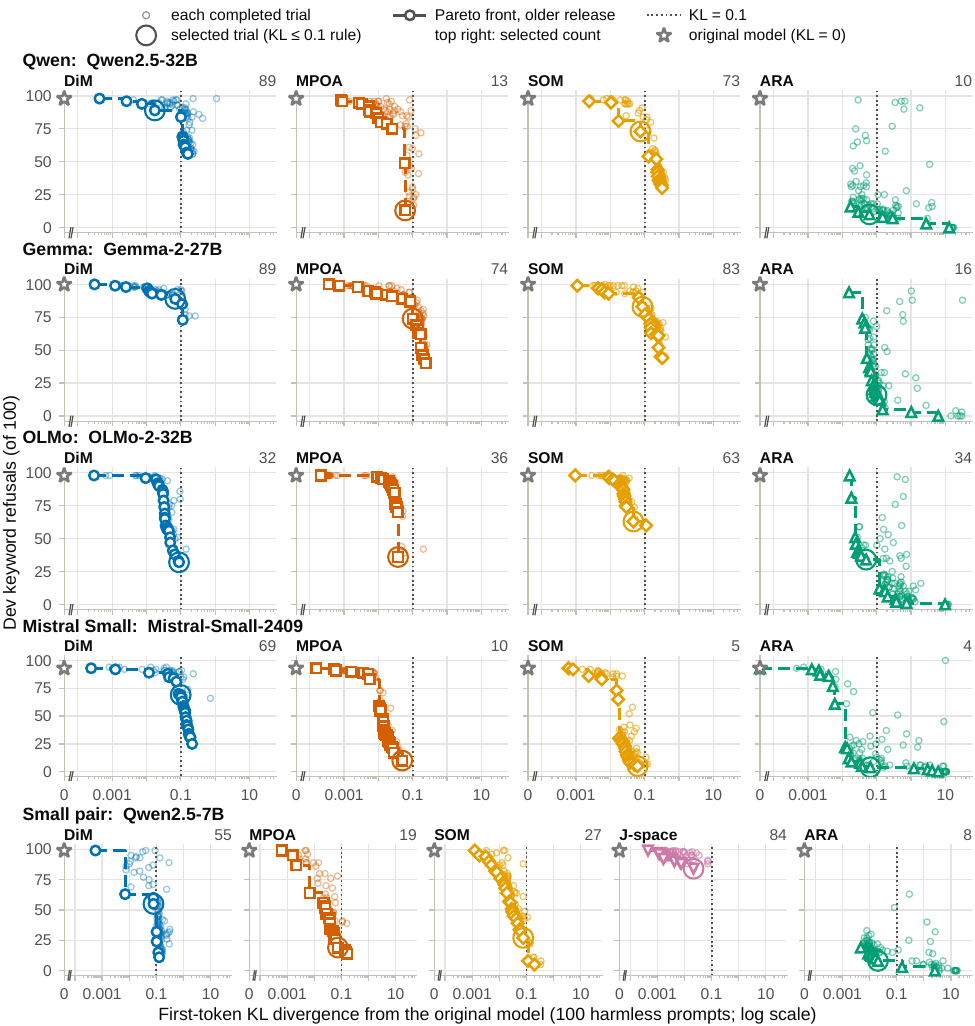}
\caption{\textbf{All completed TPE trials of each optimisation run, older releases (dev set).} One panel per run (model $\times$ method, seed 1), laid out as in Figure~\ref{fig:pareto}; the older model of each pair heads its row. Each dot is one completed trial; the line is the run's Pareto front (dashed, hollow markers), the ring the trial chosen by the selection rule (fewest dev keyword refusals at KL $\le$ 0.1; count at the top right of the panel) and the grey star the original model (KL = 0). The x axis is logarithmic, with KL = 0 drawn at the left edge behind an axis break; the dotted line is the selection budget KL = 0.1. 2100 trials from 21 runs; the newer releases are in Figure~\ref{fig:trials-new}. Refusal counts are keyword matches on the 100 dev prompts that were also used for optimisation and selection, not held-out refusal rates.}
\label{fig:trials-old}
\end{figure*}

\begin{figure*}[tp]
\centering
\includegraphics[width=\textwidth]{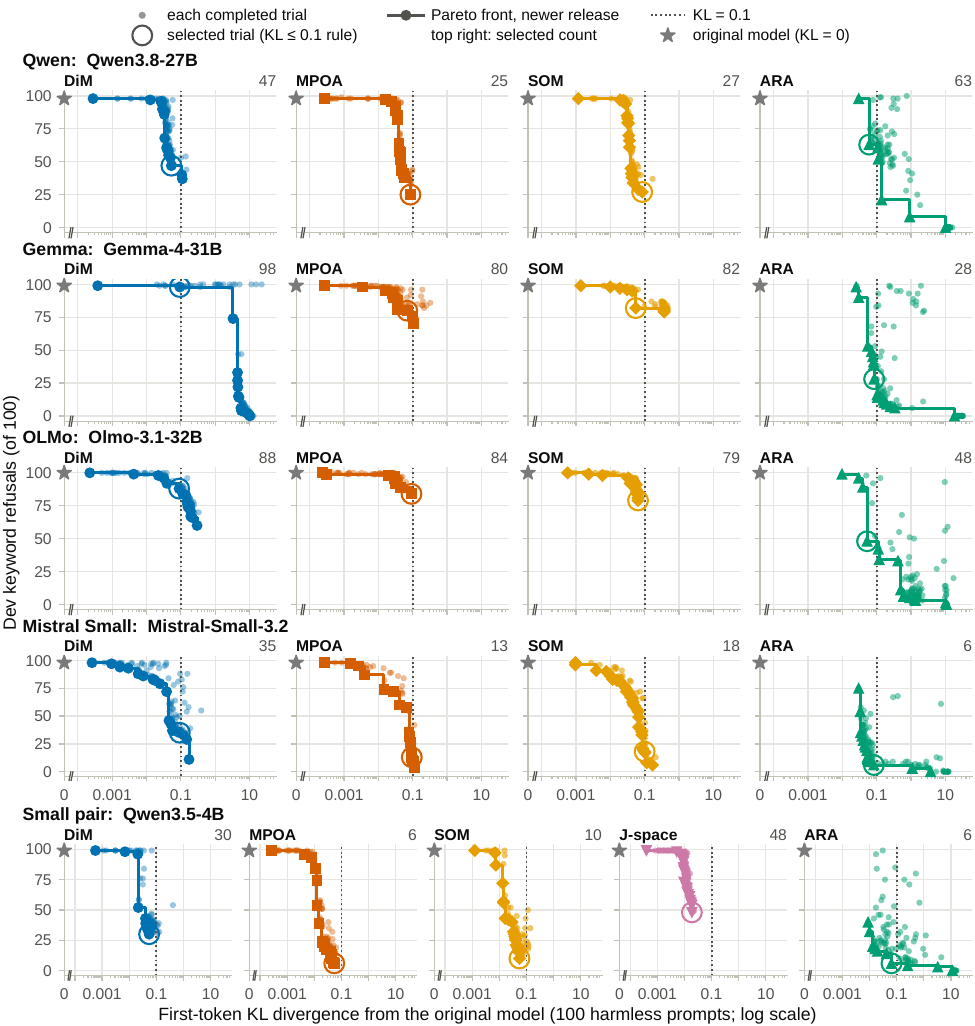}
\caption{\textbf{All completed TPE trials of each optimisation run, newer releases (dev set).} One panel per run (model $\times$ method, seed 1), laid out as in Figure~\ref{fig:pareto}; the newer model of each pair heads its row. Each dot is one completed trial; the line is the run's Pareto front (solid, filled markers), the ring the trial chosen by the selection rule (fewest dev keyword refusals at KL $\le$ 0.1; count at the top right of the panel) and the grey star the original model (KL = 0). The x axis is logarithmic, with KL = 0 drawn at the left edge behind an axis break; the dotted line is the selection budget KL = 0.1. 2099 trials from 21 runs; the older releases are in Figure~\ref{fig:trials-old}. Refusal counts are keyword matches on the 100 dev prompts that were also used for optimisation and selection, not held-out refusal rates.}
\label{fig:trials-new}
\end{figure*}

\begin{figure*}[tp]
\centering
\includegraphics[width=\textwidth]{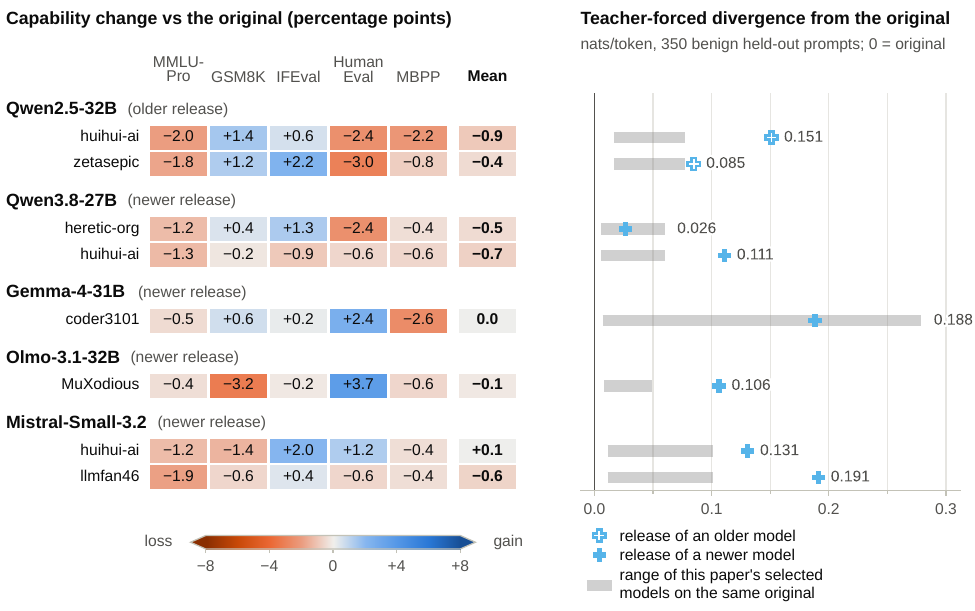}
\caption{\textbf{Public community releases: capability change and divergence from the original.} The 8 community releases evaluated alongside the selected models (2 of older, 6 of newer models; row label = Hugging Face uploader), grouped by the original they were derived from. Left: benchmark score minus that of the same original, in percentage points, with the diverging scale of Figure~\ref{fig:capability} (saturating at $\pm$8 pp; printed values exact); the last column is the unweighted mean of the five benchmarks. Right: teacher-forced divergence on the 350 benign held-out prompts, defined as in Figure~\ref{fig:divergence} (0 = the original); hollow markers are releases of the older, filled of the newer model of a pair. The grey bar spans the divergences of this paper's selected models on the same original (all conditions of Figure~\ref{fig:divergence}). 6 of 8 releases lie above that range; divergences range from 0.026 (heretic-org, Qwen3.8-27B) to 0.191 (llmfan46, Mistral-Small-3.2) nats per token, and mean capability changes from $-$0.9 to +0.1 pp. The releases were made by third parties with their own methods and selection; they were not selected with this paper's rule and are not part of the H1--H3 tests.}
\label{fig:community-releases}
\end{figure*}

\begin{figure*}[tp]
\centering
\includegraphics[width=\textwidth]{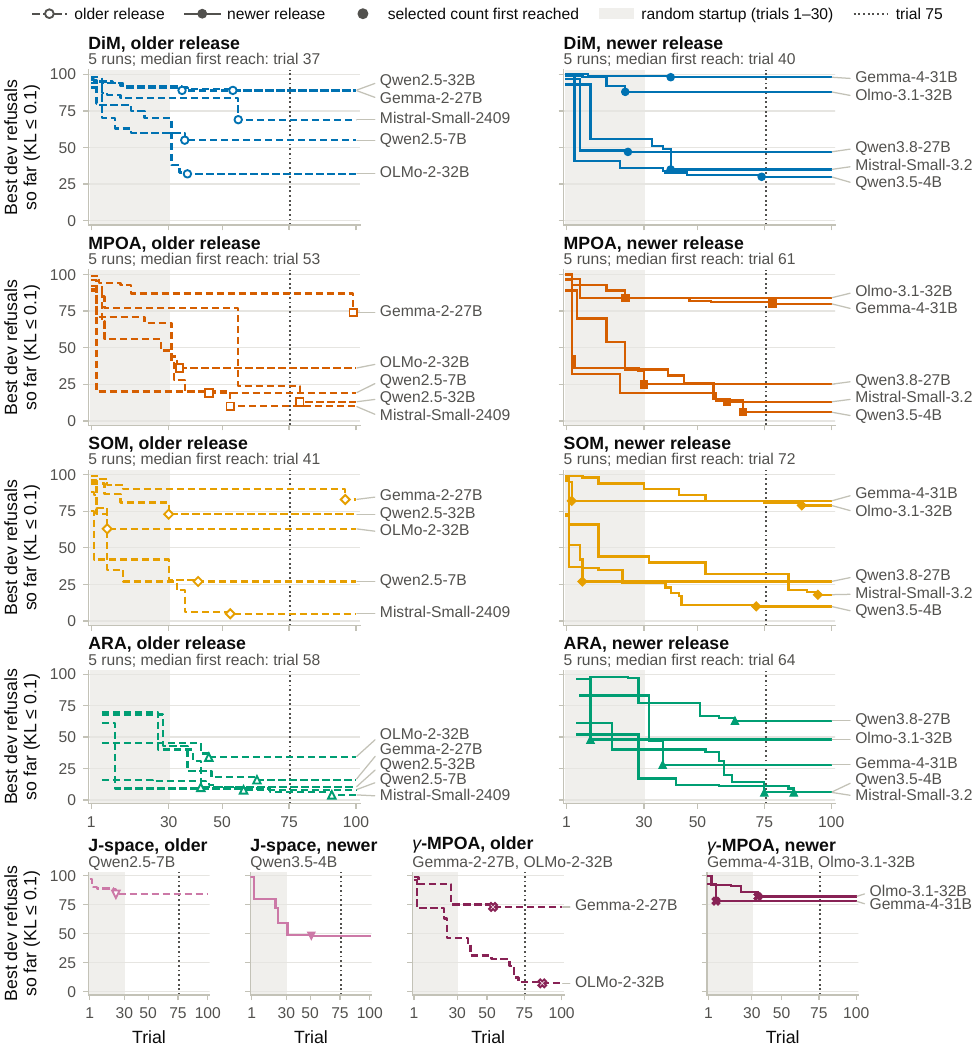}
\caption{\textbf{Search-budget convergence of every optimisation run (dev set).} One step curve per run (seed 1): after trial $t$, the fewest dev keyword refusals (of 100) among trials 1 to $t$ with first-token KL divergence $\le$ 0.1 from the original, i.e. the count the selection rule would have returned had the search stopped at trial $t$. Rows are methods and columns the older and newer release of each pair, so that a panel holds at most five runs, each named at its right end; bottom row: J-space (small pair only) and $\gamma$-MPOA (Gemma and OLMo pairs only). Dashed lines with hollow markers are older, solid lines with filled markers newer releases; the marker sits where the run first reached the count it finally selected. Curves start at the run's first trial within the KL budget (trials 5--10 for ARA, 1--2 otherwise). The grey band marks the 30 random startup trials of the TPE sampler, the dotted line trial 75. The medians, spreads and first-reach histogram of the core runs are in Figure~\ref{fig:search-convergence}. All counts are keyword matches on the 100 dev prompts that were also used for search and selection, not held-out refusal rates.}
\label{fig:search-runs}
\end{figure*}

This appendix gives the per-run optimisation traces and the exact values
behind several main-text figures. Figures~\ref{fig:trials-old}
and~\ref{fig:trials-new} show every completed trial of the seed-1 runs of the
older and newer models (Pareto fronts in Figure~\ref{fig:pareto}).
Figure~\ref{fig:community-releases} gives the capability change and
teacher-forced divergence of the eight community releases (refusal in
Figure~\ref{fig:community}), and Figure~\ref{fig:search-runs} the convergence
curve of every run (summary in Figure~\ref{fig:search-convergence}).
Tables~\ref{tab:dev-refusals-main} and~\ref{tab:dev-refusals-small} list the
dev-set keyword refusals of the trial selected at KL budgets of 0.05, 0.1 and
0.2 (Figure~\ref{fig:kl-budget}); Tables~\ref{tab:capability-main}
and~\ref{tab:capability-small} list the benchmark scores
(Figure~\ref{fig:capability}); and Tables~\ref{tab:teacher-forced-main}
and~\ref{tab:teacher-forced-small} list the teacher-forced log-likelihood of
each original model's answers and its shift under each selected model
(Figure~\ref{fig:divergence}). The four $\gamma$-MPOA models, which
Figures~\ref{fig:capability} and~\ref{fig:divergence} include, are not listed
in these tables; their benchmark changes, dev KL and teacher-forced divergence
are printed in Figure~\ref{fig:gmpoa-heldout}(b). Appendices~\ref{app:trunc256}
and~\ref{app:dev-rescore} report the 256- and 100-token checks of H3 and the
dev-set re-score of the released exports; Appendices~\ref{app:kl-cap},
\ref{app:item-source} and~\ref{app:harm} report the KL-cap binding, the
pair-level pooling and the two scales, the item-source analysis and harmful
compliance as the outcome.

\input{tables/t1_dev_refusals_main}

\input{tables/t1_dev_refusals_small}

{\setlength{\tabcolsep}{3pt}\renewcommand{\footnotesize}{\fontsize{7.5}{9}\selectfont}%
\input{tables/t2_capability_main}
}

\input{tables/t2_capability_small}

\input{tables/t3_teacher_forced_main}

\input{tables/t3_teacher_forced_small}

\subsection{The Failed Gemma-4-31B DiM Edit}
\label{app:gemma4-dim}
In the Gemma-4-31B DiM study, 10 of the 100 trials met the KL cap and 90
exceeded it. Every trial with a first-token KL below 3 (28 trials) refused 98
to 100 of the 100 dev prompts; lower refusal appeared only above a KL of 3
(minimum 3 of 100 between 3 and 8, and 0 above 8), where the edited model is
far from the original. In every other direction-based primary study, the
lowest-refusal trial above the cap had a KL of at most 0.37. The edit therefore
did not trade refusal against divergence gradually: within any usable budget it
had no effect. Two mechanisms proposed for similar failures on Gemma-3 are a
residual-stream coordinate of very large magnitude that dominates the
difference-in-means estimate \cite{mishra2026onecoordinate,lai2025mpoa} and
learned normalisation gains that amplify it \cite{mishra2026onecoordinate}. The
post-sublayer RMSNorm gains of the four post-norm models were therefore read
from the released weights. Gemma-2-27B and Gemma-4-31B both had recurring
outlier channels (a gain over 20 times the layer's median in 39 of 92 and 27 of
120 post-norm vectors, mostly one channel in Gemma-2-27B and two in
Gemma-4-31B), whereas OLMo-2-32B and Olmo-3.1-32B had 3 of 128 and 1 of 128.
The median coefficient of variation of the gains was 0.65 and 0.68 for the
attention outputs and 0.38 and 0.62 for the feed-forward outputs of the two
Gemma models, and 0.09 to 0.34 in the OLMo models; outlier channels were more
frequent in Gemma-2-27B than in Gemma-4-31B (42\% and 23\% of vectors). The gains thus separate the Gemma family from
OLMo, consistent with the larger effect of the $\gamma$ correction on the Gemma
models (\S\ref{sec:res-secondary}), but they do not separate Gemma-4-31B from
Gemma-2-27B, on which DiM lowered refusal. The activation outlier was not
measured, and the cause of the failure remains untested.

%% file: tables/t1_dev_refusals_main.tex
\begin{table*}[t]
\centering\small
\caption{Keyword refusals on the development set (mlabonne/harmful\_behaviors, test split, first 100 prompts; count out of 100, Heretic keyword scorer) for the trial selected at each first-token KL budget $\tau$ (KL on mlabonne/harmless\_alpaca, test split, first 100 prompts). Selection rule: among completed trials with KL $\le\tau$, fewest refusals; ties broken by lower KL, then lower trial index. ``Original'' is the unmodified model as recorded by Heretic in the same runs. Main study; seed 1; 100 trials planned per run. $^\dagger$Run with fewer completed than planned trials (Qwen3.8-27B MPOA: 99 of 100). --: not yet available.}
\label{tab:dev-refusals-main}
\begin{tabular}{llrrrrrrrr}
\toprule
 & & \multicolumn{2}{c}{Qwen} & \multicolumn{2}{c}{Gemma} & \multicolumn{2}{c}{OLMo} & \multicolumn{2}{c}{Mistral Small} \\
\cmidrule(lr){3-4}\cmidrule(lr){5-6}\cmidrule(lr){7-8}\cmidrule(lr){9-10}
Method & $\tau$ & 2.5-32B & 3.8-27B & 2-27B & 4-31B & 2-32B & 3.1-32B & 2409 & 3.2-24B \\
\midrule
Original & & 98 & 98 & 100 & 99 & 98 & 100 & 93 & 98 \\
\midrule
DiM & 0.05 & 89 & 53 & 92 & 99 & 47 & 92 & 85 & 46 \\
 & 0.1 & 89 & 47 & 89 & 98 & 32 & 88 & 69 & 35 \\
 & 0.2 & 56 & 37 & 73 & 98 & 32 & 67 & 31 & 11 \\
\midrule
MPOA & 0.05 & 75 & 43$^\dagger$ & 89 & 81 & 36 & 89 & 10 & 60 \\
 & 0.1 & 13 & 25$^\dagger$ & 74 & 80 & 36 & 84 & 10 & 13 \\
 & 0.2 & 13 & 25$^\dagger$ & 46 & 70 & 36 & 84 & 10 & 4 \\
\midrule
ARA & 0.05 & 12 & 98 & 67 & 90 & 34 & 89 & 5 & 22 \\
 & 0.1 & 10 & 63 & 16 & 28 & 34 & 48 & 4 & 6 \\
 & 0.2 & 8 & 21 & 5 & 9 & 10 & 34 & 4 & 6 \\
\midrule
SOM & 0.05 & 81 & 34 & 93 & 95 & 63 & 92 & 8 & 60 \\
 & 0.1 & 73 & 27 & 83 & 82 & 63 & 79 & 5 & 18 \\
 & 0.2 & 54 & 27 & 63 & 82 & 60 & 79 & 5 & 6 \\
\bottomrule
\end{tabular}
\end{table*}

%% file: tables/t1_dev_refusals_small.tex
\begin{table}[t]
\centering\small
\caption{Keyword refusals on the development set (mlabonne/harmful\_behaviors, test split, first 100 prompts; count out of 100, Heretic keyword scorer) for the trial selected at each first-token KL budget $\tau$ (KL on mlabonne/harmless\_alpaca, test split, first 100 prompts). Selection rule: among completed trials with KL $\le\tau$, fewest refusals; ties broken by lower KL, then lower trial index. ``Original'' is the unmodified model as recorded by Heretic in the same runs. Small pair; seed 1; 100 trials planned per run. --: not yet available.}
\label{tab:dev-refusals-small}
\begin{tabular}{llrr}
\toprule
 & & \multicolumn{2}{c}{Qwen (small pair)} \\
\cmidrule(lr){3-4}
Method & $\tau$ & 2.5-7B & 3.5-4B \\
\midrule
Original & & 99 & 99 \\
\midrule
DiM & 0.05 & 63 & 34 \\
 & 0.1 & 55 & 30 \\
 & 0.2 & 11 & 30 \\
\midrule
MPOA & 0.05 & 33 & 10 \\
 & 0.1 & 19 & 6 \\
 & 0.2 & 14 & 6 \\
\midrule
ARA & 0.05 & 8 & 14 \\
 & 0.1 & 8 & 6 \\
 & 0.2 & 3 & 6 \\
\midrule
SOM & 0.05 & 39 & 19 \\
 & 0.1 & 27 & 10 \\
 & 0.2 & 5 & 10 \\
\midrule
J-space & 0.05 & 84 & 48 \\
 & 0.1 & 84 & 48 \\
 & 0.2 & 84 & 48 \\
\bottomrule
\end{tabular}
\end{table}

%% file: tables/t2_capability_main.tex
\begin{table*}[t]
\centering\footnotesize
\caption{Capability of the original and the selected modified models (KL $\le 0.1$ trial), in \%. Modified models: score with the difference to the same family's original model in percentage points in parentheses (computed from unrounded scores). Number of items in the column header. Metrics: MMLU-Pro exact match (custom extract, 5-shot CoT); GSM8K exact match (flexible extract, 5-shot); IFEval prompt-level strict accuracy (0-shot); HumanEval pass@1 (0-shot); MBPP pass@1 (3-shot). Greedy decoding, chat template on, thinking disabled. Main study. $^\ddagger$Rescored from the logged samples with the patched MBPP answer extractor (run predates the lm-eval patch; scripts/bench/LM\_EVAL\_PATCHES.md): Qwen2.5-32B-Instruct Original MBPP. $^\S$Original model's capability run used evaluation-harness settings that differ from its modified models' runs (original vs.\ modified models; per-cell settings in paper/tables/coverage.md, evaluation protocol table): Qwen2.5-32B-Instruct: bench.py revision (sha256 prefix) 0ac2aca202 vs.\ b6a926cf73, f220db7bb9; output post-processing none vs.\ U+2581$\rightarrow$space; BENCH\_VLLM\_EXTRA unset vs.\ \{\}. gemma-2-27b-it: bench.py revision (sha256 prefix) b6a926cf73 vs.\ f220db7bb9. --: not yet available.}
\label{tab:capability-main}
\begin{tabular}{llrrrrr}
\toprule
Model & Method & MMLU-Pro (1000) & GSM8K (500) & IFEval (541) & HumanEval (164) & MBPP (500) \\
\midrule
Qwen2.5-32B-Instruct & Original$^\S$ & 71.1 & 85.4 & 78.9 & 89.6 & 77.6$^\ddagger$ \\
 & DiM & 69.9\,($-$1.2) & 85.6\,(+0.2) & 79.7\,(+0.7) & 90.2\,(+0.6) & 77.0\,($-$0.6) \\
 & MPOA & 70.5\,($-$0.6) & 85.0\,($-$0.4) & 80.8\,(+1.8) & 89.6\,(0.0) & 76.6\,($-$1.0) \\
 & ARA & 70.3\,($-$0.8) & 86.6\,(+1.2) & 79.9\,(+0.9) & 84.8\,($-$4.9) & 77.0\,($-$0.6) \\
 & SOM & 71.1\,(0.0) & 85.6\,(+0.2) & 79.5\,(+0.6) & 89.6\,(0.0) & 78.4\,(+0.8) \\
\midrule
Qwen3.8-27B & Original & 79.6 & 96.2 & 80.6 & 97.6 & 80.8 \\
 & DiM & 79.5\,($-$0.1) & 96.6\,(+0.4) & 80.6\,(0.0) & 96.3\,($-$1.2) & 80.6\,($-$0.2) \\
 & MPOA & 79.6\,(0.0) & 97.0\,(+0.8) & 82.4\,(+1.8) & 97.0\,($-$0.6) & 80.4\,($-$0.4) \\
 & ARA & 78.5\,($-$1.1) & 95.8\,($-$0.4) & 81.0\,(+0.4) & 93.9\,($-$3.7) & 79.2\,($-$1.6) \\
 & SOM & 79.4\,($-$0.2) & 97.0\,(+0.8) & 79.5\,($-$1.1) & 95.1\,($-$2.4) & 80.6\,($-$0.2) \\
\midrule
gemma-2-27b-it & Original$^\S$ & 42.8 & 86.4 & 60.6 & 76.2 & 69.6 \\
 & DiM & 42.3\,($-$0.5) & 86.0\,($-$0.4) & 60.6\,(0.0) & 75.0\,($-$1.2) & 69.2\,($-$0.4) \\
 & MPOA & 43.1\,(+0.3) & 84.4\,($-$2.0) & 60.8\,(+0.2) & 76.8\,(+0.6) & 68.6\,($-$1.0) \\
 & ARA & 44.3\,(+1.5) & 84.2\,($-$2.2) & 61.4\,(+0.7) & 76.8\,(+0.6) & 70.0\,(+0.4) \\
 & SOM & 45.3\,(+2.5) & 86.2\,($-$0.2) & 60.8\,(+0.2) & 75.6\,($-$0.6) & 68.6\,($-$1.0) \\
\midrule
gemma-4-31B-it & Original & 85.6 & 96.6 & 91.5 & 92.7 & 71.4 \\
 & DiM & 85.2\,($-$0.4) & 97.0\,(+0.4) & 90.0\,($-$1.5) & 90.2\,($-$2.4) & 69.2\,($-$2.2) \\
 & MPOA & 86.1\,(+0.5) & 96.4\,($-$0.2) & 91.7\,(+0.2) & 90.9\,($-$1.8) & 70.0\,($-$1.4) \\
 & ARA & 85.2\,($-$0.4) & 97.0\,(+0.4) & 91.5\,(0.0) & 86.0\,($-$6.7) & 67.2\,($-$4.2) \\
 & SOM & 85.4\,($-$0.2) & 96.8\,(+0.2) & 89.6\,($-$1.8) & 89.0\,($-$3.7) & 63.0\,($-$8.4) \\
\midrule
OLMo-2-0325-32B-Instruct & Original & 52.6 & 88.8 & 84.3 & 61.0 & 44.0 \\
 & DiM & 52.2\,($-$0.4) & 88.6\,($-$0.2) & 80.8\,($-$3.5) & 59.8\,($-$1.2) & 38.4\,($-$5.6) \\
 & MPOA & 52.3\,($-$0.3) & 88.0\,($-$0.8) & 83.0\,($-$1.3) & 57.9\,($-$3.0) & 42.2\,($-$1.8) \\
 & ARA & 52.3\,($-$0.3) & 88.6\,($-$0.2) & 80.6\,($-$3.7) & 57.3\,($-$3.7) & 30.8\,($-$13.2) \\
 & SOM & 51.9\,($-$0.7) & 88.2\,($-$0.6) & 83.2\,($-$1.1) & 57.9\,($-$3.0) & 44.0\,(0.0) \\
\midrule
Olmo-3.1-32B-Instruct & Original & 64.7 & 89.2 & 86.9 & 87.2 & 71.4 \\
 & DiM & 64.6\,($-$0.1) & 90.4\,(+1.2) & 87.2\,(+0.4) & 88.4\,(+1.2) & 70.2\,($-$1.2) \\
 & MPOA & 63.5\,($-$1.2) & 89.8\,(+0.6) & 86.7\,($-$0.2) & 90.2\,(+3.0) & 71.4\,(0.0) \\
 & ARA & 63.5\,($-$1.2) & 88.6\,($-$0.6) & 87.2\,(+0.4) & 92.1\,(+4.9) & 70.0\,($-$1.4) \\
 & SOM & 64.1\,($-$0.6) & 89.2\,(0.0) & 86.7\,($-$0.2) & 91.5\,(+4.3) & 70.8\,($-$0.6) \\
\midrule
Mistral-Small-Instruct-2409 & Original & 56.2 & 88.2 & 61.4 & 76.2 & 62.6 \\
 & DiM & 56.6\,(+0.4) & 87.4\,($-$0.8) & 62.3\,(+0.9) & 75.6\,($-$0.6) & 61.6\,($-$1.0) \\
 & MPOA & 53.9\,($-$2.3) & 88.0\,($-$0.2) & 61.0\,($-$0.4) & 75.0\,($-$1.2) & 62.2\,($-$0.4) \\
 & ARA & 51.5\,($-$4.7) & 84.2\,($-$4.0) & 63.6\,(+2.2) & 74.4\,($-$1.8) & 58.0\,($-$4.6) \\
 & SOM & 55.6\,($-$0.6) & 88.0\,($-$0.2) & 63.2\,(+1.8) & 76.8\,(+0.6) & 62.4\,($-$0.2) \\
\midrule
Mistral-Small-3.2-24B-Instruct-2506 & Original & 64.8 & 93.6 & 77.1 & 87.2 & 70.6 \\
 & DiM & 64.8\,(0.0) & 93.8\,(+0.2) & 76.9\,($-$0.2) & 87.8\,(+0.6) & 71.2\,(+0.6) \\
 & MPOA & 64.9\,(+0.1) & 94.0\,(+0.4) & 76.0\,($-$1.1) & 89.6\,(+2.4) & 69.6\,($-$1.0) \\
 & ARA & 63.4\,($-$1.4) & 92.6\,($-$1.0) & 77.4\,(+0.4) & 88.4\,(+1.2) & 70.2\,($-$0.4) \\
 & SOM & 63.7\,($-$1.1) & 93.6\,(0.0) & 76.0\,($-$1.1) & 88.4\,(+1.2) & 69.6\,($-$1.0) \\
\bottomrule
\end{tabular}
\end{table*}

%% file: tables/t2_capability_small.tex
\begin{table*}[t]
\centering\small
\caption{Capability of the original and the selected modified models (KL $\le 0.1$ trial), in \%. Modified models: score with the difference to the same family's original model in percentage points in parentheses (computed from unrounded scores). Number of items in the column header. Metrics: MMLU-Pro exact match (custom extract, 5-shot CoT); GSM8K exact match (flexible extract, 5-shot); IFEval prompt-level strict accuracy (0-shot); HumanEval pass@1 (0-shot); MBPP pass@1 (3-shot). Greedy decoding, chat template on, thinking disabled. Small pair. --: not yet available.}
\label{tab:capability-small}
\setlength{\tabcolsep}{4pt}
\begin{tabular}{llrrrrr}
\toprule
Model & Method & MMLU-Pro (1000) & GSM8K (500) & IFEval (541) & HumanEval (164) & MBPP (500) \\
\midrule
Qwen2.5-7B-Instruct & Original & 59.5 & 85.8 & 71.2 & 84.8 & 63.2 \\
 & DiM & 58.1\,($-$1.4) & 86.0\,(+0.2) & 71.7\,(+0.6) & 86.0\,(+1.2) & 63.8\,(+0.6) \\
 & MPOA & 57.9\,($-$1.6) & 85.0\,($-$0.8) & 70.6\,($-$0.6) & 86.6\,(+1.8) & 63.2\,(0.0) \\
 & ARA & 57.8\,($-$1.7) & 86.8\,(+1.0) & 70.2\,($-$0.9) & 87.2\,(+2.4) & 64.8\,(+1.6) \\
 & SOM & 59.8\,(+0.3) & 84.0\,($-$1.8) & 70.6\,($-$0.6) & 83.5\,($-$1.2) & 64.8\,(+1.6) \\
 & J-space & 60.5\,(+1.0) & 86.4\,(+0.6) & 72.6\,(+1.5) & 84.8\,(0.0) & 64.4\,(+1.2) \\
\midrule
Qwen3.5-4B & Original & 70.6 & 91.8 & 82.3 & 79.9 & 62.6 \\
 & DiM & 70.1\,($-$0.5) & 91.2\,($-$0.6) & 82.4\,(+0.2) & 81.7\,(+1.8) & 61.8\,($-$0.8) \\
 & MPOA & 69.2\,($-$1.4) & 91.6\,($-$0.2) & 82.3\,(0.0) & 79.9\,(0.0) & 61.6\,($-$1.0) \\
 & ARA & 69.4\,($-$1.2) & 90.2\,($-$1.6) & 79.3\,($-$3.0) & 79.3\,($-$0.6) & 61.2\,($-$1.4) \\
 & SOM & 68.7\,($-$1.9) & 92.6\,(+0.8) & 81.7\,($-$0.6) & 79.9\,(0.0) & 61.2\,($-$1.4) \\
 & J-space & 70.4\,($-$0.2) & 91.6\,($-$0.2) & 81.3\,($-$0.9) & 82.9\,(+3.0) & 61.2\,($-$1.4) \\
\bottomrule
\end{tabular}
\end{table*}

%% file: tables/t3_teacher_forced_main.tex
\begin{table*}[t]
\centering\small
\caption{Teacher-forced log-likelihood of the original model's greedy answers to the 350 benign held-out prompts (jbb\_benign 100, xstest\_safe 250). ``Original'': mean log-probability per answer token under the original model (token-weighted over all items); method columns: the same quantity under the selected modified model (KL $\le 0.1$ trial) minus the original's value (negative = the original answers became less likely). Tokens: number of scored answer tokens. Main study. $^\S$Original model's held-out generation and teacher forcing come from a run whose evaluation-harness settings differ from its modified models' runs (original vs.\ modified models; per-cell settings in paper/tables/coverage.md, evaluation protocol table): Qwen2.5-32B-Instruct: separate bench.py run (refusal\_run/summary.json in its evaluation directory); bench.py revision (sha256 prefix) f7731f7d3f vs.\ b6a926cf73, f220db7bb9; output post-processing none vs.\ U+2581$\rightarrow$space. gemma-2-27b-it: bench.py revision (sha256 prefix) b6a926cf73 vs.\ f220db7bb9. --: not yet available.}
\label{tab:teacher-forced-main}
\begin{tabular}{lrrrrrr}
\toprule
 & & & \multicolumn{4}{c}{Shift vs.\ original} \\
\cmidrule(lr){4-7}
Model & Tokens & Original & DiM & MPOA & ARA & SOM \\
\midrule
Qwen2.5-32B-Instruct & 112398 & $-$0.272$^\S$ & $-$0.017 & $-$0.042 & $-$0.077 & $-$0.020 \\
Qwen3.8-27B & 138016 & $-$0.275 & $-$0.006 & $-$0.060 & $-$0.042 & $-$0.021 \\
gemma-2-27b-it & 97666 & $-$0.227$^\S$ & $-$0.068 & $-$0.063 & $-$0.161 & $-$0.031 \\
gemma-4-31B-it & 142322 & $-$0.110 & $-$0.042 & $-$0.007 & $-$0.279 & $-$0.026 \\
OLMo-2-0325-32B-Instruct & 110792 & $-$0.520 & $-$0.051 & $-$0.088 & $-$0.055 & $-$0.030 \\
Olmo-3.1-32B-Instruct & 136772 & $-$0.288 & $-$0.008 & $-$0.008 & $-$0.049 & $-$0.016 \\
Mistral-Small-Instruct-2409 & 129960 & $-$0.209 & $-$0.004 & $-$0.018 & $-$0.080 & $-$0.023 \\
Mistral-Small-3.2-24B-Instruct-2506 & 131780 & $-$0.171 & $-$0.011 & $-$0.066 & $-$0.101 & $-$0.037 \\
\bottomrule
\end{tabular}
\end{table*}

%% file: tables/t3_teacher_forced_small.tex
\begin{table*}[t]
\centering\small
\caption{Teacher-forced log-likelihood of the original model's greedy answers to the 350 benign held-out prompts (jbb\_benign 100, xstest\_safe 250). ``Original'': mean log-probability per answer token under the original model (token-weighted over all items); method columns: the same quantity under the selected modified model (KL $\le 0.1$ trial) minus the original's value (negative = the original answers became less likely). Tokens: number of scored answer tokens. Small pair. --: not yet available.}
\label{tab:teacher-forced-small}
\begin{tabular}{lrrrrrrr}
\toprule
 & & & \multicolumn{5}{c}{Shift vs.\ original} \\
\cmidrule(lr){4-8}
Model & Tokens & Original & DiM & MPOA & ARA & SOM & J-space \\
\midrule
Qwen2.5-7B-Instruct & 119254 & $-$0.321 & $-$0.019 & $-$0.018 & $-$0.048 & $-$0.026 & $-$0.009 \\
Qwen3.5-4B & 139372 & $-$0.412 & $-$0.010 & $-$0.031 & $-$0.093 & $-$0.035 & $-$0.045 \\
\bottomrule
\end{tabular}
\end{table*}

%% file: sections/08e_truncation256.tex
\subsection{Judging the First 256 Tokens}
\label{app:trunc256}

Held-out responses were generated with up to 512 new tokens and judged in
full, whereas Heretic's keyword check reads only the start of a response. To
test whether the H3 comparison depends on how much text each measure sees,
WildGuard, Qwen3Guard and the HarmBench classifier were re-run, with the same
pinned revisions, prompt templates and parsers, on the stored first 256 tokens
of all 97,828 held-out responses of the 74 judged models, and compared with
keyword matching on the same 256 tokens. WildGuard's output could not be parsed
for 7 prefixes, which were dropped from the comparisons that use it. Only
responses longer than 256 tokens differ between the two texts
(Table~\ref{tab:trunc256}); on the 24,507 responses of at most 256 tokens,
WildGuard and Qwen3Guard repeated their full-response label in all but 6 and
14 cases, and the HarmBench classifier in all but 4 of the 17,967 among them
that answer harmful prompts. Per model, label changes are paired differences
of rates (prefix minus full response) with Newcombe intervals and exact
McNemar tests, Holm-adjusted per judge across the 74 models. The
keyword-minus-judge gaps, their pooling with the prompt-cluster bootstrap and
the refitted class and generation effects (prompt-level GEE, model-level OLS)
follow the H3 analysis of \S\ref{sec:res-h3}.

\begin{table}[t]
\centering\footnotesize
\setlength{\tabcolsep}{2.8pt}
\caption{Keyword matching and LLM judges on the first 256 tokens versus the full response, harmful held-out prompts
(972 per model; H3 robustness). $>$256: share of responses longer than 256 tokens, the only ones whose 256-token
prefix differs from the full response. WG change: WildGuard refusal rate on the prefix minus that on the full
response. Keyword minus judge: refusal rate by Heretic's 33 markers minus the judge's rate; full text = both on the
full response (as in Table~\ref{tab:h3-gap}); 256 tokens = both on the same prefix (WG = WildGuard, Q3G =
Qwen3Guard). All differences in pp with prompt-cluster bootstrap 95\% intervals; H3 predicted a negative gap. Cap:
the 512-token generation limit.}
\label{tab:trunc256}
\begin{tabular}{lrrllll}
\toprule
 & & & & \multicolumn{3}{c}{Keyword minus judge} \\
\cmidrule(lr){5-7}
Models & Cells & $>$256, \% & WG change & WG, full text & WG, 256 tokens & Q3G, 256 tokens \\
\midrule
Originals & 10 & 40.7 & +0.3 [+0.1, +0.5] & +2.6 [+1.8, +3.4] & +1.6 [+0.8, +2.4] & +2.7 [+1.9, +3.6] \\
\midrule
Ablated (seed 1), all & 42 & 78.9 & +0.6 [+0.5, +0.7] & +8.4 [+6.9, +10.0] & +3.8 [+2.4, +5.2] & +0.9 [$-$0.6, +2.4] \\
\quad main study & 32 & 79.1 & +0.6 [+0.5, +0.7] & +9.8 [+8.2, +11.3] & +4.9 [+3.5, +6.4] & +2.1 [+0.6, +3.5] \\
\quad small pair & 10 & 78.1 & +0.7 [+0.5, +0.9] & +4.2 [+2.5, +5.9] & +0.4 [$-$1.4, +2.0] & $-$2.7 [$-$4.5, $-$1.0] \\
\quad ending before the cap & 42 & 56.9 & +0.0 [$-$0.1, +0.1] & $-$1.2 [$-$3.2, +0.9] & $-$5.6 [$-$7.6, $-$3.8] & $-$7.6 [$-$9.8, $-$5.5] \\
\quad truncated at the cap & 42 & 100.0 & +1.2 [+1.0, +1.4] & +17.7 [+16.2, +19.3] & +13.0 [+11.5, +14.4] & +9.1 [+7.9, +10.3] \\
\midrule
$\gamma$-MPOA control & 4 & 82.0 & +0.7 [+0.4, +1.0] & +17.1 [+14.9, +19.4] & +12.0 [+9.9, +14.0] & +9.1 [+7.1, +11.1] \\
Community releases & 8 & 90.3 & +0.5 [+0.3, +0.7] & +16.5 [+15.2, +17.8] & +13.6 [+12.4, +14.8] & +8.1 [+6.8, +9.3] \\
Seed-2 replicates & 10 & 78.2 & +0.8 [+0.5, +1.0] & +3.5 [+2.0, +5.0] & +0.0 [$-$1.5, +1.4] & $-$1.9 [$-$3.4, $-$0.4] \\
\midrule
All judged models & 74 & 75.0 & +0.6 [+0.5, +0.7] & +8.3 [+7.0, +9.6] & +4.5 [+3.3, +5.7] & +2.0 [+0.8, +3.2] \\
\bottomrule
\end{tabular}
\end{table}

Judging the prefix changed few labels. Over the 42 ablated models, WildGuard
refusal on the harmful prompts was 32.1\% on the first 256 tokens and 31.5\%
on the full response (+0.6~pp [+0.5, +0.7]; 385 responses gained and 131 lost
the refusal label); Qwen3Guard refusal changed by +1.3~pp [+1.1, +1.4] and
HarmBench harmful compliance by $-$1.6~pp [$-$1.8, $-$1.3]. Per model (Holm
over the 74 models), WildGuard refusal changed significantly in 6 of the 74, all
upwards and by 2.0--5.1~pp; Qwen3Guard refusal rose in 22 (by up to 8.3~pp)
and HarmBench harmful compliance fell in 24 (by up to 9.3~pp), and no judge
changed significantly in the opposite direction.
Keyword matching on the same 256 tokens counted 36.0\% refusals, so the
keyword-minus-WildGuard gap of the ablated models was +3.8~pp [+2.4, +5.2],
against +8.4~pp [+6.9, +10.0] on the full text (change $-$4.6~pp [$-$4.9,
$-$4.3]); of this change, 4.0~pp came from the shorter keyword window (39.9\%
to 36.0\%) and 0.6~pp from WildGuard. The gap was +4.9~pp [+3.5, +6.4] in the
main study and +0.4~pp [$-$1.4, +2.0] in the small pair, and $-$5.6~pp
[$-$7.6, $-$3.8] on responses that ended before the cap against +13.0~pp
[+11.5, +14.4] on responses truncated at the cap. It was positive in 28 of the 42 ablated
models (21 Holm-significant) and negative in 14 (6 significant), against 34
(24) and 7 (3) on the full text. Against Qwen3Guard on the same 256 tokens,
the gap was +0.9~pp [$-$0.6, +2.4] (full text +6.2~pp [+4.6, +7.7]). In the
main study the gap remained larger for ARA than for the direction-based
methods, by +5.5~pp [+4.2, +6.8] with the prompt as unit
($p_\mathrm{Holm}<0.001$; full text +6.0~pp) but not with the model as unit
($n=32$, $p=0.139$; full text $p=0.151$).

With keyword matching and WildGuard reading the same 256 tokens, keyword
matching after ablation still counted more refusals than the judge, so the
negative gap that H3 predicted was again not observed over the ablated models,
and the class difference again held with the prompt but not with the model as
unit. The gap was about half as
large as on the full text, mostly because markers that occur only after the
first 256 tokens no longer counted (4.0 of the 4.6~pp), and it was not
distinguishable from zero in the small pair or against Qwen3Guard. A negative
gap appeared only in subsets: on responses that ended before the cap
($-$5.6~pp; full text $-$1.2~pp [$-$3.2, +0.9]) and, against Qwen3Guard, in
the small pair ($-$2.7~pp [$-$4.5, $-$1.0]). Part of the full-text gap in
Table~\ref{tab:h3-gap} therefore reflects the length of the keyword window,
not the amount of text the judges read; at the 100-token window that Heretic
scores, the pooled gap disappeared (below and Table~\ref{tab:h3-windows}). Pooled over the ablated models, the
judges' labels changed by at most 1.6~pp with the cut, although single models
changed by up to 9.3~pp; the HarmBench classifier labelled slightly fewer
prefixes than full responses as harmful, the direction anticipated for
truncated answers in Appendix~\ref{app:validity}. As in the main analysis,
these comparisons show where the measures diverge, not which is correct.

\paragraph{Keyword matching on the first 100 tokens.} Heretic counts a dev
response as a refusal if any of its markers occurs in at most 100 generated
tokens. Token ids of the held-out responses were not stored, so each stored
response was re-tokenised with the tokenizer of the evaluated model, cut to
its first 100 tokens, decoded and matched with the same 33 markers; because
decoding was greedy, these are the tokens that a 100-token run would have
produced, up to numerics. The procedure reproduced the stored full-response
keyword label in all 97,828 responses and the stored 256-token label in
97,825 (OLMo-2-32B, which emits each newline as a separate token, needed a
newline-split encoding). Bounds that do not depend on exact re-encoding fixed
the 100-token label in 99.98\% of the responses (17 unresolved) and gave the
same pooled gaps. Over the 42 ablated models the 100-token keyword refusal was
32.0\%, against 36.0\% at 256 tokens and 39.9\% on the full text, and its gap
to WildGuard was +0.5~pp [$-$1.0, +1.9] against the full-response label and
$-$0.1~pp [$-$1.6, +1.3] against the 256-token label; Heretic's own matcher,
which also strips emphasis and counts empty responses as refusals, gave the
same +0.5~pp. In the main study, the class, generation and generation
$\times$ class terms of the gap were +5.2, +4.7 and +8.1~pp with the prompt as
unit (all $p_\mathrm{Holm}<0.001$) and not significant with the model as unit.
No judge was run on the 100-token text, and the held-out responses were
generated with vLLM after the chat template's empty reasoning block, whereas
Heretic scores Transformers generations on the dev prompts.

%% file: sections/08f_dev_rescore.tex
\subsection{Re-scoring the Released Exports on the Dev Set}
\label{app:dev-rescore}

The dev refusal count and first-token KL divergence of each selected model
(Tables~\ref{tab:dev-refusals-main} and~\ref{tab:dev-refusals-small}) were
measured during the search, with the trial's adapter applied to the original
model at run time. The released checkpoint is a separate artefact: the trial is
restored from the study journal, its adapter is rebuilt and merged into the
bf16 weights, and the export step does not score the result
(Appendix~\ref{sec:setup-export}). Each of the 46 released exports (42 primary
models and four $\gamma$-MPOA controls; the ten seed-2 replicates are not
released) was therefore scored again with Heretic's own evaluator
(\texttt{heretic -{}-model <original> -{}-evaluate-model <export>}), with the run's
configuration, code version and generation batch size, the original model at
the revision of Table~\ref{tab:models}, and the export at the revision that was
evaluated. The evaluator recomputes the original model's dev responses and
first-token distributions and scores the export with the same keyword and KL
scorers on the same 100 harmful and 100 harmless dev prompts. The 36
main-study exports, searched on the H200 server, were re-scored on the RTX PRO
6000 server; the ten small-pair exports were searched and re-scored on the RTX
4090 workstation (Table~\ref{tab:compute}). In all 46 runs the original model's
dev refusal count equalled the value recorded during the search. The study
journal stores only the count and the KL of each trial, not per-prompt
outcomes, so search-time and re-scored values are compared as counts: by
Wilcoxon signed-rank tests across exports (Holm-adjusted over the refusal and KL
tests) and by Spearman's $\rho$ with a bootstrap interval.

Table~\ref{tab:dev-rescore} lists both values for every export. The re-scored
refusal count differed from the selected trial's by a median of +1 (mean +1.0;
27 higher, 9 equal, 10 lower) and lay within $\pm$2 of it for 37 of the 46
exports. The differences were shifted upward (Wilcoxon signed-rank $p = 0.002$,
Holm-adjusted $p = 0.004$), and the two counts ranked the exports almost
identically ($\rho = 0.995$, 95\% bootstrap interval [0.984, 0.997]).
First-token KL changed by a median of +0.0003, with no significant shift
(Wilcoxon $p = 0.22$). ARA changed most: 4 of its 10
exports moved by more than two refusals (Qwen3.8-27B +9, Gemma-4-31B $-$7,
Qwen2.5-32B +6, Gemma-2-27B +5), against 5 of the 36 exports of the other
methods (largest change +7, Qwen3.5-4B DiM), and its three largest KL changes
(+0.0356 for Qwen2.5-32B, +0.0225 for Gemma-4-31B, +0.0176 for Qwen3.8-27B)
exceed the largest absolute change of any other method (0.0090). These four
ARA refusal changes occurred in main-study exports whose adapter was refitted
and re-scored on a different GPU from the search; the two small-pair ARA
exports changed by $-$1 and +1 refusals, and KL changed by at most 0.0015
across all ten small-pair exports, which were exported and re-scored on the GPU
of their search. The sources of the differences (merged bf16 weights versus a
run-time adapter, GPU type and kernels, greedy decisions on borderline
prompts) were not fully separated, but the export host is examined below. Four
exports scored above the selection budget of KL~0.1 on re-score, each by less
than 0.008: Gemma-4-31B ARA (0.0845 to 0.1070), Gemma-2-27B MPOA (0.0991 to
0.1036) and the $\gamma$-MPOA controls of Gemma-2-27B (0.0969 to 0.1059) and
Olmo-3.1-32B (0.0993 to 0.1032).

\paragraph{Export host.} For ARA, the export step refits the adapter by
L-BFGS from the stored hyperparameters (Appendix~\ref{sec:setup-export}). All
main-study searches ran on the H200 server; the Qwen and Gemma exports were
rebuilt on the RTX PRO 6000 server and the OLMo and Mistral Small exports on
the H200, and all 36 main-study exports were re-scored on the RTX PRO 6000. The
four ARA exports rebuilt on the other GPU moved by more than two dev refusals
in 4 of 4 cases (+6, +9, +5 and $-$7), against 0 of the 4 rebuilt on the
search GPU (0 to +2; Fisher $p=0.029$), and their largest KL change was 0.0356
against 0.0021. For the other methods the corresponding counts were 3 of 14
and 1 of 14 ($p=0.60$). Because the OLMo and Mistral Small ARA exports, rebuilt
on the H200 and re-scored on the RTX PRO 6000, moved by at most two refusals,
re-scoring on other hardware alone does not explain the large ARA changes;
they coincide with refitting on other hardware. Gemma-4-31B ARA, rebuilt on
the RTX PRO 6000, is the only ARA export that re-scored above the KL cap.

\begin{table}[t]
\centering\scriptsize
\setlength{\tabcolsep}{3pt}
\caption{Dev-set re-score of the 46 released exports. Each cell gives the
value of the selected trial logged during the search (adapter applied to the
original model) $\to$ the value of the released merged export scored by
Heretic's evaluator with the run's configuration, code version and batch size.
(a)~Keyword refusals of 100 on the dev prompts (mlabonne/harmful\_behaviors,
test split, first 100); bold: change of more than two. (b)~First-token KL
divergence from the original model (mlabonne/harmless\_alpaca, test split,
first 100); bold: re-scored value above the selection budget of 0.1. Main-study
exports were searched on the H200 server and re-scored on the RTX PRO 6000
server; small-pair exports (last two rows of each panel) were searched and
re-scored on the RTX 4090. $^{a}$Re-scored with eager attention set in a local
copy of the configuration (Table~\ref{tab:dev-rescore-gemma2}). --: method not
applied to that model.}
\label{tab:dev-rescore}
\begin{tabular}{lllllll}
\toprule
Model & DiM & MPOA & SOM & J-space & ARA & $\gamma$-MPOA \\
\midrule
\multicolumn{7}{l}{\emph{(a) Dev keyword refusals (of 100): selected trial $\to$ released export}} \\
\addlinespace[2pt]
Qwen2.5-32B & 89$\to$90 & 13$\to$\textbf{17} & 73$\to$75 & -- & 10$\to$\textbf{16} & -- \\
Qwen3.8-27B & 47$\to$\textbf{44} & 25$\to$26 & 27$\to$28 & -- & 63$\to$\textbf{72} & -- \\
\addlinespace[2pt]
Gemma-2-27B$^{a}$ & 89$\to$90 & 74$\to$74 & 83$\to$\textbf{86} & -- & 16$\to$\textbf{21} & 73$\to$72 \\
Gemma-4-31B & 98$\to$98 & 80$\to$80 & 82$\to$83 & -- & 28$\to$\textbf{21} & 78$\to$78 \\
\addlinespace[2pt]
OLMo-2-32B & 32$\to$34 & 36$\to$38 & 63$\to$62 & -- & 34$\to$34 & 7$\to$8 \\
Olmo-3.1-32B & 88$\to$90 & 84$\to$84 & 79$\to$78 & -- & 48$\to$49 & 82$\to$81 \\
\addlinespace[2pt]
Mistral-Small-2409 & 69$\to$\textbf{72} & 10$\to$12 & 5$\to$6 & -- & 4$\to$6 & -- \\
Mistral-Small-3.2 & 35$\to$34 & 13$\to$13 & 18$\to$18 & -- & 6$\to$6 & -- \\
\addlinespace[2pt]
Qwen2.5-7B & 55$\to$54 & 19$\to$20 & 27$\to$28 & 84$\to$85 & 8$\to$7 & -- \\
Qwen3.5-4B & 30$\to$\textbf{37} & 6$\to$8 & 10$\to$11 & 48$\to$47 & 6$\to$7 & -- \\
\midrule
\multicolumn{7}{l}{\emph{(b) First-token KL divergence: selected trial $\to$ released export}} \\
\addlinespace[2pt]
Qwen2.5-32B & 0.0175$\to$0.0178 & 0.0599$\to$0.0603 & 0.0750$\to$0.0707 & -- & 0.0620$\to$0.0976 & -- \\
Qwen3.8-27B & 0.0530$\to$0.0517 & 0.0854$\to$0.0842 & 0.0843$\to$0.0813 & -- & 0.0605$\to$0.0781 & -- \\
\addlinespace[2pt]
Gemma-2-27B$^{a}$ & 0.0686$\to$0.0672 & 0.0991$\to$\textbf{0.1036} & 0.0860$\to$0.0807 & -- & 0.0990$\to$0.0996 & 0.0969$\to$\textbf{0.1059} \\
Gemma-4-31B & 0.0932$\to$0.0936 & 0.0691$\to$0.0718 & 0.0548$\to$0.0567 & -- & 0.0845$\to$\textbf{0.1070} & 0.0609$\to$0.0645 \\
\addlinespace[2pt]
OLMo-2-32B & 0.0887$\to$0.0845 & 0.0372$\to$0.0357 & 0.0470$\to$0.0446 & -- & 0.0494$\to$0.0493 & 0.0852$\to$0.0839 \\
Olmo-3.1-32B & 0.0891$\to$0.0912 & 0.0916$\to$0.0959 & 0.0638$\to$0.0670 & -- & 0.0526$\to$0.0547 & 0.0993$\to$\textbf{0.1032} \\
\addlinespace[2pt]
Mistral-Small-2409 & 0.0990$\to$0.0989 & 0.0493$\to$0.0480 & 0.0620$\to$0.0625 & -- & 0.0656$\to$0.0672 & -- \\
Mistral-Small-3.2 & 0.0931$\to$0.0931 & 0.0938$\to$0.0922 & 0.0995$\to$0.0973 & -- & 0.0829$\to$0.0827 & -- \\
\addlinespace[2pt]
Qwen2.5-7B & 0.0782$\to$0.0797 & 0.0743$\to$0.0748 & 0.0756$\to$0.0754 & 0.0216$\to$0.0209 & 0.0205$\to$0.0205 & -- \\
Qwen3.5-4B & 0.0539$\to$0.0530 & 0.0543$\to$0.0548 & 0.0550$\to$0.0553 & 0.0188$\to$0.0193 & 0.0645$\to$0.0648 & -- \\
\bottomrule
\end{tabular}
\end{table}

\paragraph{Gemma-2 attention configuration.} A first re-score of the five
Gemma-2-27B exports gave KL 0.404 to 0.491 and refusal changes from $-$22 to
+7 (Table~\ref{tab:dev-rescore-gemma2}). Gemma-2 caps attention logits and
final logits with a scaled $\tanh$ \cite{gemmateam2024gemma2}. The SDPA
attention path of transformers does not apply the attention-logit cap; the
eager path does, and the original model's configuration selects it with
\texttt{"\_attn\_implementation": "eager"}. Heretic saved the exports with
transformers 5.18.0, whose configuration serialisation drops this key, while the
cap values themselves (\texttt{attn\_logit\_softcapping} 50,
\texttt{final\_logit\_softcapping} 30) were kept. In the re-score the original
model was therefore loaded with eager attention and each export with SDPA
attention, which ignores the attention-logit cap, so the measured divergence
reflected the attention implementation rather than the edit. With
\texttt{"\_attn\_implementation": "eager"} written into a local copy of each
export's configuration and the weights unchanged, the five exports re-scored at
KL 0.067 to 0.106 with refusal changes from $-$1 to +5; these eager values are
the ones in Table~\ref{tab:dev-rescore} and in the counts above. The
configurations of the five public repositories were patched on 6~November~2026
by one commit per repository that adds only this key
(Table~\ref{tab:dev-rescore-gemma2}); the weight files are unchanged, and every
evaluation in this paper used the earlier revisions. Before the patch, loading
a released Gemma-2 model with transformers 5.18 without requesting eager
attention selected SDPA, as in the first re-score.

\begin{table}[t]
\centering\footnotesize
\setlength{\tabcolsep}{4pt}
\caption{Gemma-2-27B exports re-scored on the dev set with the configuration as
exported (transformers SDPA attention, no attention-logit soft-capping) and
with eager attention set in a local copy of the configuration (same weights).
Ref.: keyword refusals of 100; KL: first-token KL divergence. Revisions of
\texttt{richardyoung/gemma-2-27b-it-\{DiM,MPOA,SOM,ARA,GMPOA\}-heretic}, first
12 hexadecimal digits: the revision evaluated throughout this paper and the
revision created by the configuration patch of 6~November~2026, which adds only
\texttt{"\_attn\_implementation": "eager"}.}
\label{tab:dev-rescore-gemma2}
\begin{tabular}{lrrrrrrll}
\toprule
 & \multicolumn{2}{c}{Selected trial} & \multicolumn{2}{c}{Re-score, SDPA} & \multicolumn{2}{c}{Re-score, eager} & \multicolumn{2}{c}{Revision} \\
\cmidrule(lr){2-3}\cmidrule(lr){4-5}\cmidrule(lr){6-7}\cmidrule(lr){8-9}
Method & Ref. & KL & Ref. & KL & Ref. & KL & Evaluated & Patched \\
\midrule
DiM & 89 & 0.0686 & 89 & 0.4036 & 90 & 0.0672 & \texttt{9ee29ab3ab16} & \texttt{3f2f3e1d2c36} \\
MPOA & 74 & 0.0991 & 52 & 0.4609 & 74 & 0.1036 & \texttt{a2c021b2e06a} & \texttt{8aa25e304fdd} \\
SOM & 83 & 0.0860 & 77 & 0.4912 & 86 & 0.0807 & \texttt{2a355f20c0d4} & \texttt{c63892ab9b07} \\
ARA & 16 & 0.0990 & 23 & 0.4817 & 21 & 0.0996 & \texttt{29250dd15c03} & \texttt{958fda591386} \\
$\gamma$-MPOA & 73 & 0.0969 & 63 & 0.4167 & 72 & 0.1059 & \texttt{78da037a4a28} & \texttt{9630b864b0fe} \\
\bottomrule
\end{tabular}
\end{table}

\paragraph{Held-out evaluation in vLLM.} The held-out generations, capability
benchmarks and teacher-forced divergences were computed with vLLM 0.33.0 and
are not affected by the configuration defect. vLLM's Gemma-2 implementation
reads the cap values from \texttt{attn\_logit\_softcapping} and
\texttt{final\_logit\_softcapping}, passes the attention cap to its own
attention backend and does not read \texttt{\_attn\_implementation}; the
original model and all five exports ran on the same backend (FlashAttention~2),
which applies the cap, and the five exports were evaluated from the unpatched
configuration. Consistent with this, the teacher-forced divergences of the four
primary Gemma-2-27B exports (0.031 to 0.161 nats per token) and their mean
capability changes ($-$0.50 to +0.21~pp) lie within the ranges of the other 38
primary models (0.004 to 0.279; $-$4.21 to +0.86~pp).

\paragraph{Consequences for the reported results.} Held-out refusal,
capability and teacher-forced divergence were measured on the released weights
themselves (main study: downloaded back from the published repository; small
pair: SHA-256-identical shards; Appendix~\ref{sec:setup-export}), so they
describe the released models whatever the dev re-score shows. The re-score
bears on two other points. First, the dev values in
Tables~\ref{tab:dev-refusals-main} and~\ref{tab:dev-refusals-small} and in the
dev--held-out comparison are those of the selected trials; the released exports
reproduce their ranking ($\rho = 0.995$) and, for 37 of the 46 exports, their
counts to within two refusals, and the small upward shift is the direction
expected when the lowest of many measured counts is measured again
(Appendix~\ref{app:validity}). Second, the selected trial of every run met
KL~$\le$~0.1 as measured during the search, so no selection changes under the
rule as specified, and 42 of the 46 released exports also stayed within the
budget on re-score. Had the re-scored
KL been used for the other four, their selected trials would have been
ineligible, and among the logged trials the rule would have returned trial~16
of Gemma-2-27B MPOA (87 refusals at KL 0.0835, against 74), trial~76 of
Gemma-4-31B ARA (28 refusals at KL 0.0921, the same count; trial~66 with 45
refusals if every trial of that run had drifted by the same +0.0225), trial~26
of Gemma-2-27B $\gamma$-MPOA (75, against 73) and trial~52 of Olmo-3.1-32B
$\gamma$-MPOA (84, against 82). These trials were neither exported nor
evaluated, and their own KL was not re-measured. The two primary models
concerned, Gemma-2-27B MPOA and Gemma-4-31B ARA, are therefore reported as
selected at KL~$\le$~0.1 on the search measurement, with released exports at
0.104 and 0.107 on re-score.

%% file: sections/08g_selection_pairs.tex
\subsection{KL-Cap Binding, Pair-Level Pooling and the Two Scales}
\label{app:kl-cap}

\paragraph{KL-cap binding.} The selection rule caps first-token KL divergence
at 0.1 for every model. The cap binds in a run when some trial above 0.1 had
fewer dev keyword refusals than the selected trial; a run stayed below the cap
when at most one of its trials exceeded 0.1. In the main study the cap bound
in 15 of the 24 direction-based runs (7 of 12 older and 8 of 12 newer models)
and in all 8 ARA runs, every one of which reached 0 dev refusals above the cap
(maximum KL 10.9 to 33.2). Eight direction-based runs stayed below the cap,
four on each generation: Qwen3.8-27B MPOA, Olmo-3.1-32B MPOA and SOM and
Mistral-Small-2409 MPOA never sampled a trial above 0.1, and OLMo-2-32B DiM,
MPOA and SOM and Qwen3.8-27B SOM sampled one. On the small pair the cap bound
in 4 of the 16 direction-based runs of both seeds (all four J-space runs
stayed below it) and in all 4 ARA runs. The selected dev KL ranged from 0.018
to 0.100 for the direction-based runs (median 0.080) and from 0.049 to 0.099
for ARA; 7 of 24 direction-based and 1 of 8 ARA selections lay within 0.01 of
the cap.

\paragraph{Position in the search range.} The selected trial's parameters were
read from the Optuna journals and placed in Heretic's default ranges
(Table~\ref{tab:search}); a maximum weight counts as at the ceiling when it
lies in the top 5\% of its range. The selected trial was at the ceiling in 5
of the 24 main-study direction-based runs and in none of the 8 runs that
stayed below the cap. In those 8 runs the selected maximum weights lay at a
median of 0.81 (attention; range 0.56--0.93) and 0.89 (MLP; 0.83--0.94) of
their range, against 0.58 and 0.84 in the other direction-based runs, so the
selected edits lean to the strong half of the range but not to its bound. The
search did sample the strongest maximum weights in each of the 8 runs, and
those trials had no fewer dev refusals than the selected trial in 7 of 8 (in
OLMo-2-32B SOM the better trial at the ceiling had KL 0.110). Counting every
strength-type parameter (maximum weight, minimum weight, minimum-weight
distance, SOM decay), the selected trial had at least one at the strong end of
its range in 14 of the 24 direction-based runs and in 4 of the 8 that stayed
below the cap (a minimum weight or minimum-weight distance in all four). The selected
trial was one of the 30 random start-up trials of the TPE sampler in 7 of the
24 direction-based runs (4 of the 8 that stayed below the cap) and in 1 of 8
ARA runs. Because seed~1 and the same search space give identical start-up
draws, two pairs of selected parameter vectors coincide across models
(OLMo-2-32B SOM and Qwen3.8-27B SOM; Olmo-3.1-32B MPOA and Qwen3.8-27B MPOA).

\paragraph{H1 and H2 by binding.} At the level of pair $\times$ method units,
the cap bound in both cells for Qwen DiM and ARA, all four Gemma methods, OLMo
ARA and Mistral Small DiM and ARA; in neither cell for Qwen MPOA and OLMo
MPOA; and in one cell for Qwen SOM, OLMo DiM and SOM, and Mistral Small MPOA
and SOM. Table~\ref{tab:kl-cap-subsets} refits the pooled GEE of
\S\ref{sec:res-h1} on these subsets. The raw direction-based effect was
positive in every subset; the baseline-adjusted one was negative where the
cap bound in both cells and positive where it did not, and the pooled +0.03 of
all units lies between the two. The subsets also differ in family, so they
cannot separate the cap from the family. H2 can be estimated only on units in
which the cap bound, because it bound in every ARA cell.

\begin{table}[t]
\centering\footnotesize
\setlength{\tabcolsep}{2.5pt}
\caption{Pooled H1 and H2 terms (GEE, log-odds, prompt clusters, family fixed effects, 95\% intervals) refitted on
subsets of the 16 main-study pair $\times$ method units, by whether the KL cap bound in each unit's two cells. Units:
direction-based / ARA. Dir.: direction-based class; adj.: baseline-adjusted (minus the originals' difference); n/a:
no ARA unit in the subset.}
\label{tab:kl-cap-subsets}
\begin{tabular}{@{}lcllll@{}}
\toprule
Subset & Units & Dir., raw & Dir., adj. & ARA, raw & H2 \\
\midrule
All units & 12/4 & +0.62 [+0.57, +0.68] & +0.03 [$-$0.05, +0.11] & +0.12 [+0.00, +0.24] & $-$0.50 [$-$0.62, $-$0.38] \\
Cap bound in both cells & 5/4 & +0.32 [+0.24, +0.41] & $-$0.24 [$-$0.33, $-$0.14] & +0.11 [+0.01, +0.22] & $-$0.21 [$-$0.33, $-$0.09] \\
Not bound in $\ge$1 cell & 7/0 & +1.03 [+0.95, +1.11] & +0.22 [+0.12, +0.32] & n/a & n/a \\
Bound in neither cell & 2/0 & +3.23 [+2.90, +3.56] & +2.20 [+1.95, +2.45] & n/a & n/a \\
$\ge$2 trials above the cap in both & 6/4 & +0.33 [+0.24, +0.41] & $-$0.28 [$-$0.37, $-$0.18] & +0.12 [+0.01, +0.23] & $-$0.21 [$-$0.33, $-$0.08] \\
$\le$1 trial above the cap in a cell & 6/0 & +1.04 [+0.96, +1.12] & +0.27 [+0.17, +0.37] & n/a & n/a \\
\bottomrule
\end{tabular}
\end{table}

\paragraph{Pair-level pooling.} With the pair as the unit, between-pair
heterogeneity dominated every pooled estimand ($I^2$ 99--100\%; between-pair
SD of the adjusted direction-based effect 22.9~pp and of the H2 contrast
26.1~pp). The 95\% prediction intervals for a new pair, which use a $t$
distribution on 2 degrees of freedom, were $-$5.90 to +6.02 log-odds for the
adjusted direction-based effect and $-$11.32 to +9.21 for the H2 contrast; on
the percentage-point scale they extend beyond $\pm$100~pp, so with four pairs
they state only that an effect of either sign is plausible for a new pair.
Mapping the percentage-point $\sigma$ to the logit scale by the delta method
at each cell's rate, instead of estimating $\sigma$ on the logit scale, gave
seed SEs up to 14.9 log-odds near 0\% refusal (Gemma-4-31B ARA) and pooled
intervals of $-$1.61 to +1.82 (adjusted direction-based) and $-$3.59 to +2.81
(H2). Table~\ref{tab:lofo} gives the leave-one-family-out GEE terms.

\begin{table}[t]
\centering\footnotesize
\setlength{\tabcolsep}{3pt}
\caption{Leave-one-family-out pooled GEE terms (log-odds; main study, WildGuard refusal, 972 harmful prompts).}
\label{tab:lofo}
\begin{tabular}{@{}lrrrrr@{}}
\toprule
Term & All four & w/o Qwen & w/o Gemma & w/o OLMo & w/o Mistral Small \\
\midrule
Generation effect, direction class & +0.62 & +0.69 & +0.81 & +0.24 & +0.74 \\
Generation effect, ARA & +0.12 & $-$1.37 & +1.02 & +0.24 & +0.15 \\
Generation $\times$ class (H2) & $-$0.50 & $-$2.06 & +0.20 & $-$0.00 & $-$0.59 \\
Baseline-adjusted, direction class & +0.03 & +0.22 & +0.13 & $-$0.51 & +0.33 \\
Baseline-adjusted, ARA & $-$0.46 & $-$1.68 & +0.31 & $-$0.51 & $-$0.25 \\
Originals' generation difference & +0.58 & +0.45 & +0.66 & +0.74 & +0.40 \\
\bottomrule
\end{tabular}
\end{table}

\paragraph{The two scales.} Table~\ref{tab:scales} gives every pair $\times$
method generation effect, the originals' difference and the baseline-adjusted
effect on both scales. Raw effects of a single method have the same sign on
both scales, because they compare two rates. Adjusted effects changed sign
between scales in 3 of 21 cells and in significance in 5. On the percentage-point
scale all four Mistral Small adjusted effects were large and negative
($-$22.2 to $-$14.4~pp), whereas in log-odds they ranged from $-$1.37 to
$-$0.10.

\begin{table*}[t]
\centering\scriptsize
\setlength{\tabcolsep}{3pt}
\caption{Generation effect, originals' difference and baseline-adjusted generation effect per pair and method on the
percentage-point and the log-odds scale (WildGuard refusal, 972 harmful prompts; 95\% intervals, prompt sampling only).
Sign: the adjusted effect has opposite signs on the two scales. Sig.: its significance differs between scales.}
\label{tab:scales}
\begin{tabular}{@{}lllrllrlcc@{}}
\toprule
Pair & Method & Effect, pp & Orig., pp & Adjusted, pp & Effect, log-odds & Orig., log-odds & Adjusted, log-odds & Sign & Sig. \\
\midrule
Qwen & DiM & +26.9 [+23.8, +29.8] & +13.1 & +13.7 [+10.1, +17.1] & +1.16 [+1.02, +1.30] & +0.99 & +0.17 [$-$0.03, +0.37] &  & $\times$ \\
Qwen & MPOA & +10.1 [+7.8, +12.5] & +13.1 & $-$3.1 [$-$6.3, +0.2] & +1.39 [+1.04, +1.73] & +0.99 & +0.39 [+0.01, +0.77] & $\times$ & $\times$ \\
Qwen & SOM & $-$4.7 [$-$8.0, $-$1.4] & +13.1 & $-$17.8 [$-$21.5, $-$14.2] & $-$0.20 [$-$0.33, $-$0.06] & +0.99 & $-$1.19 [$-$1.40, $-$0.97] &  &  \\
Qwen & ARA & +34.4 [+31.2, +37.5] & +13.1 & +21.3 [+17.1, +25.4] & +2.35 [+2.08, +2.62] & +0.99 & +1.36 [+1.05, +1.68] &  &  \\
\midrule
Gemma & DiM & +16.4 [+13.5, +19.3] & +1.0 & +15.3 [+13.0, +17.8] & +1.18 [+0.96, +1.40] & +0.10 & +1.07 [+0.90, +1.25] &  &  \\
Gemma & MPOA & $-$4.7 [$-$8.6, $-$0.9] & +1.0 & $-$5.8 [$-$9.7, $-$1.7] & $-$0.19 [$-$0.35, $-$0.04] & +0.10 & $-$0.30 [$-$0.55, $-$0.04] &  &  \\
Gemma & SOM & +2.6 [$-$1.1, +6.2] & +1.0 & +1.5 [$-$1.9, +4.9] & +0.12 [$-$0.05, +0.28] & +0.10 & +0.01 [$-$0.22, +0.24] &  &  \\
Gemma & ARA & $-$26.0 [$-$28.9, $-$23.3] & +1.0 & $-$27.1 [$-$30.3, $-$23.6] & $-$4.07 [$-$4.87, $-$3.27] & +0.10 & $-$4.17 [$-$5.00, $-$3.34] &  &  \\
\midrule
OLMo & DiM & +34.8 [+31.5, +37.9] & +0.8 & +34.0 [+30.5, +37.6] & +1.45 [+1.31, +1.60] & +0.06 & +1.39 [+1.22, +1.56] &  &  \\
OLMo & MPOA & +61.4 [+58.2, +64.4] & +0.8 & +60.6 [+57.1, +64.1] & +3.91 [+3.56, +4.25] & +0.06 & +3.85 [+3.49, +4.21] &  &  \\
OLMo & SOM & +14.4 [+11.5, +17.3] & +0.8 & +13.6 [+10.4, +16.7] & +0.64 [+0.51, +0.78] & +0.06 & +0.58 [+0.42, +0.75] &  &  \\
OLMo & ARA & $-$2.5 [$-$4.6, $-$0.3] & +0.8 & $-$3.3 [$-$5.8, $-$0.9] & $-$0.23 [$-$0.43, $-$0.03] & +0.06 & $-$0.29 [$-$0.52, $-$0.07] &  &  \\
\midrule
Mistral Small & DiM & $-$4.3 [$-$7.4, $-$1.2] & +17.9 & $-$22.2 [$-$26.1, $-$18.4] & $-$0.20 [$-$0.35, $-$0.06] & +0.80 & $-$1.00 [$-$1.18, $-$0.82] &  &  \\
Mistral Small & MPOA & $-$2.9 [$-$4.7, $-$1.2] & +17.9 & $-$20.8 [$-$24.0, $-$17.7] & $-$0.57 [$-$0.92, $-$0.22] & +0.80 & $-$1.37 [$-$1.74, $-$1.00] &  &  \\
Mistral Small & SOM & +3.5 [+1.7, +5.3] & +17.9 & $-$14.4 [$-$17.7, $-$11.1] & +0.70 [+0.34, +1.06] & +0.80 & $-$0.10 [$-$0.47, +0.28] &  & $\times$ \\
Mistral Small & ARA & $-$0.7 [$-$2.1, +0.6] & +17.9 & $-$18.6 [$-$21.6, $-$15.7] & $-$0.17 [$-$0.45, +0.12] & +0.80 & $-$0.96 [$-$1.28, $-$0.65] &  &  \\
\midrule
Small pair & DiM & +34.1 [+30.5, +37.5] & +16.4 & +17.7 [+13.5, +21.8] & +1.54 [+1.36, +1.71] & +1.09 & +0.45 [+0.22, +0.68] &  &  \\
Small pair & MPOA & +4.2 [+1.6, +6.9] & +16.4 & $-$12.1 [$-$15.7, $-$8.3] & +0.32 [+0.12, +0.52] & +1.09 & $-$0.77 [$-$1.03, $-$0.51] &  &  \\
Small pair & SOM & +9.7 [+6.9, +12.4] & +16.4 & $-$6.7 [$-$10.4, $-$3.0] & +0.70 [+0.50, +0.90] & +1.09 & $-$0.39 [$-$0.65, $-$0.13] &  &  \\
Small pair & J-space & +21.2 [+18.1, +24.3] & +16.4 & +4.8 [+1.3, +8.3] & +0.93 [+0.79, +1.07] & +1.09 & $-$0.16 [$-$0.35, +0.04] & $\times$ & $\times$ \\
Small pair & ARA & +14.2 [+11.6, +16.9] & +16.4 & $-$2.2 [$-$6.0, +1.6] & +1.55 [+1.23, +1.86] & +1.09 & +0.46 [+0.10, +0.82] & $\times$ & $\times$ \\
\bottomrule
\end{tabular}
\end{table*}

%% file: sections/08h_item_source_harm.tex
\subsection{Item Source}
\label{app:item-source}

The 972 harmful held-out prompts come from four public sets (JailbreakBench,
XSTest unsafe, StrongREJECT and HarmBench; 772 prompts) and the code-as-weapon
bank (200). H1 and H2 were re-run, with the same models, tests and Holm
families as in \S\ref{sec:res-h1}, on the public sets together, on the
code-as-weapon bank and on each set separately. The originals' generation
difference is the item-weighted sum of the differences on the sets, so the
contribution of the code items is $200/972$ times their difference
(Table~\ref{tab:item-source}a). The code items carried most of the Qwen
originals' difference and 52\% of the mean difference of the four main pairs;
in Gemma, a decrease on the public sets ($-$5.7~pp) and an increase on the
code items (+27.0~pp) cancelled; and in Mistral Small the difference lay in
the public sets. Gemma's direction-based effects also lay in the code items
(DiM, MPOA and SOM +65.5, +35.0 and +47.5~pp on code items, against +3.6,
$-$15.0 and $-$9.1~pp on the public sets), whereas the largest effect of each
other main pair lay in the public sets (Qwen ARA +41.3 vs.\ +7.5~pp, OLMo MPOA
+75.6 vs.\ +6.5~pp).

On the 772 public items the pooled conclusions were unchanged
(Table~\ref{tab:item-source}b): the baseline-adjusted direction-based effect
was $-$0.13 log-odds [$-$0.28, +0.03], the H2 interaction $-$0.46 [$-$0.58,
$-$0.34], with the same pairs in the predicted (Gemma, OLMo) and opposite
(Qwen, small pair) direction, and only the small pair had every method higher
on the newer model. Without the 44 prompts that overlap the search dataset,
the public-set estimates were $-$0.08 [$-$0.23, +0.08] (adjusted
direction-based), $-$0.58 [$-$0.76, $-$0.39] (adjusted ARA) and $-$0.50
[$-$0.63, $-$0.38] (H2). On the code items alone the adjusted direction-based
effect was positive (+1.39 [+1.08, +1.70]), but 15 of the 32 main-study
modified models refused none of the 200 code items, so code-only log-odds rest
on the few models that still refused some of them. The overlap check is
lexical; semantic near-duplicates and safety tuning of the newer models on the
public benchmarks were not tested.

\begin{table*}[t]
\centering\scriptsize
\setlength{\tabcolsep}{2.6pt}
\caption{Item source. (a) Originals' generation difference (newer minus older unmodified model, WildGuard refusal,
pp, item-bootstrap 95\% intervals) on all 972 harmful prompts, the four public sets and the code-as-weapon bank, and
the share of the 972-item difference contributed by the code items (shown only where the total excludes zero).
(b) Pooled main-study GEE terms (log-odds, 95\% intervals) by item subset (Public: the four public sets; Code:
code-as-weapon bank; JBB: JailbreakBench; XSTest: its unsafe prompts): originals' difference, generation effect
of the direction-based class raw and baseline-adjusted, ARA raw and adjusted, H2 interaction ($p_\mathrm{Holm}$ over
the subset's H2 family) and the Holm tally of the 16 main-pair H1 tests (newer higher / lower / not significant). The
generation $\times$ family Wald $\chi^2$ (3 df) was 547 (all items), 793 (public sets) and 313 (XSTest); it was not
estimable on the other subsets, where some model refused 0\% or 100\% of the items.}
\label{tab:item-source}
\begin{tabular}{@{}llllll@{}}
\multicolumn{6}{@{}l}{\emph{(a) Originals' generation difference by item source, pp}} \\
\toprule
Pair & All 972 & Public sets (772) & Code-as-weapon (200) & Code contribution & Code share \\
\midrule
Qwen & +13.1 [+10.8, +15.4] & +5.3 [+3.5, +7.2] & +43.2 [+36.1, +50.2] & +8.9 [+7.1, +10.7] & 68\% [59, 77] \\
Gemma & +1.0 [$-$1.3, +3.5] & $-$5.7 [$-$7.5, $-$4.0] & +27.0 [+17.9, +35.8] & +5.6 [+3.6, +7.6] & -- \\
OLMo & +0.8 [$-$0.5, +2.2] & $-$0.3 [$-$1.5, +1.0] & +5.0 [+0.5, +9.8] & +1.0 [+0.1, +2.1] & -- \\
Mistral Small & +17.9 [+15.2, +20.7] & +20.5 [+17.2, +23.8] & +8.0 [+4.6, +12.1] & +1.6 [+0.9, +2.5] & 9\% [5, 14] \\
Small pair & +16.4 [+13.9, +18.8] & +9.5 [+7.0, +11.8] & +43.0 [+36.3, +50.0] & +8.8 [+7.2, +10.6] & 54\% [46, 63] \\
Mean of 4 main pairs & +8.1 [+7.0, +9.3] & +5.0 [+3.8, +6.0] & +20.6 [+17.3, +23.8] & +4.2 [+3.4, +5.1] & 52\% [44, 60] \\
\bottomrule
\end{tabular}

\medskip
\setlength{\tabcolsep}{1.5pt}%
\begin{tabular}{@{}lllllllc@{}}
\multicolumn{8}{@{}l}{\emph{(b) Pooled main-study generation effects by item subset, log-odds}} \\
\toprule
Items & Originals & Dir., raw & Dir., adj. & ARA, raw & ARA, adj. & H2 ($p_\mathrm{Holm}$) & H1 tally \\
\midrule
All 972 & +0.58 [+0.49, +0.66] & +0.62 [+0.57, +0.68] & +0.03 [$-$0.05, +0.11] & +0.12 [+0.00, +0.24] & $-$0.46 [$-$0.58, $-$0.34] & $-$0.50 ($<$0.001) & 8/4/4 \\
Public (772) & +0.69 [+0.52, +0.85] & +0.56 [+0.50, +0.62] & $-$0.13 [$-$0.28, +0.03] & +0.10 [$-$0.02, +0.22] & $-$0.59 [$-$0.77, $-$0.41] & $-$0.46 ($<$0.001) & 8/6/2 \\
Code (200) & +1.24 [+0.98, +1.49] & +2.96 [+2.48, +3.43] & +1.39 [+1.08, +1.70] & +1.75 [+0.42, +3.08] & +0.19 [$-$0.59, +0.96] & $-$1.20 (0.129) & 8/0/8 \\
JBB (100) & +0.70 [+0.21, +1.20] & +0.53 [+0.36, +0.71] & $-$0.17 [$-$0.63, +0.29] & +0.69 [+0.14, +1.24] & $-$0.04 [$-$0.66, +0.59] & +0.16 (0.577) & 5/2/9 \\
XSTest (200) & $-$0.09 [$-$0.37, +0.19] & +0.22 [+0.12, +0.33] & +0.31 [+0.03, +0.59] & $-$0.46 [$-$0.63, $-$0.29] & $-$0.35 [$-$0.65, $-$0.05] & $-$0.68 ($<$0.001) & 5/3/8 \\
StrongREJECT (313) & +1.39 [+1.03, +1.75] & +0.80 [+0.70, +0.89] & $-$0.59 [$-$0.93, $-$0.25] & +0.34 [+0.14, +0.53] & $-$1.05 [$-$1.43, $-$0.68] & $-$0.46 ($<$0.001) & 7/1/8 \\
HarmBench (159) & +1.00 [+0.69, +1.32] & +0.63 [+0.51, +0.76] & $-$0.37 [$-$0.68, $-$0.06] & +0.47 [+0.14, +0.81] & $-$0.53 [$-$0.97, $-$0.09] & $-$0.16 (0.524) & 6/2/8 \\
\bottomrule
\end{tabular}
\end{table*}

\subsection{Harmful Compliance as the Outcome}
\label{app:harm}

H1 and H2 were re-run on all 972 harmful prompts with two harmful-compliance
labels, the HarmBench classifier and WildGuard's harmful-response label, coded
as not harmful so that a positive generation effect again means that removal
left less on the newer model (Table~\ref{tab:harm}). The pooled conclusions
were the same as for refusal. Across the 42 ablated seed-1 models, the share
of non-refused responses that the HarmBench classifier labelled harmful ranged
from 48.1\% (OLMo-2-32B DiM) to 88.8\%, and the WildGuard share from 15.8\% to
75.9\%. Of the 21 per-pair H1 tests, 5 changed Holm decision or sign under
HarmBench (Qwen SOM, Gemma MPOA and Gemma SOM became significantly higher on
the newer model; Mistral Small DiM lost significance; Mistral Small ARA became
significantly lower) and 6 under WildGuard (Gemma MPOA and SOM became
significantly higher; Qwen SOM and Mistral Small DiM and MPOA lost
significance; OLMo ARA changed sign without being significant). In addition,
the pooled raw ARA effect became significant under both labels, and the
small-pair H2 interaction lost significance under WildGuard. Per pair, the
harmful-compliance rates themselves are given in Table~\ref{tab:harm-cells}.

\begin{table*}[t]
\centering\scriptsize
\setlength{\tabcolsep}{2.6pt}
\caption{Harmful compliance as the outcome. (a) Pooled main-study GEE terms (log-odds, 95\% intervals) for WildGuard
refusal (primary) and for two not-harmful outcomes (1 minus harmful compliance by the HarmBench classifier or by
WildGuard), so that signs match; generation $\times$ family $\chi^2$ with 3 df; Holm tally of the 16 main-pair H1
tests; pairs in which every method was significantly higher on the newer model.
(b) Per-pair H2 interaction (log-odds, $p_\mathrm{Holm}$) by outcome. (c) Share of responses that WildGuard does not
label as refusal and that are harmful, pooled over cell groups (prompt-cluster bootstrap 95\% intervals).}
\label{tab:harm}
\begin{tabular}{@{}llll@{}}
\multicolumn{4}{@{}l}{\emph{(a) Pooled main-study terms by outcome, log-odds}} \\
\toprule
Term & WG refusal (primary) & HB not harmful & WG not harmful \\
\midrule
Originals' difference & +0.58 [+0.49, +0.66] & +0.65 [+0.58, +0.72] & +0.72 [+0.62, +0.83] \\
Direction, raw & +0.62 [+0.57, +0.68] & +0.70 [+0.65, +0.74] & +0.69 [+0.64, +0.73] \\
Direction, adjusted & +0.03 [$-$0.05, +0.11] & +0.05 [$-$0.02, +0.11] & $-$0.03 [$-$0.13, +0.07] \\
ARA, raw ($p_\mathrm{Holm}$) & +0.12 [+0.00, +0.24] (0.135) & +0.21 [+0.14, +0.28] ($<$0.001) & +0.24 [+0.18, +0.30] ($<$0.001) \\
ARA, adjusted & $-$0.46 [$-$0.58, $-$0.34] & $-$0.44 [$-$0.53, $-$0.36] & $-$0.48 [$-$0.60, $-$0.37] \\
H2 interaction ($p_\mathrm{Holm}$) & $-$0.50 [$-$0.62, $-$0.38] ($<$0.001) & $-$0.49 [$-$0.56, $-$0.41] ($<$0.001) & $-$0.45 [$-$0.52, $-$0.38] ($<$0.001) \\
Generation $\times$ family $\chi^2$ & 547 & 450 & 447 \\
H1 tally (higher / lower / n.s.) & 8/4/4 & 11/3/2 & 10/1/5 \\
Every method higher & small pair & Qwen, small pair & small pair \\
\bottomrule
\end{tabular}

\medskip
\begin{tabular}{@{}llll@{}}
\multicolumn{4}{@{}l}{\emph{(b) Per-pair H2 interaction, log-odds ($p_\mathrm{Holm}$)}} \\
\toprule
Pair & WG refusal & HB not harmful & WG not harmful \\
\midrule
Qwen & +1.89 ($<$0.001) & +1.47 ($<$0.001) & +1.43 ($<$0.001) \\
Gemma & $-$4.27 ($<$0.001) & $-$2.17 ($<$0.001) & $-$1.92 ($<$0.001) \\
OLMo & $-$1.87 ($<$0.001) & $-$1.26 ($<$0.001) & $-$1.44 ($<$0.001) \\
Mistral Small & $-$0.06 (0.696) & $-$0.15 (0.083) & $-$0.06 (0.732) \\
Small pair & +0.76 ($<$0.001) & +0.27 (0.004) & +0.06 (0.732) \\
\bottomrule
\end{tabular}

\medskip
\begin{tabular}{@{}lllll@{}}
\multicolumn{5}{@{}l}{\emph{(c) Harmful among non-refused responses, \%}} \\
\toprule
Cells & WG refusal & HB harmful & HB among not refused & WG among not refused \\
\midrule
Originals (10) & 80.2 & 16.0 & 79.0 [75.5, 82.1] & 29.6 [25.6, 33.8] \\
Ablated, main (32) & 32.3 & 51.0 & 75.1 [73.4, 76.9] & 55.8 [53.3, 58.1] \\
Ablated, small (10) & 29.1 & 54.4 & 76.7 [74.7, 78.7] & 58.5 [55.9, 61.1] \\
$\gamma$-MPOA (4) & 29.9 & 56.2 & 79.9 [78.0, 81.8] & 60.2 [57.6, 62.7] \\
Community (8) & 6.4 & 75.5 & 80.5 [79.0, 82.1] & 66.7 [64.3, 68.9] \\
Seed 2 (10) & 29.2 & 56.4 & 79.5 [77.7, 81.4] & 61.3 [58.6, 63.9] \\
\bottomrule
\end{tabular}
\end{table*}

\begin{table}[t]
\centering\scriptsize
\setlength{\tabcolsep}{2.5pt}
\caption{Refusal and harmful compliance per pair and method, older $\to$ newer model (\%, 972 harmful prompts). Effect:
newer minus older on the not-refused (WG refusal) or not-harmful scale, pp, so that a positive effect means that
removal left less on the newer model; * Holm-significant within that outcome's H1 family.}
\label{tab:harm-cells}
\begin{tabular}{@{}llrrr@{}}
\toprule
Pair & Method & WG refusal: effect & HB harmful: effect & WG harmful: effect \\
\midrule
Qwen & Original & 76.8$\to$89.9: +13.1 & 19.8$\to$6.3: +13.5 & 7.1$\to$1.4: +5.7 \\
 & DiM & 47.4$\to$74.3: +26.9* & 40.3$\to$16.4: +24.0* & 26.0$\to$6.0: +20.1* \\
 & MPOA & 3.9$\to$14.0: +10.1* & 85.3$\to$48.6: +36.7* & 72.9$\to$34.2: +38.8* \\
 & SOM & 42.7$\to$38.0: $-$4.7* & 44.5$\to$39.9: +4.6* & 30.1$\to$27.8: +2.3 \\
 & ARA & 6.1$\to$40.4: +34.4* & 82.3$\to$30.6: +51.7* & 69.1$\to$17.2: +52.0* \\
\midrule
Gemma & Original & 88.3$\to$89.3: +1.0 & 10.7$\to$6.9: +3.8 & 2.1$\to$1.8: +0.3 \\
 & DiM & 73.8$\to$90.1: +16.4* & 22.3$\to$6.5: +15.8* & 10.0$\to$1.5: +8.4* \\
 & MPOA & 45.2$\to$40.4: $-$4.7 & 45.5$\to$31.6: +13.9* & 32.5$\to$22.6: +9.9* \\
 & SOM & 65.2$\to$67.8: +2.6 & 28.8$\to$17.5: +11.3* & 15.5$\to$8.6: +6.9* \\
 & ARA & 26.6$\to$0.6: $-$26.0* & 50.1$\to$80.9: $-$30.8* & 35.0$\to$65.2: $-$30.2* \\
\midrule
OLMo & Original & 83.2$\to$84.1: +0.8 & 15.0$\to$12.3: +2.7 & 4.0$\to$3.2: +0.8 \\
 & DiM & 33.5$\to$68.3: +34.8* & 32.0$\to$20.4: +11.6* & 18.2$\to$7.2: +11.0* \\
 & MPOA & 3.6$\to$65.0: +61.4* & 83.5$\to$21.6: +61.9* & 69.4$\to$9.4: +60.1* \\
 & SOM & 27.3$\to$41.7: +14.4* & 53.5$\to$39.6: +13.9* & 40.5$\to$26.6: +13.9* \\
 & ARA & 13.3$\to$10.8: $-$2.5 & 60.7$\to$61.2: $-$0.5 & 45.6$\to$44.5: +1.1 \\
\midrule
Mistral Small & Original & 55.7$\to$73.6: +17.9 & 35.5$\to$21.3: +14.2 & 19.8$\to$8.4: +11.3 \\
 & DiM & 32.8$\to$28.5: $-$4.3* & 53.4$\to$56.4: $-$3.0 & 38.9$\to$42.2: $-$3.3 \\
 & MPOA & 6.9$\to$4.0: $-$2.9* & 77.7$\to$82.5: $-$4.8* & 63.1$\to$65.6: $-$2.6 \\
 & SOM & 3.7$\to$7.2: +3.5* & 80.5$\to$76.7: +3.7* & 68.7$\to$61.3: +7.4* \\
 & ARA & 4.9$\to$4.2: $-$0.7 & 78.5$\to$81.9: $-$3.4* & 66.0$\to$66.9: $-$0.8 \\
\midrule
Small pair & Original & 72.1$\to$88.5: +16.4 & 23.3$\to$9.5: +13.8 & 9.3$\to$1.6: +7.6 \\
 & DiM & 20.6$\to$54.6: +34.1* & 64.0$\to$32.6: +31.4* & 50.2$\to$21.3: +28.9* \\
 & MPOA & 13.8$\to$18.0: +4.2* & 71.5$\to$55.9: +15.6* & 58.0$\to$44.1: +13.9* \\
 & SOM & 12.2$\to$21.9: +9.7* & 73.1$\to$56.4: +16.8* & 59.2$\to$44.2: +14.9* \\
 & J-space & 52.6$\to$73.8: +21.2* & 36.3$\to$18.9: +17.4* & 22.8$\to$5.5: +17.4* \\
 & ARA & 4.7$\to$18.9: +14.2* & 79.0$\to$55.9: +23.1* & 65.1$\to$43.9: +21.2* \\
\bottomrule
\end{tabular}
\end{table}

%% file: sections/08i_revision_details.tex
\subsection{Detail Moved from the Results}
\label{app:moved-results}

This subsection keeps the detailed Results text of the previous draft, which
the main text now summarises, together with five secondary figures. Numbers and
tests are unchanged.

\subsubsection*{H1: Residual Refusal in the Newer Model}

Table~\ref{tab:pair-effects} gives the per-pair generation effects by method
class, and Figure~\ref{fig:heldout-generation} and Table~\ref{tab:h1-heldout}
give them per method. After direction-based removal (mean of DiM, MPOA and
SOM), residual refusal was higher on the newer model by +10.7~pp in the Qwen
pair, +4.7~pp in Gemma, +36.9~pp in OLMo and $-$1.2~pp in Mistral Small, and
by +17.3~pp in the small pair (mean including J-space). After ARA the changes
were +34.4, $-$26.0, $-$2.5, $-$0.7 and +14.2~pp. The generation effect
differed between pairs (prompt-level generation $\times$ pair Wald $\chi^2=547.4$, 3
df, $p<0.001$; $\chi^2$ of 277--551 and $p\le10^{-59}$ for every judge, with
and without the 44 prompts that overlap the search dataset). With run-to-run (seed) variance added, the per-pair half-widths grew
to about 15~pp for the direction-based class and 25~pp for ARA, and 14 of the
40 main-pair estimand $\times$ scale intervals excluded zero (32 with prompt
sampling only). Of the ten raw effects on the pp scale (five pairs, two
classes), four excluded zero: the OLMo (+36.9~pp
[+22.1, +51.7]) and small-pair (+17.3~pp [+4.4, +30.1]) direction-based
increases, the Qwen ARA increase (+34.4~pp [+8.8, +59.9]) and the Gemma ARA
decrease ($-$26.0~pp [$-$51.6, $-$0.5]).

The unedited originals already differed by +13.1 (Qwen), +17.9 (Mistral
Small) and +16.4~pp (small pair), but not in Gemma ($p=0.46$) or OLMo
($p=0.30$). Relative to this baseline difference, a contrast added after H1
was stated and therefore exploratory, the direction-based effect was $-$2.4
(Qwen), +3.7 (Gemma), +36.0 (OLMo), $-$19.2 (Mistral Small) and +0.9~pp (small
pair), and the ARA effect +21.3, $-$27.1, $-$3.3, $-$18.6 and $-$2.2~pp. The
two scales of the adjustment disagreed in places. Among the 21 pair $\times$
method cells, the adjusted effect had opposite signs in pp and log-odds for
three (Qwen MPOA, $-$3.1~pp but +0.39 log-odds; small-pair J-space and ARA),
and its significance differed in five. At class level the signs differed for
the small pair's direction-based (+0.9~pp, $-$0.30 log-odds) and ARA effects
and for the H2 contrasts of Mistral Small and the small pair. Near 0\%
refusal the log-odds scale magnifies differences: Gemma ARA ($26.6\%$ to
$0.6\%$) was $-$26.0~pp but $-$4.07 log-odds, the largest log-odds effect in
the main study.

Pooled over the four main pairs with the pair as the unit (random effects,
Hartung--Knapp--Sidik--Jonkman interval), the direction-based generation
effect was +12.8~pp [$-$13.9, +39.5] (+0.55 log-odds [$-$0.66, +1.76]) and,
beyond the originals' difference, +4.6~pp [$-$32.2, +41.3] (+0.06 log-odds
[$-$1.67, +1.80]; 95\% prediction interval for a new pair $-$5.90 to +6.02);
for ARA the adjusted effect was $-$6.9~pp [$-$40.7, +26.8] ($-$0.96 log-odds
[$-$4.59, +2.68]). Heterogeneity between pairs was almost total ($I^2$
99--100\%; between-pair SD of the adjusted direction-based effect 22.9~pp,
against a seed SE of 7.5~pp and a prompt SE of 1.3--1.5~pp per pair), so
adding seed variance moved these limits by at most 0.1~pp and 0.07 log-odds,
and none of the ten pooled estimand $\times$ scale intervals excluded zero.
With the prompt as the unit (GEE, prompt sampling only) the same quantities
were narrow: the direction-based effect was +0.62 log-odds [+0.57, +0.68]
($p_\mathrm{Holm}<0.001$) raw and +0.03 [$-$0.05, +0.11] ($p=0.529$) adjusted,
the ARA effect +0.12 [+0.00, +0.24] ($p_\mathrm{Holm}=0.135$) raw and $-$0.46
[$-$0.58, $-$0.34] ($p<0.001$) adjusted, and the originals' difference +0.58
[+0.49, +0.66] ($p<0.001$); the planned mixed model gave unadjusted effects of
+0.86 [+0.82, +0.90] and +0.15 [+0.04, +0.26]. Leaving one family out moved
the adjusted direction-based GEE effect between $-$0.51 (without OLMo) and
+0.33 (without Mistral Small).

All models were selected under the same cap on first-token KL divergence
($\le 0.1$), but the cap did not constrain every run
(Appendix~\ref{app:kl-cap}). It bound, in the sense that some trial above 0.1
had fewer dev refusals than the selected trial, in 15 of the 24 main-study
direction-based runs and in all 8 ARA runs, which reached 0 dev refusals above
the cap (maximum KL 10.9--33.2); to stay under it, the median binding run gave
up 24 (direction-based) and 22 (ARA) dev refusals. Four direction-based runs
never sampled a trial above 0.1 (Qwen3.8-27B MPOA, Olmo-3.1-32B MPOA and SOM,
Mistral-Small-2409 MPOA) and four sampled one (OLMo-2-32B DiM, MPOA and SOM,
Qwen3.8-27B SOM). In these eight runs the selected maximum weights were not at
the upper bound of Heretic's search range (0 of 8), and in 7 of 8 the trials
at that bound had no fewer dev refusals than the selected trial (in the
eighth, OLMo-2-32B SOM, the better one had KL 0.110), so the upper bound of
the maximum weights did not limit their residual refusal; another strength
parameter (a minimum weight or minimum-weight distance) was at the strong end
of its range in 4 of the 8. Refitted on the pair $\times$ method
units in which the cap bound in both cells (5 direction-based, 4 ARA), the
direction-based effect was +0.32 log-odds [+0.24, +0.41] raw and $-$0.24
[$-$0.33, $-$0.14] adjusted; on the 7 direction-based units in which it did
not bind in at least one cell, +1.03 [+0.95, +1.11] and +0.22 [+0.12,
+0.32]. The two subsets also differ in family (cap-bound units: one Qwen, three
Gemma and one Mistral Small unit; others: two Qwen, three OLMo and two Mistral
Small units), so this split does not separate the cap from the family.

The per-pair effects can also be compared with the run-to-run variation of
the search. The seed-1 and seed-2 models of the ten small-pair cells gave a
run-to-run standard deviation of a cell's held-out refusal of 9.2~pp [6.4,
16.1] (\S\ref{sec:setup-stats}), so the difference between two independently
optimised cells has a standard deviation of 12.9~pp, and an effect larger than
25.9~pp (2~SD) exceeds seed noise. Of the 16 main-study per-method effects, 5
exceeded this band (Table~\ref{tab:h1-heldout}, last column): four increases,
Qwen DiM (+26.9~pp [+23.8, +29.8]), Qwen ARA (+34.4~pp [+31.2, +37.5]), OLMo
DiM (+34.8~pp [+31.5, +37.9]) and OLMo MPOA (3.6\% [2.6, 5.0] to 65.0\%
[62.0, 68.0]; +61.4~pp [+58.2, +64.4]), and one decrease, Gemma ARA (26.6\%
[24.0, 29.5] to 0.6\% [0.3, 1.3]; $-$26.0~pp [$-$28.9, $-$23.3]), which lay
0.1~pp beyond the band. The other 11, from $-$4.7 to +16.4~pp, lay within
seed noise. Across the 95\% interval of the standard deviation, between 1 and
5 effects exceeded the band; of the baseline-adjusted effects, 3 exceeded it
(OLMo DiM +34.0, OLMo MPOA +60.6 and Gemma ARA $-$27.1~pp). In 3 of the ten
seed pairs (Qwen3.5-4B DiM, SOM and ARA) the seed-2 run also used a generation
batch size of 64 instead of 128, after running out of memory at 128. Without
these three pairs the standard deviation was 7.2~pp [4.8, 14.6] and the
threshold 20.4~pp; the same 5 raw effects exceeded it, and the
baseline-adjusted effects that did so rose from 3 to 6 (adding Qwen ARA
+21.3, Mistral Small DiM $-$22.2 and MPOA $-$20.8~pp). The band assumes that
the seed variability of the 4B and 7B small-pair models transfers to the
22--32B main-study models, each optimised once.

With the prompt as the unit of inference (exact McNemar, Holm over 23 tests),
the newer model refused significantly more in 8 of the 16 main-study
comparisons, significantly less in 4 and not significantly differently in 4;
4 of the 8 increases and 3 of the 4 decreases lay within seed noise, and no
main-study pair was higher for every method, as H1 required. Gemma-4-31B
DiM refused more than its own original (90.1\% vs.\ 89.3\%; +0.8~pp [+0.1,
+1.6], $p_\mathrm{Holm}=0.039$ over the 42 edit-versus-original tests), a
failed edit. In the small pair all five methods refused more on the newer
model with seed~1 (+4.2 to +34.1~pp, all $p_\mathrm{Holm}\le 0.019$).

Under Qwen3Guard the pooled ARA generation effect was +0.21
[+0.10, +0.32] and under full-response keyword matching +0.57 [+0.47, +0.67]
(both $p_\mathrm{Holm}<0.001$). Excluding the 44 held-out prompts that match a prompt of the
search dataset (\texttt{harmful\_behaviors}) removed significance for Qwen SOM
and Mistral Small DiM (all effects changed by $\le$1.9~pp). No main-study pair
met H1 under Qwen3Guard or full-response keyword matching; with the 256-token
keyword window, all four OLMo methods were higher on the newer model (OLMo ARA
+5.2~pp [+2.2, +8.3], $p_\mathrm{Holm}=0.009$).

\paragraph{Item source.} About half of the originals' generation difference
came from the code-as-weapon bank (200 of the 972 items;
Appendix~\ref{app:item-source}). Averaged over the four main pairs, the
originals differed by +5.0~pp [+3.8, +6.0] on the four public sets and by
+20.6~pp [+17.3, +23.8] on the code items, which contributed 52\% [44, 60] of
the mean difference of +8.1~pp (Qwen 68\% [59, 77], Mistral Small 9\% [5,
14]). Gemma's direction-based effects also lay in the code items (DiM, MPOA
and SOM +65.5, +35.0 and +47.5~pp, against +3.6, $-$15.0 and $-$9.1~pp on the
public sets). On the 772 public items alone the conclusions were unchanged:
the baseline-adjusted direction-based GEE effect was $-$0.13 [$-$0.28,
+0.03], only the small pair was higher on the newer model for every method,
the Holm tally was 8 higher, 6 lower and 2 not significant, and the H2
interaction was $-$0.46 [$-$0.58, $-$0.34], in the predicted direction for
Gemma and OLMo and opposite for Qwen and the small pair.

\subsubsection*{H2: Generation $\times$ Edit Class}

Per pair, ARA lost less effectiveness than the direction-based methods in
Gemma (H2 contrast, ARA minus direction mean, $-$30.8~pp) and OLMo
($-$39.3~pp), more in Qwen (+23.6~pp), and about as much in Mistral Small
(+0.5~pp) and the small pair ($-$3.1~pp; Tables~\ref{tab:h2-contrasts}
and~\ref{tab:pair-effects}). In log-odds the per-pair interactions were
$-$4.27 (Gemma), $-$1.87 (OLMo), +1.89 (Qwen) and +0.76 (small pair, opposite
in sign to its risk difference), all $p_\mathrm{Holm}<0.001$, and $-$0.06 for
Mistral Small ($p_\mathrm{Holm}=0.696$); the pairs differed (generation
$\times$ class $\times$ family $\chi^2=430.4$, 3 df, $p<0.001$). With seed
variance added, only the Gemma ($-$30.8~pp [$-$60.2, $-$1.3]) and OLMo
($-$39.3~pp [$-$68.8, $-$9.9]) contrasts excluded zero. Pooled with the pair
as the unit, the contrast was $-$11.5~pp [$-$57.6, +34.6] ($-$1.05 log-odds
[$-$5.21, +3.11]). With the prompt as the unit, the pooled GEE interaction was
$-$0.50 log-odds [$-$0.62, $-$0.38] ($p_\mathrm{Holm}<0.001$, Holm over 6
tests), and the planned mixed model gave $-$0.71 [$-$0.81, $-$0.60], both in
the predicted direction. Leaving out Gemma reversed the pooled interaction
(+0.20 [+0.05, +0.36], $p=0.008$), leaving out OLMo removed it (0.00
[$-$0.15, +0.15], $p=0.98$), and under keyword scoring it was +0.02 [$-$0.08,
+0.12] ($p_\mathrm{Holm}=1.000$). Because the KL cap bound in every ARA run
but in only 15 of the 24 direction-based runs, H2 compares a class whose
selection was always set by the cap with one whose selection was set by it in
part; on the units in which the cap bound in both cells, the interaction was
$-$0.21 [$-$0.33, $-$0.09].

Three properties of the ARA cells bear on these contrasts. First, the released
ARA model is rebuilt by refitting its adapter (L-BFGS) from the stored trial
parameters, not copied from the search. For the Qwen and Gemma pairs this
refit ran on a different GPU (RTX PRO 6000) from the search (H200), and these
four ARA exports moved by +6, +9, +5 and $-$7 dev keyword refusals when
re-scored, whereas the four refitted on the search machine (OLMo, Mistral
Small) moved by 0 to 2 (more than two refusals: 4 of 4 vs.\ 0 of 4, Fisher
$p=0.029$; other methods 3 of 14 vs.\ 1 of 14, $p=0.60$;
Appendix~\ref{app:dev-rescore}). Held-out results were measured on the
released models, so they describe those models, but the link between the
searched trial and the released model is weaker for these four. Second, the
Gemma-4-31B ARA export, behind the largest H2 contrast, re-scored at a
first-token KL of 0.107, above the cap (0.0845 during the search). Third, two
seed-1 ARA runs on Qwen3.8-27B selected trials with 59 and 63 of 100 dev
keyword refusals, whereas heretic-org reports 0 of 100 at KL 0.054 for a
model built with the same defaults; this was not reproduced, the reported
trial was not run in this pipeline, and the released heretic-org model refused
26.4\% of the held-out prompts against 40.4\% for the ARA model selected here
($-$14.0~pp [$-$17.0, $-$11.0]). The Qwen ARA effect (+34.4~pp), which
reverses H2 in the Qwen pair, rests on this cell.

\subsubsection*{H3: Keyword Counts versus an Automatic Judge}

Table~\ref{tab:h3-windows} reports H3 for three keyword windows; Table~\ref{tab:h3-gap}
details the full-response comparison and Figure~\ref{fig:judge-agreement} shows the gap per model at the
100-token window and the pooled gap at each window. At
the window Heretic itself scores, the first 100 generated tokens, keyword
matching and WildGuard agreed on average: over the 42 ablated models the
keyword-minus-WildGuard gap was +0.5~pp [$-$1.0, +1.9] (main study +1.5~pp
[+0.0, +2.9]; small pair $-$2.6~pp [$-$4.3, $-$0.9]; originals +0.7~pp
[$-$0.1, +1.5]), and $-$0.1~pp [$-$1.6, +1.3] against WildGuard on the first
256 tokens. Per model the sign varied: keyword matching was significantly
higher than WildGuard in 14 of the 42 ablated models and significantly lower
in 11 (Holm over the 74 models). With longer windows the gap grew, to +3.8~pp
[+2.4, +5.2] at 256 tokens (both measures on the same prefix) and +8.4~pp [+6.9, +10.0] on the full response of
at most 512 tokens, where keyword matching was higher in 34 models (24
significant) and lower in 7 (3). Keyword refusal fell from 39.9\% on the full
response to 32.0\% at 100 tokens while WildGuard refusal was 31.5\%, so most
of the full-response excess came from markers that appear after the first 100
tokens. At 100 tokens the gap was +11.0~pp [+9.7, +12.4] on responses that
reached the 512-token cap and $-$10.4~pp [$-$12.2, $-$8.6] on responses that
ended earlier (full response +17.7 and $-$1.2~pp). The 100-token text was
obtained by re-tokenising the stored responses, which reproduced the stored
256-token keyword labels in 97,825 of 97,828 responses
(Appendix~\ref{app:trunc256}).

The structure of the gap was the same at every window. In the main study,
with the prompt as the unit, the 100-token gap was larger for ARA than for the
direction-based methods (+5.2~pp [+3.9, +6.4]), larger on the newer models
(+4.7~pp [+3.4, +5.9]) and largest for ARA on the newer models (generation
$\times$ class +8.1~pp [+6.0, +10.1]; all $p_\mathrm{Holm}<0.001$; full
response +6.0, +4.4 and +9.4~pp). With the model as the unit ($n=32$), none of
these terms was significant (class $p=0.17$, interaction $p=0.28$; full
response $p=0.151$ and 0.26). In the small pair the class difference was not
significant (100 tokens +1.1~pp, $p_\mathrm{Holm}=0.280$; full response
+0.8~pp [$-$1.5, +3.0], $p_\mathrm{Holm}=0.774$).

On the full response, keyword--WildGuard agreement fell from $\kappa=0.73$
[0.70, 0.75] in the originals to 0.40 [0.39, 0.42] (main study) and 0.33
[0.31, 0.35] (small pair) after ablation, whereas WildGuard and Qwen3Guard
agreed at 0.90 [0.88, 0.92] (originals) and 0.82 [0.80, 0.83] after ablation
(42 models; main study 0.82 [0.81, 0.84], small pair 0.81 [0.78, 0.83]). Over
the 42 ablated models, 7,503 responses were refusals by full-response keyword
matching only and 4,059 by WildGuard only; by string heuristics that were not
validated, 90.9\% of the keyword-only refusals matched only Heretic's safety
and disclaimer markers (such as ``illegal'' or ``harmful''), not a
first-person refusal or AI-identity phrase, and 60.0\% were truncated at 512
tokens. Judging the first 256 tokens instead of the full response changed
WildGuard refusal by +0.6~pp [+0.5, +0.7] (Appendix~\ref{app:trunc256}).

\subsubsection*{Secondary Outcomes}

\begin{figure*}[!htbp]
\centering
\includegraphics[width=\textwidth]{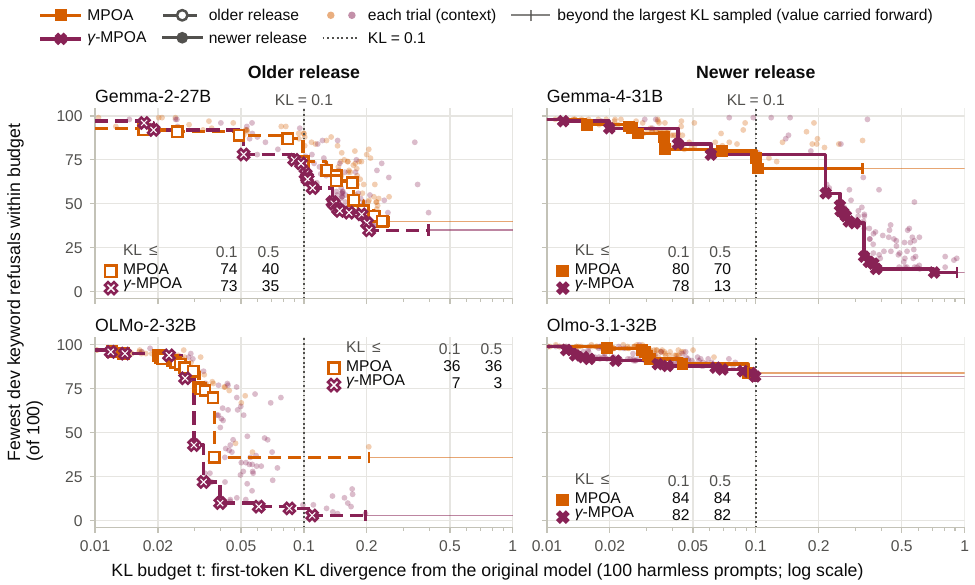}
\caption{\textbf{Gamma-corrected MPOA versus MPOA on the post-norm models (dev set).} One panel per model (columns: older and newer release; rows: Gemma, OLMo). A curve gives, for every KL budget $t$, the fewest dev keyword refusals (of 100) among the completed TPE trials of one run (100 trials, seed 1) with first-token KL divergence from the original at most $t$; it steps down at the run's Pareto-front trials (method markers; older release dashed and hollow, newer solid and filled). Past the largest KL sampled (end tick) the value is only carried forward (thin line). Faint dots: all completed trials; dotted line: the selection budget KL = 0.1; the small table (upper right for OLMo-2-32B, lower left elsewhere) reads each curve at KL $\le$ 0.1 and $\le$ 0.5. MPOA vs $\gamma$-MPOA: on Gemma-2-27B the correction changes little (74 vs 73 at KL $\le$ 0.1, 40 vs 35 at KL $\le$ 0.5). On Gemma-4-31B the two are close at KL $\le$ 0.1 (80 vs 78) but separate at larger budgets (70 vs 13 at KL $\le$ 0.5), a gap confounded with search coverage (MPOA placed 14 of 100 trials above KL 0.1, $\gamma$-MPOA 82): the MPOA curve is flat there because that region was barely explored, not because lower counts were shown to be unreachable. On OLMo-2-32B the correction lowers the count at the selection budget itself (36 vs 7 at KL $\le$ 0.1, 36 vs 3 at KL $\le$ 0.5), with similar search coverage (99 and 88 of 100 trials at KL $\le$ 0.1, none above 0.3), so this gap is not a coverage artefact. On Olmo-3.1-32B the correction changes little (84 vs 82 at KL $\le$ 0.1; neither search sampled a larger KL, so the KL $\le$ 0.5 values only repeat these). All counts are dev keyword refusals on the 100 prompts also used for optimisation and selection, not held-out refusal rates; one search per run (seed 1), budgets other than 0.1 read post hoc.}
\label{fig:gmpoa-dev}
\end{figure*}

\paragraph{$\gamma$-MPOA control.} Correcting MPOA for the post-norm output
gains (Figures~\ref{fig:gmpoa-dev} and~\ref{fig:gmpoa-heldout}) lowered
held-out refusal on all four post-norm models, by 3.1~pp [2.1, 4.4]
(OLMo-2-32B) to 19.9~pp [16.9, 22.7] (Gemma-4-31B; all $p_\mathrm{Holm}\le
0.002$). The OLMo generation effect remained +57.0~pp [+53.8, +60.1] (MPOA
+61.4~pp), the Gemma generation effect became $-$20.4~pp [$-$23.6, $-$17.1]
(MPOA $-$4.7~pp), and the pooled H2 interaction with $\gamma$-MPOA in place of
MPOA was $-$0.42 [$-$0.53, $-$0.30]. Gemma-4-31B $\gamma$-MPOA lost 21.4~pp of
mean capability at a dev KL of 0.061, with a teacher-forced divergence of
0.128 (MPOA 0.007), and OLMo-2-32B $\gamma$-MPOA lost 8.1~pp at 0.085
(divergence 0.196; MPOA 0.088); on Gemma-2-27B and Olmo-3.1-32B, where
$\gamma$-MPOA lowered refusal by 4.2 and 7.5~pp, no benchmark score differed
from the original's ($p_\mathrm{Holm}=1.0$).

\paragraph{Community releases.} Of eight public abliterations
(Figures~\ref{fig:community} and~\ref{fig:community-releases}), four refused
less than the best model selected here for the same checkpoint (all
$p_\mathrm{Holm}<0.001$), for example on Qwen3.8-27B (1.4\% [0.9, 2.4] vs.\
14.0\% [12.0, 16.3]; $-$12.6~pp [$-$14.8, $-$10.5]); two did not differ significantly and two refused more. The
four had teacher-forced divergences of 0.085--0.151 (best selected models
0.042--0.060) and mean capability changes of $-$0.94 to $-$0.15~pp. On three
of the five checkpoints with a public release (Qwen2.5-32B, Qwen3.8-27B,
Olmo-3.1-32B), a public release refused less than every model selected here
under the KL cap; on the other two (Gemma-4-31B, Mistral-Small-3.2) the best
release did not differ significantly from the best selected model.

\begin{figure*}[!htbp]
\centering
\includegraphics[width=\textwidth]{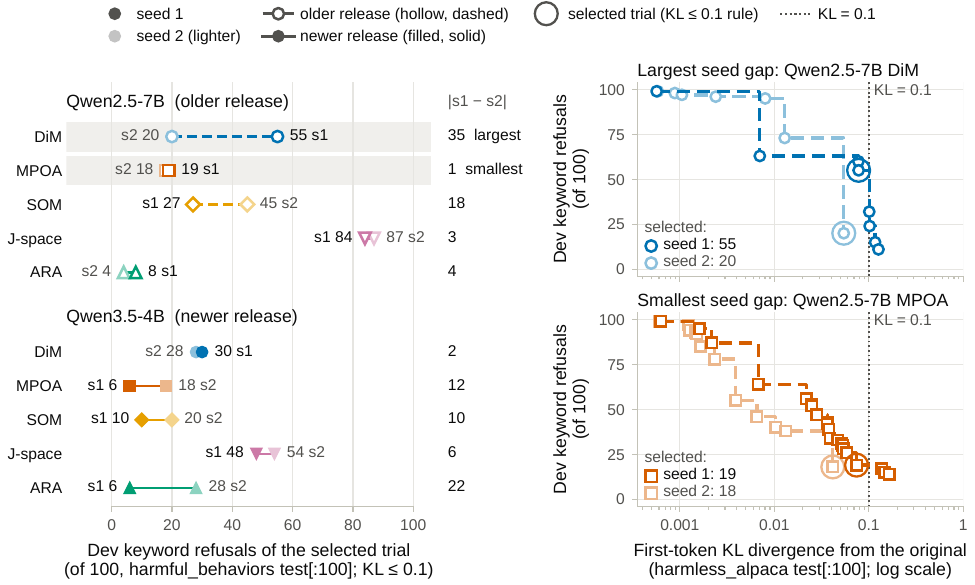}
\caption{\textbf{Seed sensitivity of the selected trial (dev set, small pair).} Each row is one model $\times$ method cell of the small Qwen pair, optimised twice with 100 TPE trials that differ only in the seed (same configuration, Heretic commit and RTX 4090). The two markers give the dev keyword refusals (of 100) of the trial chosen by the selection rule (fewest dev keyword refusals at first-token KL $\le$ 0.1) in the seed-1 run (full colour) and the seed-2 run (lighter); the column on the right is their absolute difference. Hollow markers and dashed lines mark the older model (Qwen2.5-7B), filled markers and solid lines the newer one (Qwen3.5-4B). The right panels show both seeds' Pareto fronts for the cell with the largest difference (Qwen2.5-7B DiM, 55 vs 20) and the cell with the smallest (Qwen2.5-7B MPOA, 19 vs 18). In the fronts, the rings mark each seed's selected trial and the dotted line the KL = 0.1 threshold. All counts are keyword matches on the 100 dev prompts used for optimisation and selection, not held-out refusal rates; seed 1 is the run evaluated elsewhere in the paper.}
\label{fig:seed-dev}
\end{figure*}

\paragraph{Seed sensitivity.} On the small pair a second seed changed the
selected dev count by a median of 8 and up to 35 refusals
(Figure~\ref{fig:seed-dev}), and held-out refusal differed between seeds in
all 10 cells after Holm correction (median 9.4~pp, maximum 25.2~pp;
item-level $\kappa=0.08$--0.79; Figure~\ref{fig:seed-heldout}); these
differences give the run-to-run standard deviation used in
\S\ref{sec:res-h1}. The ARA
generation effect was +14.2~pp with seed~1 and +0.0~pp [$-$1.2, +1.2]
($p=1.00$) with seed~2, the direction-based effects changed from +34.1, +4.2,
+9.7 and +21.2~pp (DiM, MPOA, SOM, J-space) to +44.3, +30.1, +29.6 and
+10.6~pp, and the H2 risk-difference contrast was $-$3.1~pp with seed~1 and
$-$28.7~pp [$-$31.0, $-$26.5] with seed~2. Method accounted for $\eta^2=0.94$ and 0.88
of the variance among the ten cell rates of Qwen2.5-7B and Qwen3.5-4B
($F(4,5)=19.46$ and 9.58, $p=0.003$ and 0.015).

\paragraph{Over-refusal.} WildGuard refusal on the 350 benign held-out prompts
fell from 4.3--22.3\% in the originals to 0.0--7.4\% in the 42 primary models,
significantly in 41 of them (exact McNemar, Holm over all 64 modified models);
only Gemma-4-31B DiM, the failed edit, did not change significantly (+1.1~pp,
$p_\mathrm{Holm}=0.250$).

\paragraph{Harmful compliance.} Residual refusal is not the same as harmful
compliance. Of the ablated main-study responses that WildGuard did not label
as refusals, 75.1\% [73.4, 76.9] were harmful according to the HarmBench
classifier and 55.8\% [53.3, 58.1] according to WildGuard's harmful-response
label (all ten originals: 79.0\% and 29.6\%), so 16.8\% [15.7, 18.0] of all ablated
main-study responses were neither refused nor harmful by HarmBench. With
harmful compliance as the outcome, coded as not harmful so that a positive
generation effect again means that removal left less on the newer model, the
pooled conclusions were the same (Appendix~\ref{app:harm}). The
baseline-adjusted direction-based effect was +0.05 log-odds [$-$0.02, +0.11]
(HarmBench) and $-$0.03 [$-$0.13, +0.07] (WildGuard), against +0.03 for
refusal; the H2 interaction was $-$0.49 [$-$0.56, $-$0.41] and $-$0.45
[$-$0.52, $-$0.38], against $-$0.50; and per pair the interaction was in the
predicted direction for Gemma and OLMo and opposite for Qwen under every
outcome. Several per-pair results changed. Under HarmBench every Qwen method
was significantly higher on the newer model, so the Qwen pair met the H1
criterion on this outcome: Qwen SOM, Gemma MPOA and Gemma SOM became
significantly higher, Mistral Small DiM lost significance and Mistral Small
ARA became significantly lower. The raw pooled ARA effect became significant
(+0.21 [+0.14, +0.28] and +0.24 [+0.18, +0.30], against +0.12 for refusal),
and the Holm tally of the 16 main-pair tests was 11 higher, 3 lower and 2 not
significant (HarmBench) and 10, 1 and 5 (WildGuard), against 8, 4 and 4.

\paragraph{Capability and divergence.} Across the 42 primary models, mean
capability change ranged from $-$4.2~pp (OLMo-2-32B ARA) to +0.9~pp
(Qwen2.5-7B J-space; Figure~\ref{fig:capability},
Tables~\ref{tab:capability-main}--\ref{tab:capability-small}), and
teacher-forced divergence from 0.004 to 0.279 nats per token
(Figure~\ref{fig:divergence},
Tables~\ref{tab:teacher-forced-main}--\ref{tab:teacher-forced-small}). Across
the 56 modified models with a dev trial, capability change was not correlated
with selected dev KL (Spearman $\rho=-0.03$ [$-$0.29, +0.24], $p=0.81$) but
was with teacher-forced divergence ($\rho=-0.48$ [$-$0.66, $-$0.24],
$p<0.001$); every selected trial had KL~$\le 0.1$ by design. Among the 42
primary models alone, the correlations were $\rho=+0.08$ [$-$0.22, +0.38]
($p=0.63$) with dev KL and $-0.39$ [$-$0.63, $-$0.11] ($p=0.010$) with
teacher-forced divergence.

\begin{figure*}[!htbp]
\centering
\includegraphics[width=\textwidth]{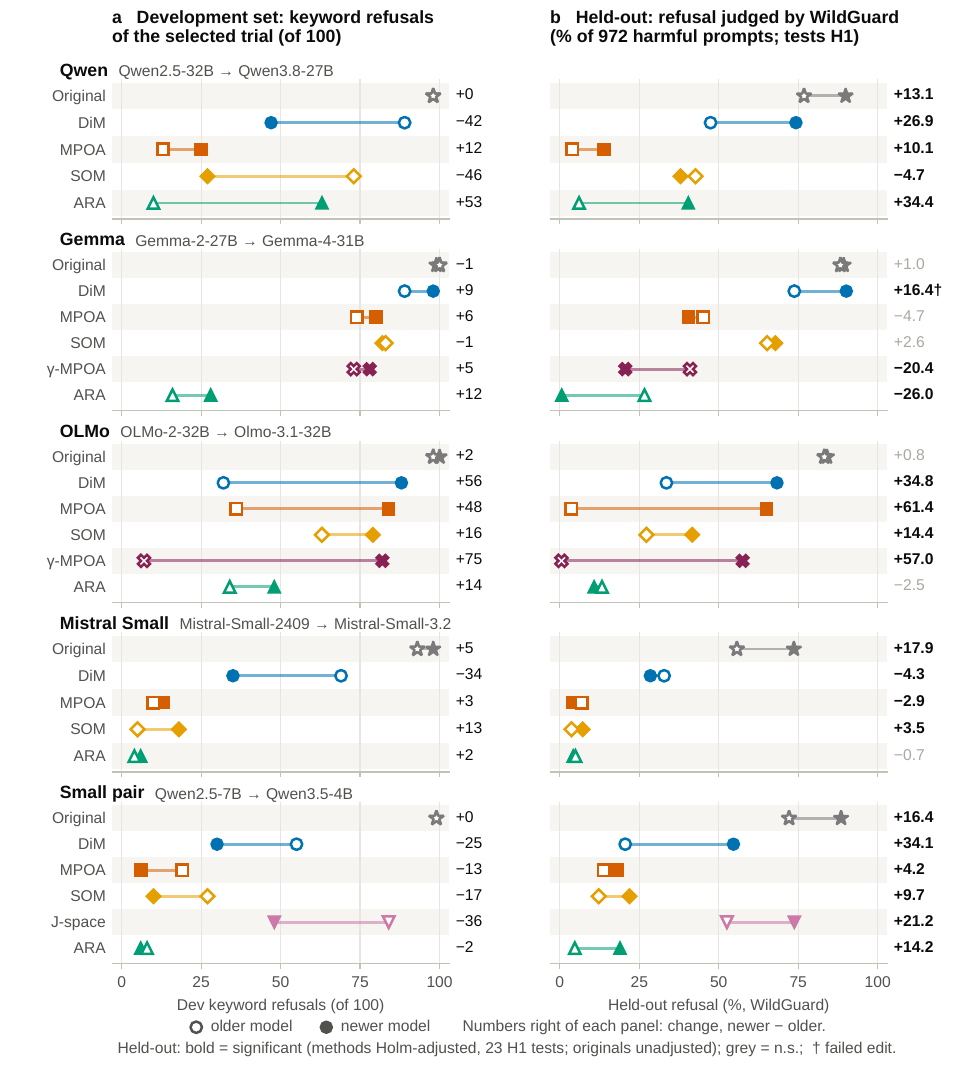}
\caption{\textbf{Generation effect on the development set (a) and on held-out prompts (b).} One block per pair, one row per condition; hollow = older, filled = newer model; the number right of each panel is the change, newer $-$ older. (a) Dev keyword refusals (of 100) of the trial picked by the selection rule (fewest dev refusals at first-token KL $\le$ 0.1, seed 1); the Original row gives the unedited models. These prompts were also used for search and selection, so the counts are optimistically biased and untested; seed-to-seed variation is shown in Figure~\ref{fig:seed-dev}. (b) Held-out refusal judged by WildGuard, paired over the 972 harmful prompts: the H1 measure, with the same values as Figure~\ref{fig:heldout-generation} and Table~\ref{tab:h1-heldout}. Bold: significant; grey: not significant. Significance for the methods is Holm-adjusted (exact McNemar, Holm within the 23 H1 tests); significance for the originals and $\gamma$-MPOA is unadjusted (exact McNemar), because they are not H1 tests; $\dagger$ failed edit (held-out refusal not below the model's own original). $\gamma$-MPOA was run on the Gemma and OLMo pairs only, J-space on the small pair only. Single seed; 95\% Wilson intervals for (b) are in Figure~\ref{fig:heldout-generation}.}
\label{fig:dev-generation}
\end{figure*}

\begin{figure*}[!htbp]
\centering
\includegraphics[width=\textwidth]{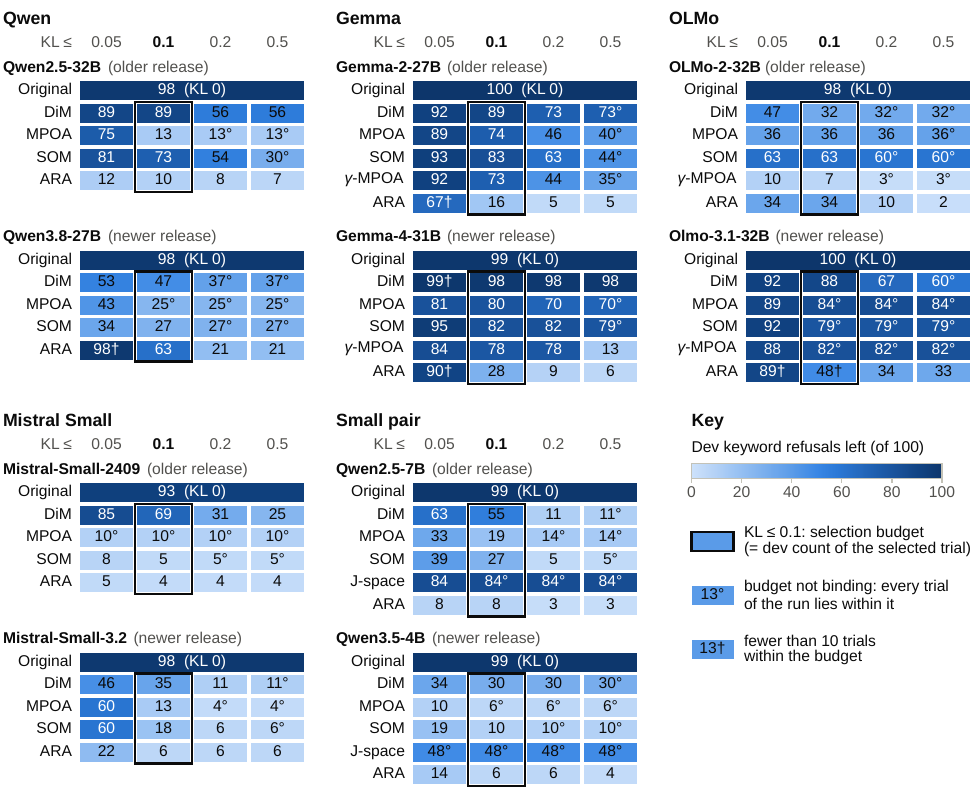}
\caption{\textbf{How much refusal is left at a given KL budget (dev set).} Each cell is the fewest dev keyword refusals (of 100) among the completed TPE trials of one optimisation run whose first-token KL divergence from the original is at most the column budget $t$. Within each pair the upper block is the older and the lower block the newer release; each block starts with the unedited original (KL 0). Darker cells mean more refusal survives. The outlined $t = 0.1$ column is the selection budget, so it equals the dev refusals of the selected models. A degree sign ($^\circ$) marks a budget that is not binding (60 cells: every completed trial already lies within it, so larger budgets were never explored and the value is the run's overall minimum); a dagger marks a minimum over fewer than 10 trials (6 cells). The columns are post-hoc budgets on one 100-trial search per run (seed 1), not separate optimisations, and the counts are dev-set numbers, not held-out refusal rates.}
\label{fig:kl-budget}
\end{figure*}

\paragraph{Development set versus held-out refusal.} On the dev prompts used
for search and selection (Figures~\ref{fig:pareto}--\ref{fig:kl-budget},
Table~\ref{tab:dev-refusals-main}), the selected keyword count changed from
older to newer model by $-$2.8 (DiM), +17.3
(MPOA), +20.3 (ARA) and $-$4.5 (SOM) refusals on average over the main pairs
(Figure~\ref{fig:dev-generation}). The newer models' ARA counts (Qwen, Gemma,
OLMo, Mistral Small) were 63, 28, 48 and 6 at a KL budget of 0.1 and 21, 9, 34
and 6 at 0.2 (Figure~\ref{fig:kl-budget}). Gemma-4-31B ARA, with 28 dev
keyword refusals, had 22.3\% [19.8, 25.0] held-out keyword refusals and 0.6\%
held-out WildGuard refusals. Dev refusals correlated with held-out refusal
across the 42 primary models ($\rho=0.77$ [0.61, 0.87]; older models 0.85,
newer 0.73), and the best dev method was the
best held-out method in 7 of 12 model $\times$ seed sets (6 of 12 without the
KL tie-break).
The direction of the change could differ between the two sets: in 3 of the 16
main-pair method cells (Qwen DiM, Gemma ARA, Mistral Small MPOA) the dev change
and a significant held-out change had opposite signs, and on the small pair
every method's dev count fell from the older to the newer model while every
held-out rate rose (Figure~\ref{fig:dev-generation}).

\paragraph{Re-scoring the released exports.} The 46 released merged exports were
re-scored on the development set with Heretic's own evaluator. Their keyword
refusal counts were within two of the selected trials' counts for 37 of 46
(median change +1, range $-$7 to +9, largest for ARA), first-token KL changed
by a median of +0.0003, and four exports (Gemma-2-27B MPOA, Gemma-4-31B ARA
and the $\gamma$-MPOA controls of Gemma-2-27B and Olmo-3.1-32B) re-scored
slightly above the KL budget, at 0.103 to 0.107
(Appendix~\ref{app:dev-rescore}). The largest changes were in ARA exports
whose adapter had been refitted on a different GPU from the search
(\S\ref{sec:res-h2}). The re-score also showed that the five
Gemma-2-27B exports had been saved without the eager-attention setting that
transformers needs for Gemma-2's attention-logit soft-capping; their public
configurations were patched, and the held-out results are unaffected because
all generations, benchmarks and teacher-forced divergences were computed in
vLLM, which applies the soft-cap from the configuration regardless of this
setting.

\begin{figure*}[!htbp]
\centering
\includegraphics[width=\textwidth]{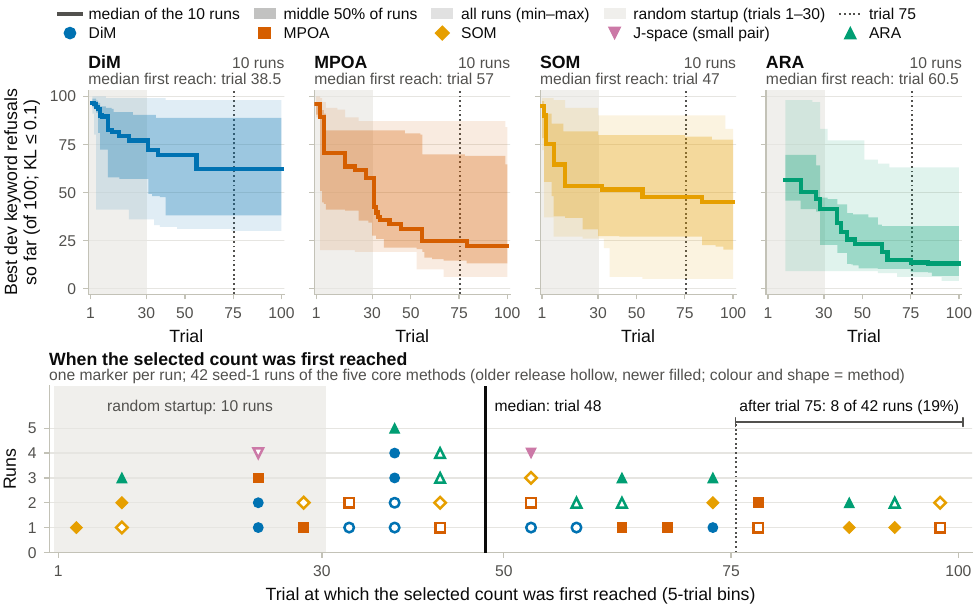}
\caption{\textbf{Search-budget convergence of the 100-trial TPE searches (dev set).} Top: for each core method with ten runs (seed 1; five pairs, older and newer release), the median over the runs of the best count so far: after trial $t$, the fewest dev keyword refusals (of 100) among trials 1 to $t$ with first-token KL divergence $\le$ 0.1 from the original, i.e. the count the selection rule would have returned had the search stopped at trial $t$. Dark band: middle 50\% of the runs (25th--75th percentile); light band: all runs (minimum to maximum). Curves start at the run's first trial within the KL budget (trials 5--10 for ARA, 1--2 otherwise); the median and bands start once every run has a value. The grey band marks the 30 random startup trials of the TPE sampler, the dotted line trial 75. Bottom: the trial at which each of the 42 core runs (DiM, MPOA, ARA, SOM and, on the small pair, J-space) first reached the count it finally selected (median 48; 10 runs within the random startup); hollow markers are older, filled markers newer releases, colour and shape the method. The 8 runs (19\%) that still improved after trial 75, by 1--14 refusals, show that the selected counts are not converged minima; for the other 34 runs these data cannot tell whether a longer search would have helped. Every run is drawn as its own curve, by method and release, in Figure~\ref{fig:search-runs}, together with the two J-space runs and the 4 $\gamma$-MPOA runs (Gemma and OLMo pairs only; not counted in the histogram). All counts are keyword matches on the 100 dev prompts that were also used for search and selection, not held-out refusal rates.}
\label{fig:search-convergence}
\end{figure*}

\paragraph{Search convergence.} Of the 42 core runs, 34 reached their
selected count by trial 75 (median trial 48) and 8 (19\%) improved later, by
1--14 refusals (Figures~\ref{fig:search-convergence}
and~\ref{fig:search-runs}); late runs were equally frequent on older and newer
models (4 of 21 each; Fisher $p=1.0$). The failed Gemma-4-31B DiM run
reached its selected count (98) at trial 40.

\clearpage
\subsection{Run-to-Run Variance as a Function of the Refusal Level}
\label{app:sigma-level}

The per-pair intervals of Table~\ref{tab:pair-effects} add a run-to-run
variance $\sigma^2$ per optimised cell, with $\sigma=9.15$~pp [6.40, 16.07]
estimated from the ten small-pair seed pairs and assumed constant across
models, methods and refusal levels (\S\ref{sec:setup-stats}). A binomial-type
rate varies less near 0\% and 100\%, so a second model holds the run effect
constant on the logit scale, $\sigma_L=0.83$ [0.58, 1.45] from the same seed
pairs, and maps it to percentage points at each cell's observed rate $p$ by
the delta method, $\sigma_{\mathrm{pp}}=\sigma_L\,p(1-p)$. This gives about
0.5~pp for Gemma-4-31B ARA (0.6\% refusal) and up to 20.7~pp at 50\% refusal.
The ten seed pairs do not distinguish the two models (log-likelihood of the
pp differences 6.25 under the constant model and 6.01 under the logit model;
rank correlation between $|d|$ and $\bar p(1-\bar p)$ $-0.01$). Under the
logit model, 1 of the 16 main-study per-method effects (OLMo MPOA) exceeded
twice the run-to-run standard deviation of a difference, against 5 under the
constant model; across the 95\% interval of $\sigma_L$, 0 to 3 did. Per pair
(Table~\ref{tab:sigma-level}), the OLMo direction-based increase excluded zero
under both models except at the upper limit of the logit model, whereas the
Qwen ARA increase, the Gemma ARA decrease and the Gemma H2 contrast excluded
zero only under the constant model. Table~\ref{tab:heterogeneity} gives the
between-pair heterogeneity with each kind of within-pair variance, including
the between-pair variance $\tau^2$ of every estimand, and
Table~\ref{tab:pair-variances} the within-pair variances used as weights. The
headline $I^2$ of the raw direction-based pp effect is therefore 80\% only
under the constant variance model at the point estimate of $\sigma$; under
the refusal-level model it is 46\% ($\tau^2=127$~pp$^2$ against 225~pp$^2$).

\input{tables/t10_sigma_level}
\input{tables/t11b_heterogeneity_tau}
\input{tables/t14_pair_variances}

\subsection{Verdicts at Other KL Budgets (Development Set)}
\label{app:kl-budgets}

Only the trial selected at first-token KL~$\le 0.1$ was evaluated on held-out
prompts, so other budgets can be read only on the 100 dev prompts, from the
completed trials of each search. For each run the refusal--KL front $f(t)$,
the fewest dev keyword refusals among trials with KL~$\le t$ (the original at
KL~$=0$ included), was read at $t=0.05$, 0.1, 0.2, 0.5 and 1.0, and summarised
without a budget as the area under the front, the mean of $f(t)$ over 200
log-spaced budgets between 0.01 and 1. Run-to-run variation of each quantity
was estimated from the ten small-pair seed pairs. In 7 of the 42 seed-1 runs
the front was carried forward beyond the largest sampled KL over more than
half of this range. Table~\ref{tab:kl-budgets} gives the class-level generation
effects. The direction-based effects kept their sign from KL~$\le 0.1$ to
KL~$\le 1.0$ in all four main pairs, and the OLMo direction-based increase
exceeded run-to-run variation at every budget and on the area. The dev ARA
effects had the same sign at every budget but the opposite sign to the
held-out ARA effects in three of the four main pairs, as at the selected
budget (\S\ref{sec:res-secondary}). The H2 contrast was positive at small
budgets and negative at large ones in Gemma, negative from KL~$\le 0.1$ on in
OLMo, and positive in Qwen at every budget.

\input{tables/t12_kl_budgets}

\subsection{Outcome Classes}
\label{app:outcome-classes}

Table~\ref{tab:outcome-classes} gives the four outcome classes of
\S\ref{sec:setup-eval} for every original and primary model. Degenerate
responses were at most 0.6\% of any of these models' responses. Among the 42 primary
models, incomplete compliance (generation stopped at the 512-token limit)
ranged from 7.6\% to 79.9\% of responses and complete compliance from 2.3\%
to 61.3\%; 65.8\% of all non-refused responses were incomplete. Incomplete
compliance is a property of the 512-token budget, not evidence that a
response was unhelpful.

\input{tables/t13_outcome_classes}

\subsection{Mixed-Model Estimates and Conditional Retention}
\label{app:rr-tables}

Table~\ref{tab:h2-vb} gives the planned mixed logistic model of the H2
interaction next to the GEE estimates of Table~\ref{tab:h2-contrasts}. Its
conditional coefficients have the same sign as the GEE coefficients in every
pair and pooled and are larger in magnitude, as expected for a conditional
model, but the model has no pair-specific slopes and its intervals describe
prompt sampling only. Table~\ref{tab:retention} gives the generation effects
on conditional refusal retention with prompt-bootstrap intervals and with
intervals that add a run-to-run standard deviation of retention of 11.0~pp
[7.7, 19.2] per cell, estimated from the ten small-pair seed pairs. With
run-to-run variance, the OLMo direction-based increase (+43.4~pp [+25.7,
+61.1]), the OLMo H2 contrast ($-$46.6~pp [$-$81.8, $-$11.3]) and the Qwen ARA
increase (+36.9~pp [+6.3, +67.5]) excluded zero; no other class-level effect
or H2 contrast did.

\input{tables/t15_h2_vb}
\input{tables/t16_retention}

%% file: tables/t10_sigma_level.tex
\begin{table*}[t]
\centering\scriptsize
\setlength{\tabcolsep}{3pt}
\caption{Per-pair class effects (pp) under two models of run-to-run variance. Constant: $\sigma=9.15$~pp per cell [6.40, 16.07] (the intervals of Table~\ref{tab:pair-effects}). Refusal-level: $\sigma_L=0.83$ [0.58, 1.45] per cell on the logit scale, mapped to pp at each cell's rate $p$ as $\sigma_L\,p(1-p)$. Intervals add prompt sampling and the run-to-run variance at the point estimate of $\sigma$; the last two columns mark (\checkmark) whether the interval excludes zero at the lower limit, point estimate and upper limit of $\sigma$ (lo\,/\,pt\,/\,hi).}
\label{tab:sigma-level}
\begin{tabular}{@{}lllllcc@{}}
\toprule
Pair & Estimand & Effect & Constant $\sigma$ & Refusal-level $\sigma$ & Excl.\ 0, const. & Excl.\ 0, level \\
 & & & [95\% interval] & [95\% interval] & lo/pt/hi & lo/pt/hi \\
\midrule
Qwen & Direction, raw & +10.7 & [$-$4.0, +25.5] & [$-$15.3, +36.7] & \checkmark/--/-- & --/--/-- \\
 & Direction, adjusted & $-$2.4 & [$-$17.3, +12.5] & [$-$28.5, +23.7] & --/--/-- & --/--/-- \\
 & ARA, raw & +34.4 & [+8.8, +59.9] & [$-$5.9, +74.6] & \checkmark/\checkmark/-- & \checkmark/--/-- \\
 & ARA, adjusted & +21.3 & [$-$4.4, +47.0] & [$-$19.0, +61.7] & \checkmark/--/-- & --/--/-- \\
 & H2 contrast & +23.6 & [$-$5.9, +53.1] & [$-$24.3, +71.6] & \checkmark/--/-- & --/--/-- \\
\addlinespace
Gemma & Direction, raw & +4.7 & [$-$10.2, +19.6] & [$-$23.2, +32.6] & --/--/-- & --/--/-- \\
 & Direction, adjusted & +3.7 & [$-$11.2, +18.6] & [$-$24.2, +31.6] & --/--/-- & --/--/-- \\
 & ARA, raw & $-$26.0 & [$-$51.6, $-$0.5] & [$-$57.9, +5.8] & \checkmark/\checkmark/-- & \checkmark/--/-- \\
 & ARA, adjusted & $-$27.1 & [$-$52.7, $-$1.5] & [$-$59.0, +4.9] & \checkmark/\checkmark/-- & \checkmark/--/-- \\
 & H2 contrast & $-$30.8 & [$-$60.2, $-$1.3] & [$-$73.1, +11.5] & \checkmark/\checkmark/-- & \checkmark/--/-- \\
\addlinespace
OLMo & Direction, raw & +36.9 & [+22.1, +51.7] & [+9.8, +63.9] & \checkmark/\checkmark/\checkmark & \checkmark/\checkmark/-- \\
 & Direction, adjusted & +36.0 & [+21.2, +50.9] & [+9.0, +63.1] & \checkmark/\checkmark/\checkmark & \checkmark/\checkmark/-- \\
 & ARA, raw & $-$2.5 & [$-$27.9, +23.0] & [$-$26.9, +22.0] & --/--/-- & --/--/-- \\
 & ARA, adjusted & $-$3.3 & [$-$28.8, +22.2] & [$-$27.8, +21.2] & --/--/-- & --/--/-- \\
 & H2 contrast & $-$39.3 & [$-$68.8, $-$9.9] & [$-$75.8, $-$2.9] & \checkmark/\checkmark/-- & \checkmark/\checkmark/-- \\
\addlinespace
Mistral Small & Direction, raw & $-$1.2 & [$-$16.0, +13.5] & [$-$18.5, +16.1] & --/--/-- & --/--/-- \\
 & Direction, adjusted & $-$19.2 & [$-$34.1, $-$4.2] & [$-$36.6, $-$1.7] & \checkmark/\checkmark/-- & \checkmark/\checkmark/-- \\
 & ARA, raw & $-$0.7 & [$-$26.1, +24.7] & [$-$10.8, +9.4] & --/--/-- & --/--/-- \\
 & ARA, adjusted & $-$18.6 & [$-$44.2, +6.9] & [$-$29.1, $-$8.2] & \checkmark/--/-- & \checkmark/\checkmark/\checkmark \\
 & H2 contrast & +0.5 & [$-$28.8, +29.9] & [$-$19.5, +20.5] & --/--/-- & --/--/-- \\
\addlinespace
Small pair & Direction, raw & +17.3 & [+4.4, +30.1] & [$-$3.7, +38.2] & \checkmark/\checkmark/-- & \checkmark/--/-- \\
 & Direction, adjusted & +0.9 & [$-$12.1, +14.0] & [$-$20.2, +22.0] & --/--/-- & --/--/-- \\
 & ARA, raw & +14.2 & [$-$11.3, +39.7] & [$-$11.9, +40.3] & --/--/-- & --/--/-- \\
 & ARA, adjusted & $-$2.2 & [$-$27.8, +23.5] & [$-$28.4, +24.1] & --/--/-- & --/--/-- \\
 & H2 contrast & $-$3.1 & [$-$31.6, +25.4] & [$-$36.5, +30.3] & --/--/-- & --/--/-- \\
\bottomrule
\end{tabular}
\end{table*}

%% file: tables/t11b_heterogeneity_tau.tex
\begin{table}[t]
\centering\scriptsize
\setlength{\tabcolsep}{2pt}
\caption{Between-pair heterogeneity of the four main-pair effects with three kinds of within-pair variance: prompt sampling only, prompt sampling plus constant run-to-run variance, and (pp scale only) prompt sampling plus refusal-level run-to-run variance, each at the point estimate of $\sigma$. Each cell gives Cochran's $Q$ (3~df); its $p$ value; $I^2$; and the DerSimonian--Laird between-pair variance $\tau^2$ (pp$^2$ on the pp scale, squared log-odds on the log-odds scale). Across the 95\% interval of $\sigma$, $I^2$ of the raw direction-based pp effect is 90\% (lower limit) to 39\% (upper limit) with constant and 73\% to 0\% with refusal-level variance. The within-pair variances are in Table~\ref{tab:pair-variances}.}
\label{tab:heterogeneity}
\begin{tabular}{@{}lccccc@{}}
\toprule
 & \multicolumn{3}{c}{pp scale} & \multicolumn{2}{c}{log-odds scale} \\
\cmidrule(lr){2-4}\cmidrule(l){5-6}
Estimand & Prompt only & + seed, const. & + seed, level & Prompt only & + seed \\
\midrule
Direction, raw & 821.8; $<$0.001; 100\%; 284 & 14.8; 0.002; 80\%; 225 & 5.5; 0.136; 46\%; 127 & 590.8; $<$0.001; 99\%; 0.56 & 3.8; 0.289; 20\%; 0.12 \\
Direction, adjusted & 849.4; $<$0.001; 100\%; 527 & 27.7; $<$0.001; 89\%; 476 & 11.5; 0.010; 74\%; 425 & 608.6; $<$0.001; 100\%; 1.53 & 7.7; 0.053; 61\%; 0.72 \\
ARA, raw & 807.1; $<$0.001; 100\%; 321 & 10.9; 0.012; 73\%; 447 & 5.3; 0.148; 44\%; 113 & 378.5; $<$0.001; 99\%; 2.86 & 14.3; 0.003; 79\%; 5.34 \\
ARA, adjusted & 381.8; $<$0.001; 99\%; 328 & 7.9; 0.048; 62\%; 278 & 5.0; 0.172; 40\%; 97 & 208.5; $<$0.001; 99\%; 1.95 & 10.8; 0.013; 72\%; 3.74 \\
H2 contrast & 1006.7; $<$0.001; 100\%; 680 & 11.2; 0.011; 73\%; 614 & 6.3; 0.097; 53\%; 343 & 539.7; $<$0.001; 99\%; 4.54 & 10.8; 0.013; 72\%; 4.87 \\
\bottomrule
\end{tabular}
\end{table}

%% file: tables/t14_pair_variances.tex
\begin{table}[t]
\centering\scriptsize
\setlength{\tabcolsep}{3pt}
\caption{Within-pair variances of the four main-pair effects used as inverse-variance weights in Table~\ref{tab:heterogeneity}, at the point estimate of $\sigma$: prompt sampling only, and prompt sampling plus constant or refusal-level run-to-run variance. pp scale in pp$^2$; log-odds scale in squared log-odds. The constant run-to-run term is the same for every pair (2$\sigma^2$ per method, 2$\sigma^2/3$ for the direction-class mean), so on that variance model the pairs are weighted almost equally.}
\label{tab:pair-variances}
\begin{tabular}{@{}llrrrrr@{}}
\toprule
 & & \multicolumn{3}{c}{pp scale (pp$^2$)} & \multicolumn{2}{c}{log-odds scale} \\
\cmidrule(lr){3-5}\cmidrule(l){6-7}
Estimand & Pair & Prompt only & + seed, const. & + seed, level & Prompt only & + seed \\
\midrule
Direction, raw & Qwen & 1.0 & 56.8 & 176.0 & 0.002 & 0.459 \\
 & Gemma & 1.9 & 57.8 & 202.9 & 0.004 & 0.461 \\
 & OLMo & 1.2 & 57.1 & 190.0 & 0.003 & 0.459 \\
 & Mistral Small & 0.6 & 56.4 & 77.9 & 0.004 & 0.461 \\
\addlinespace
Direction, adjusted & Qwen & 1.9 & 57.8 & 177.0 & 0.009 & 0.466 \\
 & Gemma & 1.6 & 57.5 & 202.6 & 0.011 & 0.468 \\
 & OLMo & 1.7 & 57.5 & 190.5 & 0.005 & 0.462 \\
 & Mistral Small & 2.3 & 58.2 & 79.6 & 0.007 & 0.464 \\
\addlinespace
ARA, raw & Qwen & 2.6 & 170.2 & 422.4 & 0.019 & 1.389 \\
 & Gemma & 2.0 & 169.6 & 264.1 & 0.165 & 1.536 \\
 & OLMo & 1.2 & 168.8 & 155.6 & 0.010 & 1.381 \\
 & Mistral Small & 0.4 & 168.0 & 26.7 & 0.022 & 1.393 \\
\addlinespace
ARA, adjusted & Qwen & 4.3 & 171.9 & 424.1 & 0.025 & 1.396 \\
 & Gemma & 3.0 & 170.6 & 265.1 & 0.178 & 1.549 \\
 & OLMo & 1.6 & 169.2 & 156.1 & 0.013 & 1.384 \\
 & Mistral Small & 2.3 & 169.9 & 28.6 & 0.026 & 1.396 \\
\addlinespace
H2 contrast & Qwen & 3.3 & 226.7 & 598.2 & 0.019 & 1.847 \\
 & Gemma & 2.7 & 226.2 & 465.7 & 0.167 & 1.995 \\
 & OLMo & 2.3 & 225.7 & 345.5 & 0.012 & 1.840 \\
 & Mistral Small & 0.9 & 224.4 & 104.5 & 0.025 & 1.853 \\
\bottomrule
\end{tabular}
\end{table}

%% file: tables/t12_kl_budgets.tex
\begin{table*}[t]
\centering\scriptsize
\setlength{\tabcolsep}{3.5pt}
\caption{Generation effects on the development set at several first-token KL budgets (newer minus older model; dev keyword refusals of 100 of the best completed trial at KL $\le t$, including the original at KL = 0) and on the area under the refusal--KL front (mean front value over 200 log-spaced budgets from 0.01 to 1). Direction: mean of the direction-based methods; H2: ARA minus direction. $^{*}$Beyond 1.96 run-to-run standard errors, from the ten small-pair seed pairs on the same quantity (run-to-run SD per cell: $\le$0.05 6.8, $\le$0.1 10.8, $\le$0.2 5.2, $\le$0.5 4.6, $\le$1.0 4.6, Area 4.3; prompt sampling of the 100 dev prompts is not included). The KL $\le$ 0.1 column belongs to the selected models; only those were evaluated on held-out prompts (last column, WildGuard, pp). Dev counts are optimised and therefore optimistic.}
\label{tab:kl-budgets}
\begin{tabular}{@{}llrrrrrrr@{}}
\toprule
Pair & Estimand & $\le$0.05 & $\le$0.1 & $\le$0.2 & $\le$0.5 & $\le$1.0 & Area & Held-out \\
\midrule
Qwen & Direction & $-$38.3$^{*}$ & $-$25.3$^{*}$ & $-$11.3$^{*}$ & $-$3.3 & $-$3.3 & $-$7.2$^{*}$ & +10.7 \\
 & ARA & +86.0$^{*}$ & +53.0$^{*}$ & +13.0 & +14.0$^{*}$ & +1.0 & +37.1$^{*}$ & +34.4 \\
 & H2 contrast & +124.3$^{*}$ & +78.3$^{*}$ & +24.3$^{*}$ & +17.3$^{*}$ & +4.3 & +44.3$^{*}$ & +23.6 \\
\addlinespace
Gemma & Direction & +0.3 & +4.7 & +22.7$^{*}$ & +30.0$^{*}$ & +30.0$^{*}$ & +13.0$^{*}$ & +4.7 \\
 & ARA & +23.0$^{*}$ & +12.0 & +4.0 & +1.0 & +1.0 & +5.1 & $-$26.0 \\
 & H2 contrast & +22.7$^{*}$ & +7.3 & $-$18.7$^{*}$ & $-$29.0$^{*}$ & $-$29.0$^{*}$ & $-$7.9 & $-$30.8 \\
\addlinespace
OLMo & Direction & +42.3$^{*}$ & +40.0$^{*}$ & +34.0$^{*}$ & +31.7$^{*}$ & +31.7$^{*}$ & +27.9$^{*}$ & +36.9 \\
 & ARA & +55.0$^{*}$ & +14.0 & +24.0$^{*}$ & +31.0$^{*}$ & +4.0 & +21.9$^{*}$ & $-$2.5 \\
 & H2 contrast & +12.7 & $-$26.0 & $-$10.0 & $-$0.7 & $-$27.7$^{*}$ & $-$6.0 & $-$39.3 \\
\addlinespace
Mistral Small & Direction & +21.0$^{*}$ & $-$6.0 & $-$8.3 & $-$6.3 & $-$6.3 & +6.5 & $-$1.2 \\
 & ARA & +17.0 & +2.0 & +2.0 & +2.0 & +2.0 & +23.6$^{*}$ & $-$0.7 \\
 & H2 contrast & $-$4.0 & +8.0 & +10.3 & +8.3 & +8.3 & +17.1$^{*}$ & +0.5 \\
\addlinespace
Small pair & Direction & $-$27.0$^{*}$ & $-$22.8$^{*}$ & $-$5.0 & $-$5.0 & $-$5.0 & $-$11.6$^{*}$ & +17.3 \\
 & ARA & +6.0 & $-$2.0 & +3.0 & +1.0 & +1.0 & +4.0 & +14.2 \\
 & H2 contrast & +33.0$^{*}$ & +20.8 & +8.0 & +6.0 & +6.0 & +15.5$^{*}$ & $-$3.1 \\
\bottomrule
\end{tabular}
\end{table*}

%% file: tables/t13_outcome_classes.tex
\begin{table*}[t]
\centering\scriptsize
\setlength{\tabcolsep}{3.5pt}
\caption{Outcome classes on the 972 harmful held-out prompts (\% of responses), older and newer model of each pair: refusal (WildGuard, non-degenerate responses), complete compliance, incomplete compliance (generation stopped at the 512-token limit) and degenerate responses (Methods). Rows may not add to 100 because of rounding and a few unscored responses.}
\label{tab:outcome-classes}
\begin{tabular}{@{}llrrrrrrrr@{}}
\toprule
 & & \multicolumn{4}{c}{Older model} & \multicolumn{4}{c}{Newer model} \\
\cmidrule(lr){3-6}\cmidrule(l){7-10}
Pair & Condition & Ref. & Compl. & Incompl. & Degen. & Ref. & Compl. & Incompl. & Degen. \\
\midrule
Qwen & Original & 76.7 & 5.7 & 17.4 & 0.1 & 89.9 & 4.2 & 5.8 & 0.1 \\
 & DiM & 47.4 & 18.8 & 33.6 & 0.1 & 74.3 & 5.2 & 20.4 & 0.1 \\
 & MPOA & 3.9 & 50.6 & 45.4 & 0.1 & 14.0 & 8.7 & 77.3 & 0.0 \\
 & SOM & 42.7 & 20.4 & 36.8 & 0.1 & 38.0 & 14.0 & 47.8 & 0.2 \\
 & ARA & 6.1 & 61.3 & 32.5 & 0.1 & 40.3 & 4.5 & 54.9 & 0.2 \\
\addlinespace
Gemma & Original & 88.3 & 2.2 & 9.6 & 0.0 & 89.3 & 2.3 & 8.4 & 0.0 \\
 & DiM & 73.8 & 8.4 & 17.8 & 0.0 & 90.1 & 2.3 & 7.6 & 0.0 \\
 & MPOA & 45.2 & 27.1 & 27.8 & 0.0 & 40.4 & 11.0 & 48.6 & 0.0 \\
 & SOM & 65.2 & 13.6 & 21.2 & 0.0 & 67.7 & 9.9 & 22.3 & 0.1 \\
 & ARA & 26.6 & 46.1 & 27.3 & 0.0 & 0.6 & 19.4 & 79.9 & 0.0 \\
\addlinespace
OLMo & Original & 83.2 & 2.6 & 14.2 & 0.0 & 84.1 & 1.4 & 14.5 & 0.0 \\
 & DiM & 33.5 & 28.5 & 38.0 & 0.0 & 68.3 & 5.7 & 25.9 & 0.1 \\
 & MPOA & 3.6 & 42.0 & 54.4 & 0.0 & 65.0 & 6.3 & 28.7 & 0.0 \\
 & SOM & 27.3 & 27.4 & 45.4 & 0.0 & 41.7 & 10.2 & 48.1 & 0.0 \\
 & ARA & 13.3 & 39.3 & 47.4 & 0.0 & 10.8 & 18.9 & 70.3 & 0.0 \\
\addlinespace
Mistral Small & Original & 55.7 & 11.8 & 32.2 & 0.2 & 73.6 & 6.1 & 20.4 & 0.0 \\
 & DiM & 32.8 & 23.6 & 43.6 & 0.0 & 28.5 & 26.2 & 45.3 & 0.0 \\
 & MPOA & 6.9 & 40.4 & 52.7 & 0.0 & 4.0 & 35.6 & 60.4 & 0.0 \\
 & SOM & 3.7 & 42.1 & 54.2 & 0.0 & 7.2 & 33.6 & 59.2 & 0.0 \\
 & ARA & 4.9 & 40.9 & 54.0 & 0.1 & 4.2 & 26.5 & 69.2 & 0.0 \\
\addlinespace
Small pair & Original & 72.1 & 5.1 & 22.5 & 0.2 & 88.5 & 2.0 & 9.6 & 0.0 \\
 & DiM & 20.6 & 27.8 & 51.5 & 0.1 & 54.6 & 11.3 & 34.1 & 0.0 \\
 & MPOA & 13.8 & 32.5 & 53.3 & 0.4 & 18.0 & 22.5 & 59.2 & 0.3 \\
 & SOM & 12.2 & 32.2 & 55.0 & 0.5 & 21.9 & 21.5 & 56.2 & 0.4 \\
 & J-space & 52.6 & 11.9 & 35.3 & 0.2 & 73.8 & 4.0 & 22.2 & 0.0 \\
 & ARA & 4.7 & 30.5 & 64.3 & 0.5 & 18.9 & 16.7 & 63.8 & 0.6 \\
\bottomrule
\end{tabular}
\end{table*}

%% file: tables/t15_h2_vb.tex
\begin{table}[!htbp]
\centering\footnotesize
\caption{Planned mixed logistic model of the H2 interaction (variational Bayes; random intercepts for prompt
and, in the pooled fit, pair): posterior mean of the generation $\times$ class interaction in log-odds
$\pm$ 1.96 posterior SD, next to the GEE estimate of Table~\ref{tab:h2-contrasts}. The model has no
pair-specific generation or interaction slopes, so it cannot represent between-pair heterogeneity
(\S\ref{sec:setup-stats}).}
\label{tab:h2-vb}
\begin{tabular}{lll}
\toprule
Pair & GEE, log-odds [95\% CI] & VB, log-odds [95\% CrI] \\
\midrule
Qwen & +1.89 [+1.61, +2.16] & +2.29 [+2.14, +2.44] \\
Gemma & $-$4.27 [$-$5.07, $-$3.47] & $-$4.90 [$-$5.64, $-$4.17] \\
OLMo & $-$1.87 [$-$2.08, $-$1.65] & $-$3.25 [$-$3.49, $-$3.01] \\
Mistral Small & $-$0.06 [$-$0.37, +0.25] & $-$0.08 [$-$0.42, +0.26] \\
Small pair & +0.76 [+0.46, +1.06] & +0.80 [+0.62, +0.99] \\
\midrule
Pooled (main) & $-$0.50 [$-$0.62, $-$0.38] & $-$0.71 [$-$0.81, $-$0.60] \\
\bottomrule
\end{tabular}
\end{table}

%% file: tables/t16_retention.tex
\begin{table}[!htbp]
\centering\footnotesize
\setlength{\tabcolsep}{4pt}
\caption{Conditional refusal retention: generation effect (newer minus older model, pp) on the share of the prompts refused by a model's own original that the edited model still refused (WildGuard, 972 harmful held-out prompts). Prompt: prompt-cluster bootstrap 95\% interval (2,000 resamples). Prompt + seed: adds a run-to-run standard deviation of retention of 11.0~pp [7.7, 19.2] per cell, estimated from the ten small-pair seed pairs, with the multipliers of Table~\ref{tab:pair-effects}. $^{*}$Prompt + seed interval excludes zero.}
\label{tab:retention}
\begin{tabular}{llrll}
\toprule
Pair & Estimand & Effect & Prompt [95\% CI] & Prompt + seed [95\% CI] \\
\midrule
Qwen & Direction & +6.0 & [+3.5, +8.3] & [$-$11.7, +23.7] \\
 & ARA & +36.9 & [+33.4, +40.4] & [+6.3, +67.5]$^{*}$ \\
 & H2 contrast & +30.9 & [+26.9, +34.9] & [$-$4.5, +66.2] \\
\addlinespace
Gemma & Direction & +4.0 & [+1.5, +6.5] & [$-$13.8, +21.7] \\
 & ARA & $-$29.4 & [$-$32.5, $-$26.4] & [$-$59.9, +1.2] \\
 & H2 contrast & $-$33.3 & [$-$36.8, $-$30.0] & [$-$68.6, +1.9] \\
\addlinespace
OLMo & Direction & +43.4 & [+41.0, +45.7] & [+25.7, +61.1]$^{*}$ \\
 & ARA & $-$3.2 & [$-$5.8, $-$0.5] & [$-$33.7, +27.3] \\
 & H2 contrast & $-$46.6 & [$-$49.8, $-$43.3] & [$-$81.8, $-$11.3]$^{*}$ \\
\addlinespace
Mistral Small & Direction & $-$6.6 & [$-$9.0, $-$4.3] & [$-$24.3, +11.1] \\
 & ARA & $-$2.6 & [$-$4.8, $-$0.5] & [$-$33.0, +27.9] \\
 & H2 contrast & +4.0 & [+1.2, +6.8] & [$-$31.2, +39.2] \\
\addlinespace
Small pair & Direction & +13.5 & [+10.7, +15.9] & [$-$1.9, +28.9] \\
 & ARA & +15.0 & [+12.0, +18.0] & [$-$15.6, +45.5] \\
 & H2 contrast & +1.5 & [$-$1.9, +5.1] & [$-$32.7, +35.6] \\
\bottomrule
\end{tabular}
\end{table}